\documentclass{article}

\usepackage{PRIMEarxiv}

\usepackage[utf8]{inputenc} % allow utf-8 input
\usepackage[T1]{fontenc}    % use 8-bit T1 fonts
\usepackage{hyperref}       % hyperlinks
\usepackage{url}            % simple URL typesetting
\usepackage{booktabs}       % professional-quality tables
\usepackage{amsfonts}       % blackboard math symbols
\usepackage{nicefrac}       % compact symbols for 1/2, etc.
\usepackage{microtype}      % microtypography
\usepackage{lipsum}
\usepackage{fancyhdr}       % header
\usepackage{graphicx}       % graphics
\graphicspath{{media/}}     % organize your images and other figures under media/ folder

\usepackage{adjustbox}
\usepackage{amsmath}
\usepackage{bookmark}
\usepackage{subcaption}
\usepackage{comment}
\usepackage{natbib}
\usepackage{cleveref}
\usepackage{pgfplotstable}
\pgfplotsset{compat=1.18}
\usepackage{longtable}
\usepackage{array}
\usepackage{tikz}
\usetikzlibrary{positioning}
\usepackage{soul}
\usepackage{longtable}% for long tables

\usepackage{csvsimple}
\usepackage{listings}

\title{Handling and Interpreting Missing Modalities in Patient Clinical Trajectories via Autoregressive Sequence Modeling
}

\author{
  Andrew Wang \\
  Brown University \\
  \texttt{andrew\_wang3@brown.edu} 
   \And
  Ellie Pavlick \\
  Brown University \\
  \texttt{ellie\_pavlick@brown.edu} 
    \And
  Ritambhara Singh \\
  Brown University \\
  \texttt{ritambhara\_singh@brown.edu} 
}

\begin{document}
\maketitle

\begin{abstract}
An active challenge in developing multimodal machine learning (ML) models for healthcare is handling missing modalities during training and deployment. As clinical datasets are inherently temporal and sparse in terms of modality presence, capturing the underlying predictive signal via diagnostic multimodal ML models while retaining model explainability remains an ongoing challenge. 
In this work, we address this by re-framing clinical diagnosis as an autoregressive sequence modeling task, utilizing causal decoders from large language models (LLMs) to model a patient's multimodal trajectory. We first introduce a missingness-aware contrastive pre-training objective that integrates multiple modalities in datasets with missingness in a shared latent space. We then show that autoregressive sequence modeling with transformer-based architectures outperforms baselines on the MIMIC-IV and eICU fine-tuning benchmarks.
Finally, we use interpretability techniques to move beyond performance boosts and find that across various patient stays, removing modalities leads to divergent behavior that our contrastive pre-training mitigates. By abstracting clinical diagnosis as sequence modeling and interpreting patient stay trajectories, we develop a framework to profile and handle missing modalities while addressing the canonical desideratum of safe, transparent clinical AI. 
\end{abstract}


\section{Introduction}
\label{sec:intro}
A patient's clinical trajectory is inherently sequential, multimodal, and irregular in terms of information availability. Their diagnosis from a human clinician, therefore, does not always occur as a single prediction at the end of their observed stay, but often as a sequence of updated beliefs given the dynamic nature of the clinical context.  Despite this, a dominant approach in multimodal clinical ML relies on static architectures that concatenate all patient data from a predefined window into a single vector (\cite{golovanevsky2025picmepipelinecontrastivemodality}).  This "bag-of-features" approach enforces a static prediction schema, preventing clinicians from auditing the \textit{sequence} of evidence that drives the model's decision.

Furthermore, existing approaches to handling missing modalities in these architectures typically fall into two paradigms: either inferring or imputing missing data that has not been clinically collected (\cite{lee2024fill}, \cite{Poette2026}), or relying on uninformative zero-padding and sample-dropping (\cite{lee2023learningmissingmodalelectronic}).  These methods either hallucinate clinical signals or destroy the underlying geometry of the latent space, both of which are detrimental to safe diagnostic modeling.

To bridge the gap between clinical reality and ML methodology, we re-frame clinical prediction from a static classification task to the autoregressive modeling of belief states, re-purposing the established machinery of transformer-based \cite{vaswani2023attentionneed} language model decoders.  To ensure this temporal decoder can process sparse, multimodal data, we first propose a novel contrastive pre-training step.  Using a hybrid objective function we term ``Masked Global Alignment", we map all available modalities and explicitly learned placeholder tokens for missing modalities into a stable global representation of patients.  Next, we use these pretrained embeddings to fine-tune and evaluate downstream models on both the MIMIC-IV (\cite{Johnson2023_MIMIC-IV})/CXR (\cite{Johnson2019-0c}) and eICU (\cite{Pollard2018}) datasets, demonstrating performance gains of sequential models over static baselines.  Finally, we conduct a mechanistic analysis of the model's temporal dynamics, illustrating the ability to transparently and mechanistically interpret the diagnostic reasoning process of autoregressive clinical sequence models.  In doing so, we also demonstrate the utility of our contrastive pre-training approach, mitigating potential sources of bias by robustly encoding and aligning latent patient data.  Taken together, our results suggest that a framework combining missingness-aware contrastive pre-training and transformer-based sequence modeling not only serves to mitigate missing-modality-associated ML performance loss but also to retain model explainability, which is crucial to trustworthy deployment and application.

\begin{figure*}
	\centering
	\includegraphics[width=0.9\linewidth]{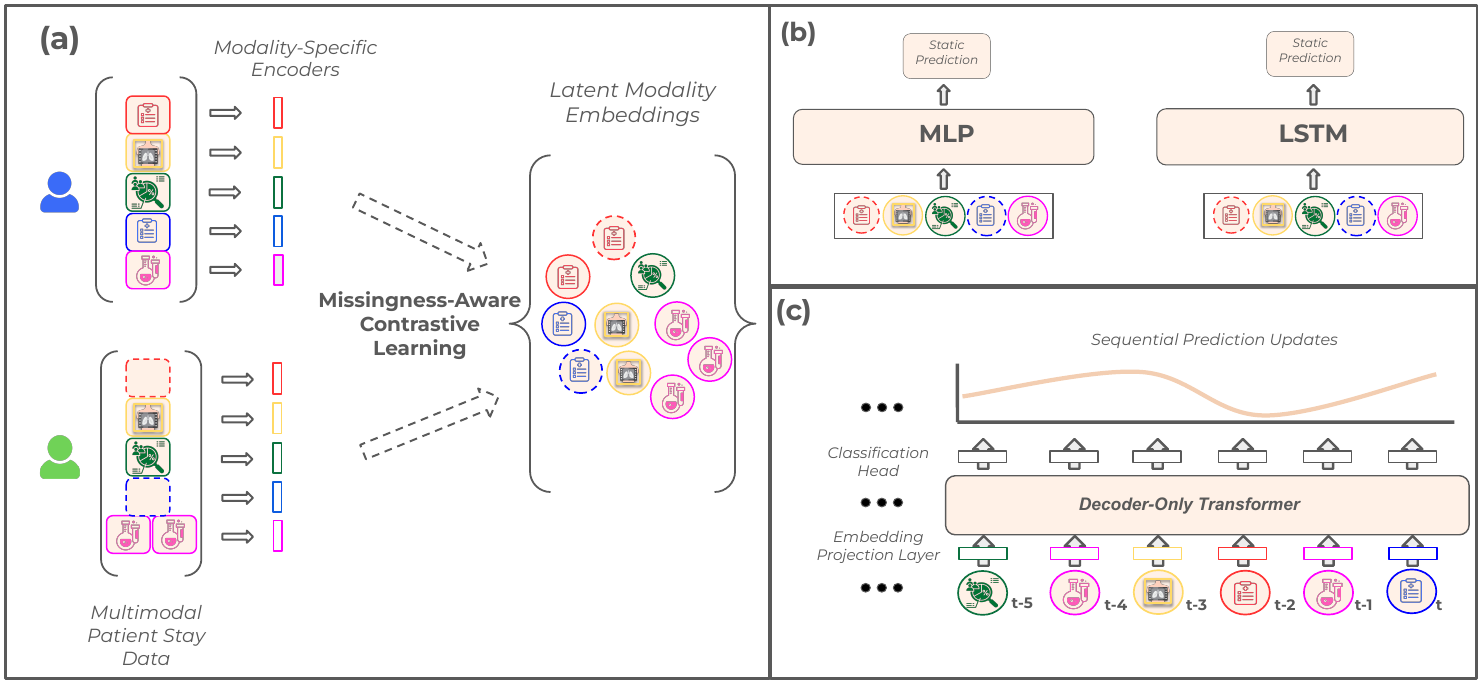}
	\caption{Overview of our proposed two-part missingness-aware framework. (a) We first develop a novel, missing modality-aware contrastive pretraining method, ensuring that the latent representations of the sparse data are aligned in a robustness manner. (b) We then re-purpose the machinery of autoregressive transformer-based models to model the sequence of contrastively pretrained embeddings for each patient stay, mimicking a realistic clinical diagnostic process and enabling more transparency into the sequential nature of decision making and updating.} 
	\label{fig:overview}
\end{figure*}

To summarize, our contributions are as follows:
\begin{itemize}
    \item{We introduce a novel, hybrid contrastive loss that learns explicit missingness tokens, aligning incomplete multimodal patient states into a global latent representation that benefits models trained on datasets with a high degree of missing modalities.}
	\item{We fuse the contrastive pretraining setup with transformer-based architectures to sequentially model patient trajectories, and show superior performance of sequential models over static baselines in the MIMIC-IV and eICU datasets.}
    \item{We provide mechanistic evidence that contrastive alignment prevents unstable safety failures under missing modality conditions, and trace the reasoning trail of sequential transformer-based architectures to highlight the transparency enabled by clinical sequence modeling.}
\end{itemize}

\section{Related Work}
\textbf{Building Multimodal Clinical Models Using MIMIC-IV and eICU}: With regards to the MIMIC dataset, \cite{Abuhamad2026}, \cite{sadanandan2026multimodaldeeplearningearly}, \cite{10.1145/3777577.3777712}, \cite{11095208}, and \cite{Lin2024} represent a subset of existing work that integrate demographics, lab results, chest X-Ray images, and other modalities to build predictive multimodal models.  In the context of applications in healthcare using the MIMIC dataset, \cite{zhu2025causaldebiasingmedicalmultimodal} covers the difficulties of debiasing multimodal representations on the MIMIC dataset, while \cite{wang2026revisitingperformanceclaimschest} presents a multimodal approach to an applied problem in handling contextual confounders in ML model evaluation, developing contextual models of future CXR labels using prior discharge summaries.  For the eICU dataset, existing work building multimodal clinical ML models includes \cite{Sheikhalishahi_2020}, which integrates demographic, laboratory measurements, and nurse bedside documentation, and \cite{WANG2024106672} integrates medical codes, multivariate time series, clinical text, and demographics.  Additionally, \cite{diagnostics11122242} and \cite{Rocheteau_2021} notably provide methods that generalize to both the MIMIC-eICU datasets.  Finally, \cite{golovanevsky2025picmepipelinecontrastivemodality} implements a similar contrastive pre-training pipeline to our study, utilizing modality-specific encoders in preparation for downstream task fine-tuning on the MIMIC-IV dataset to overcome performance degradation as modalities are removed from the evaluation set, although these approaches all make the assumption that all modalities are present in the training/evaluation sets.
\\
\textbf{Handling Missing Modalities}: In building predictive multimodal clinical ML models, existing approaches include dynamic Mixture of Experts models to address missing modalities on the fly \cite{wang2025moehealthmixtureexpertsframework}, transformer-based models to find correlations and representation alignment across potentially missing modalities in \cite{LIU2023104466},  \cite{tölle2025arbitrarydataimagesfusion}, and \cite{lee2023learningmissingmodalelectronic} exploring different modality fusion strategies to handle this missingness.  For more transparency in model prioritization, \cite{Yao_Yin_Cheung_Liu_Qin_2024} disentangles features specific to EHRs and general across EHRs and images in a multimodal framework, producing disease- and patient-specific weights from an attention layer to dynamically weight each modality during prediction.
In uncertainty quantification, \cite{gong2026embracingaleatoricuncertaintymedical}  explicitly quantifies uni-modal aleatoric uncertainty to address missing modalities, resulting in significant performance gains on tasks in the MIMIC and eICU datasets. 
Other approaches in existing work include reconstructing missing modalities at inference time to mitigate performance degradation when these modalities are absent post-training (\cite{10.1145/3746456}), handling missing modalities by anchoring everything into the strongest, richest modality, using a multimodal transformer to handle correlations within and between modalities (\cite{pmlr-v250-mordacq24a}), and dynamic mixture of experts models with comparative training in learning process to directly regularize the geometry of the loss landscape directly (\cite{li2025simmlmsimpleframeworkmultimodal}).  Additionally, \cite{10204754} and \cite{wang2023missingmodalityenabledmultimodalfusion} fuse shared features across modalities to enable modality-aware training and identify modality-specific features.  However, these approaches all operate on static models and do not use the temporal information in the datasets to build sequentially updated models.
\\
\textbf{Sequential Clinical Modeling}: Sequential modeling akin to our approach is explored in \cite{liventsev2024intensivecarebigsequence}, but this study is preliminary, as they neither interpret their sequence models nor address missing modalities in their clinical event streams.  Most similar to our language model decoder-based approach, \cite{Renc2024} tokenizes multimodal clinical event sequences to pretrain decoders for zero-shot modeling of patient trajectories.  They notably do not conduct any interpretability experiments in this work, which fundamentally limits trust and transparency, and do not address potentially "missingness" patterns in the multimodal MIMIC dataset they train on. \cite{Bornet2023.06.01.23290824} and \cite{9964038} also use transformers to encode medical concepts from multiple input modalities in MIMIC, although, once again, neither work attempts to interpret the model or its decision-making, limiting transparency, nor does it attempt to handle missing modalities in this process.  Finally, \cite{wornow2025contextcluesevaluatinglong} provides a comprehensive study comparing sub-quadratic State Space Models (SSMs) with transformer models for long-context modeling tasks on clinical datasets.  However, their focus remains on quantifying the benefits of SSM architectures for tasks with long sequence lengths and does not attempt to interpret their models or profile the impact of missing modalities during training/inference.
\\
\textbf{Interpretability in Clinical Models}: \cite{SHE202423} \cite{Soenksen2022} are examples of efforts to quantify modality-specific importances in clinical prediction tasks.  With regards to transformer-based sequence models specifically, \cite{info:doi/10.2196/74142} attempts to interpret reasoning language models via structured rationale elicitation in the prompt, but lacks a mechanistic analysis of failure modes and modality interactions.  Similarly, \cite{liao2025ehrr1reasoningenhancedfoundationallanguage} approaches the sequence modeling interpretation through the lens of generating valid reasoning chains, but once again does not attempt a mechanistic interpretation of the model.  Indeed, \cite{Qiu2025} reveals the limitations of reasoning-oriented methods, finding that many patient samples still result in factually incorrect steps in the reasoning chains, thereby motivating the need to mechanistically interpret the underlying model to prevent overreliance on medically invalid reasoning chains.

\section{Data and Experimental Setup}

\subsection{Data Preprocessing for Contrastive Pretraining}
In this study, we utilized two large-scale healthcare databases: MIMIC-IV (v1.0)  paired with MIMIC-CXR (v2.0.0) for chest radiographs, and the eICU Collaborative Research Database. The following data preprocessing pipeline was designed specifically to curate the contrastive pretraining datasets.
For details on modality-specific data preprocessing procedures, refer to Appendix section \ref{sec:data_preprocessing}.

\subsubsection{Missingness-Aware Stratification}
To ensure the contrastive encoders learn robust, missingness-aware representations, we implemented a custom hybrid stratified-splitting strategy based on \textit{modality availability combinations} rather than clinical labels. For MIMIC-IV, this tracked combinations of Time-Series, CXR, Discharge Notes, Radiology Notes, and Demographics; for eICU, it tracked Demographics, Diagnosis, Treatment, Medication, Lab, and APS (Acute Physiology Scores). 

We partitioned both pretraining datasets into training (70\%), validation (15\%), and testing (15\%) sets. To prevent data leakage between correlated ICU stays (e.g., a single patient readmitted multiple times), stratification was performed with the following approach:
\begin{enumerate}
    \item \textbf{Patient-Level Split:} Records associated with a known patient ID (\texttt{subject\_id} or \texttt{uniquepid}) were grouped. The patient's first recorded modality combination was used as their stratification stratum, and the split was performed across unique patients.
    \item \textbf{Orphan Split:} Records lacking a unifying patient ID were assigned a stratum based on their specific stay ID (\texttt{stay\_id} or \texttt{patientunitstayid}) and split independently.
\end{enumerate}
This explicit stratification ensures that the evaluation sets preserve the exact distribution of data sparsity found in the training set without violating patient isolation. For the MIMIC-IV dataset, this yielded 37,133 training stays, 6,110 validation stays, and 6,113 testing stays. For the eICU dataset, this yielded 137,153 training stays, 29,190 validation stays, and 29,348 testing stays.

\subsection{Contrastive Pretraining Architecture and Objective}
To learn robust latent representations, a standard approach is contrastive learning (\cite{chen2020simpleframeworkcontrastivelearning}), a self-supervised method that pushes representations of the same underlying sample (e.g., a patient) closer together in the latent space, while pushing representations of different samples apart.  Extensions of this contrastive setup for multimodal datasets include the One-Versus-Others (OvO) loss (used in \cite{golovanevsky2025picmepipelinecontrastivemodality}) and (\cite{Thapa2026}), which encourages each specific modality to align with the collective, averaged representation of all other modalities.  While these are highly effective for complete datasets, it implicitly assumes that all modalities are present for every sample. 

To address this, we propose a \textit{Masked Global Alignment} contrastive pre-training framework.  This framework extends the traditional multimodal contrastive objective by aligning both complete and incomplete patient data into a shared latent representation.  This contrastive setup enables modality-specific encoders to transform raw patient data into feature representations.  For full architectural details on the encoders, refer to Appendix section \ref{sec:architectures}. 

\subsubsection{Latent Imputation via Learnable Missing Tokens}
\label{sec:missing_tokens}
To handle missing data without discarding patient samples or zero-padding the latent space, we introduced a set of independently learnable modality tokens $\{\mathbf{t}_i\}_{i=1}^M$, where $\mathbf{t}_i \in \mathbb{R}^D$ and $M$ is the total number of modalities in the dataset.

Let $\mathcal{B}$ denote a mini-batch of $N$ patients.  We define the presence of modality $i$ for patient $k$ using an indicator mask $M_{i,k}$, which equals 1 if the modality is observed and 0 otherwise.  During the forward pass, if modality $i$ is present, the raw data $\mathbf{x}_{i,k}$ is passed through its respective encoder.  If the modality is missing, the encoder is bypassed, and the learnable token $\mathbf{t}_i$ is injected directly into the latent space:
\begin{equation}
\mathbf{h}_{i,k} = M_{i,k} \cdot \text{Encoder}_i(\mathbf{x}_{i,k}) + (1 - M_{i,k}) \cdot \mathbf{t}_i
\end{equation}
To simulate missingness and encourage robustness, we apply modality dropout during pre-training by randomly setting $M_{i,k} = 0$ with probability $p_{\text{drop}}=0.15$. These latent representations are subsequently $L_2$-normalized to yield $\mathbf{z}_{i,k} = \mathbf{h}_{i,k} / \|\mathbf{h}_{i,k}\|_2$, ensuring all projected modalities and missingness tokens reside on the same unit hypersphere.  The total number of available modalities for patient $k$ is given by $C_k = \sum_{j=1}^M M_{j,k}$.

\subsubsection{Unified Masked Centroid Alignment Loss}
At its core, contrastive learning optimizes a latent space by pulling a representation closer to its corresponding positive target (e.g., another clinical view of the same patient) while simultaneously pushing it away from a set of negative targets (e.g., clinical views of different patients). In our highly sparse multimodal setting, we treat a single observed modality as the base representation and the aggregated summary of the patient's remaining modalities as the positive target context.  To prevent the model from trivially aligning a modality to itself, we define a \textit{Complementary Centroid} $\mathbf{c}_{i,k}$ for each anchor modality $i$.  This vector represents the aggregated context of all \textit{other} available modalities for patient $k$:
\begin{equation}
\mathbf{c}_{i,k} = \frac{\sum_{j \neq i} M_{j,k} \mathbf{z}_{j,k}}{\left\| \sum_{j \neq i} M_{j,k} \mathbf{z}_{j,k} \right\|_2} 
\end{equation}
For patients where only a single modality is present, we define a fallback \textit{Stable Global Representation} $\bar{\mathbf{z}}_k$, computed as the normalized centroid of all active modalities:
\begin{equation}
\bar{\mathbf{z}}_k = \frac{\sum_{j=1}^M M_{j,k} \mathbf{z}_{j,k}}{\left\| \sum_{j=1}^M M_{j,k} \mathbf{z}_{j,k} \right\|_2} 
\end{equation}

For a specific modality $i$, we define the set of valid patients for contrastive alignment as $\mathcal{B}_i = \{k \in \mathcal{B} \mid M_{i,k} = 1 \text{ and } C_k \ge 2\}$.  This constraint ensures that a patient possesses at least two modalities, which is strictly required to construct a valid positive pair.  For a given embedding $\mathbf{z}_{i,k}$, the positive target is defined as the patient's own complementary centroid $\mathbf{c}_{i,k}$.  Conversely, the negative targets are drawn from the representations of all other patients $m \neq k$ within the mini-batch.To form these negative pairs, we primarily utilize the complementary centroid $\mathbf{c}_{i,m}$ of the negative patient $m$.  However, if patient $m$ possesses only modality $i$ (i.e., $C_m - M_{i,m} = 0$), computing a complementary centroid is impossible. In this highly sparse scenario, we fall back to using their global representation $\bar{\mathbf{z}}_m$.  We formalize the negative target $\mathbf{v}_{i,m}$ as:
\begin{equation}
\mathbf{v}_{i,m} = \begin{cases} 
\mathbf{c}_{i,m} & \text{if } C_m - M_{i,m} > 0 \\ 
\bar{\mathbf{z}}_m & \text{otherwise} 
\end{cases} 
\end{equation}

This fallback mechanism ensures that even patients with severe data sparsity (only one modality present) can still serve as informative negative examples without destabilizing the contrastive objective.

The contrastive loss for modality $i$ across the batch is computed using the InfoNCE objective with a learnable temperature parameter $\tau$:
\begin{equation}
\mathcal{L}_i = \frac{1}{|\mathcal{B}_i|} \sum_{k \in \mathcal{B}_i} -\log \frac{\exp(\mathbf{z}_{i,k} \cdot \mathbf{c}_{i,k} / \tau)}{\exp(\mathbf{z}_{i,k} \cdot \mathbf{c}_{i,k} / \tau) + \sum_{m \neq k} \exp(\mathbf{z}_{i,k} \cdot \mathbf{v}_{i,m} / \tau)} 
\end{equation}

This loss function penalizes the model if a modality embedding cannot correctly identify its own patient's multimodal context out of a crowd of competing negative contexts.  By minimizing this objective, we force the diverse, isolated modalities of a single patient to converge into a tightly coupled, modality-invariant region of the latent space.

\subsection{Data Preprocessing for Finetuning}
Because our study compares static ``bag-of-features'' models against our sequential transformer models, we generated two structural representations of the finetuning datasets.

\subsubsection{Static Baseline Preprocessing}
To establish fair comparisons against standard clinical ML architectures, we constructed static datasets where all temporal modalities within the observation window were collapsed into a single, fixed-size feature vector per stay.

\textbf{eICU Static Features:} We flattened the temporal dimension by aggregating features across the entire stay. Laboratory values (filtered to those with > 10,000 occurrences globally) were averaged. String-based modalities (\texttt{diagnosis}, \texttt{treatment}, \texttt{medication}) were parsed by concatenating all unique events per patient into a delimited string, converting them into binary dummy variables, and filtering out features present in fewer than 5\% of the cohort. 

\textbf{MIMIC-IV Static Features:} We selected the single most recent Anteroposterior (AP) Chest X-Ray acquired strictly between the ICU \texttt{intime} and \texttt{outtime}. Electronic Health Record (EHR) time-series were discretized into 1.0-hour bins, imputed using a carry-forward strategy, and flattened. Discharge and Radiology notes were concatenated into single documents per stay.

\subsubsection{Temporal Sequential Preprocessing}
To evaluate our sequential models, we restructured the data into chronologically ordered sequences of discrete events.

\textbf{eICU Sequential Timelines:} We extracted the exact temporal offset (in minutes from admission) for every individual clinical event. Demographics and Acute Physiology Scores (APS) were initialized at time offset $0.0$. Laboratory results, diagnoses, treatments, and medication administrations were assigned their respective recorded offsets (e.g., \texttt{labresultoffset}, \texttt{drugstartoffset}). All events for a given patient were then concatenated and strictly sorted by time.

\textbf{MIMIC-IV Sequential Timelines:} Creating the MIMIC timelines required unstacking the previously discretized 1-hour EHR bins. Using the patient's absolute \texttt{intime}, we mapped each preprocessed EHR bin back to a discrete timestamp (\texttt{intime} + elapsed hours). These unstacked EHR events were then merged with the absolute timestamps of the CXR acquisition (\texttt{StudyDateTime}) and clinical notes (\texttt{charttime}).

\textbf{Autoregressive Sequential Data:} Once all multimodal events for a patient were chronologically sorted, we algorithmically injected a special \texttt{[PREDICT]} token immediately following every individual event in the timeline. This forces the model model to autoregressively output an updated belief state and target prediction at every time step as new evidence arrives. 

\subsection{Fine-Tuning Architectures and Training Setup}
To rigorously evaluate the benefits of our sequential modeling approach, we compared our transformer-based models against static baseline models. While the underlying modality encoders and pre-trained contrastive weights were shared, the fusion mechanisms, classification heads, and optimization strategies diverged based on the temporal nature of the architecture.

\subsubsection{Static Architecture: Encoders and Late Fusion}
For our static baselines and static contrastively fine-tuned models, all multimodal events within the patient's observation window were aggregated and passed through modality-specific encoders to produce a set of fixed-size embeddings $\mathbf{z}_{i} \in \mathbb{R}^{256}$. These were then used to train downstream static architectures.
To handle inherently missing clinical modalities without discarding patient samples, we use the explicitly learnable missing tokens ($\mathbf{t}_i$) introduced during pretraining (Section \ref{sec:missing_tokens}). We evaluated two static late-fusion architectures:

\textbf{Multi-Layer Perceptron}: Modality embeddings are concatenated into a single, wide vector and passed through a multi-layer perceptron (MLP) ($\mathbb{R}^{256 \times M} \to \mathbb{R}^{512}$).

\textbf{LSTM}: The set of modalities is treated as an unordered sequence and passed through a standard LSTM , which outputs a final, integrated hidden state for downstream classification.

We use MLP as a naive static baseline, and LSTM is used to implicitly model sequence dependency between modalities in a static setting.

\subsubsection{Sequential Architecture}
To model the temporal evolution of clinical reasoning, we finetuned six different pre-trained LLM decoder backbones: Llama-3-8B \cite{grattafiori2024llama3herdmodels}, Mistral-7B-v0.1 \cite{jiang2023mistral7b}, DeepSeek-LLM-7B \cite{deepseekai2024deepseekllmscalingopensource}, Phi-3-mini-4k-instruct \cite{abdin2024phi3technicalreporthighly}, BioMistral-7B \cite{labrak2024biomistralcollectionopensourcepretrained}, and Meditron-7B \cite{chen2023meditron70bscalingmedicalpretraining}.

To model the temporal evolution of clinical reasoning, we fine-tuned six different pre-trained autoregressive large language models (LLMs): Llama-3-8B \cite{grattafiori2024llama3herdmodels}, Mistral-7B-v0.1 \cite{jiang2023mistral7b}, DeepSeek-LLM-7B \cite{deepseekai2024deepseekllmscalingopensource}, Phi-3-mini-4k-instruct \cite{abdin2024phi3technicalreporthighly}, BioMistral-7B \cite{labrak2024biomistralcollectionopensourcepretrained}, and Meditron-7B \cite{chen2023meditron70bscalingmedicalpretraining}. During training, the pre-trained modality encoders are strictly frozen and act solely as feature extractors. Each modality embedding $\mathbf{z}_{i} \in \mathbb{R}^{256}$ is mapped to the LLM's native hidden dimension $d_{llm}$ (e.g., 4096 for Llama-3) via a learnable projection block. To enable dynamic, autoregressive belief updating, a globally learnable $\texttt{[PREDICT]} \in \mathbb{R}^{256}$ embedding is injected into the chronological sequence immediately following each observed clinical event.
For full architectural and training details, refer to \ref{sec:architectures}.

\subsection{Tasks and Evaluation Metrics}
To evaluate the generalization and robustness of our representations, we benchmarked the architectures across three distinct clinical tasks.

\textbf{Pretraining Metrics}: We use \textit{Cross-Modal Retrieval} (Recall@K) to query the direct semantic alignment between different data modality embeddings. Specifically, it measures the frequency with which a patient's embedding from one modality (e.g., a chest X-ray) correctly retrieves their corresponding embedding from another modality (e.g., a clinical note) out of a batch of negative patient candidates. High retrieval accuracy indicates a tightly coupled latent space where distinct modalities share the same clinical grounding.
We also use \textit{Silhouette Scores} to calculate how similar an embedding is to its own target cluster (cohesion) compared to other clusters (separation). In our setup, a low silhouette score confirms that the contrastive objective successfully groups the different modality embeddings together.

\textbf{Finetuning Tasks and Metrics}
\textbf{Phenotyping}: A multi-label classification task predicting the presence of 25 distinct acute and chronic conditions (e.g., acute myocardial infarction, sepsis) during a patient's stay using the MIMIC-IV database.

\textbf{In-Hospital Mortality}: A binary classification task predicting whether a patient will expire before hospital discharge. To test cross-system generalizability, this task was evaluated independently on both the MIMIC-IV and eICU databases.

\textbf{Length of Stay (LOS)}: Formulated as a continuous regression task on the eICU database, models were trained to predict the total hours a patient would remain in the ICU. To ensure clinical relevance and stabilize the training objective, the cohort was filtered to exclude extreme outliers, restricting the prediction window to stays between 12 hours and 30 days.

\textbf{Classification Metrics}: To account for the inherent class imbalances in clinical datasets, we evaluated classification performance using the Area Under the Receiver Operating Characteristic curve (AUROC) and the Area Under the Precision-Recall Curve (AUPRC). For the multi-label phenotyping task, these metrics were macro-averaged across all 25 condition classes to weight each condition equally regardless of its prevalence. 

\textbf{Regression Metrics}: Predictive accuracy was measured using Mean Squared Error (MSE) and Mean Absolute Error (MAE) to capture the absolute deviation in predicted hours. To assess how well the models captured the relative trajectory and rank-order of patient stays, we additionally computed the Pearson correlation coefficient ($r$) and the Spearman rank-order correlation coefficient ($\rho$) for evaluating the alignment between the predicted and actual lengths of stay.

\subsection{Mechanistic Interpretation Analysis}
\begin{figure*}
	\centering
	\includegraphics[width=0.75\linewidth]{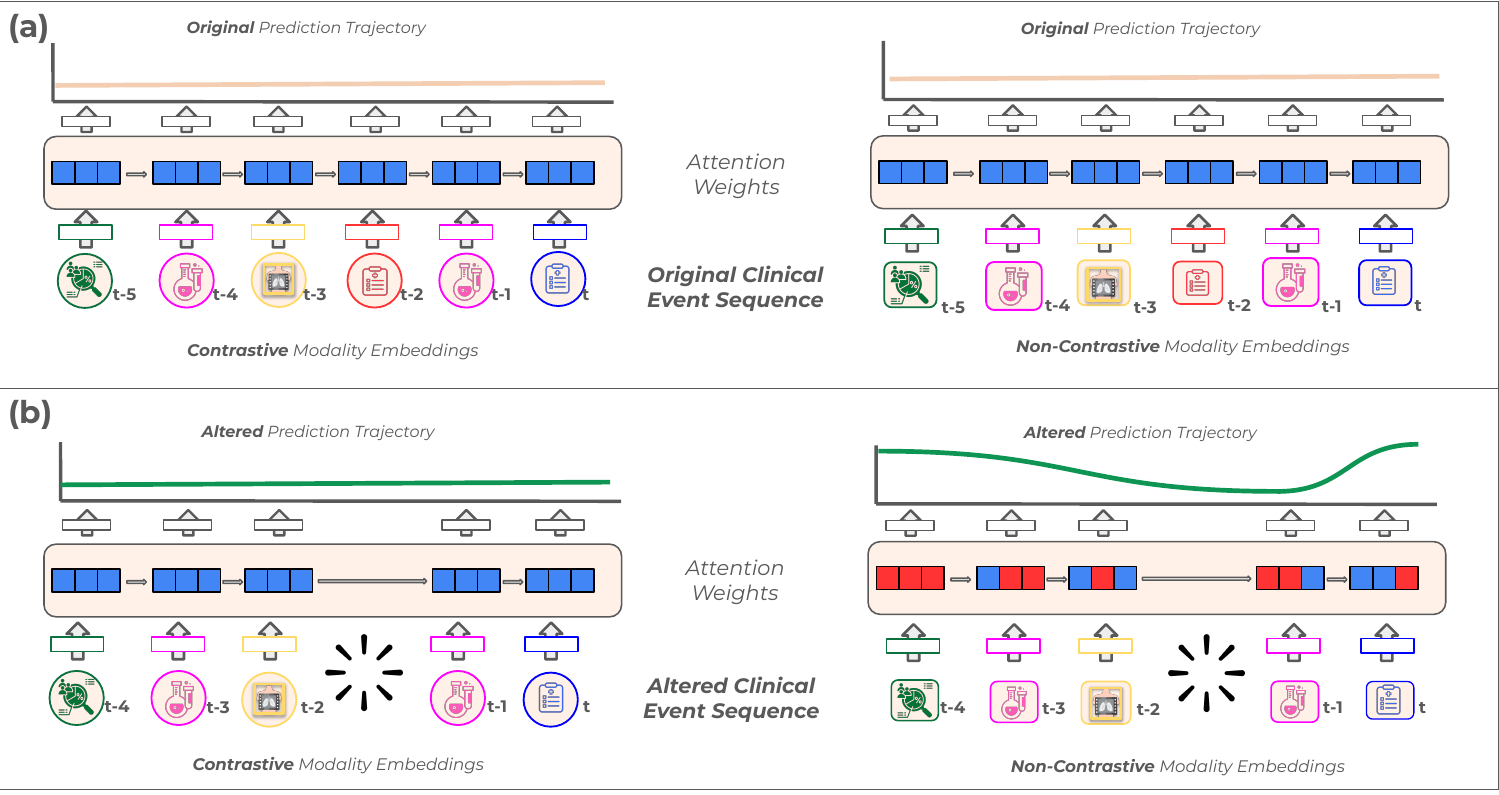}
	\caption{Overview of our mechanistic interpretability setup: (a) We first aggregate the attention weights for a given patient stay between the contrastive/non-contrastive sequential models. (b) We then remove a given modality (The red notes in this example) and analyze the impact on both the prediction trajectory and the attention weights between the contrastive/non-contrastive models. } 
	\label{fig:mech_interp_overview}
\end{figure*}
To understand exactly how missing modalities alter the internal reasoning of our sequential models, we conducted a mechanistic analysis of the transformer's attention routing. Specifically, we traced the source of predictive divergence by extracting the $N \times N$ causal self-attention matrices from the final layer of the transformer decoders, where $N$ represents the total sequence length of the patient's multimodal clinical event sequence.To obtain a holistic view of the model's temporal reasoning, we averaged the attention weights across all attention heads within this final layer . This resulting heatmap allowed us to map exactly which historical tokens the model most heavily relies upon when calculating its final diagnostic state. 

\section{Results}
\subsection{Contrastive Masked Global Alignment Pre-training Semantically Integrates Patients with Missing Modalities}
We first evaluate the benefits of our proposed Masked Global contrastive pre-training with regard to aligning modality-specific representations within the shared latent space.  To ensure that the embeddings used for downstream tasks are both well-integrated and semantically cohesive, we assess the latent geometry using two standard representation learning metrics: cross-modal retrieval and the silhouette score.

For experiments on modality dropout ablations, refer to \ref{fig:mimic_ablations} and \ref{fig:eicu_ablations}.

% \begin{figure*}
% 	\centering
% 	\includegraphics[width=0.6\linewidth]{img/mimic_pretrained_embeddings.pdf}
% 	\caption{PCA, t-SNE, and UMAP visualizations of the pretrained MIMIC-IV embeddings in the latent space. Silhouette scores are computed and displayed.} 
% 	\label{fig:mimic_latents}
% \end{figure*}

\begin{figure*}[htbp]
    \centering
    \begin{subfigure}{0.48\textwidth}
        \centering
        \includegraphics[width=\linewidth]{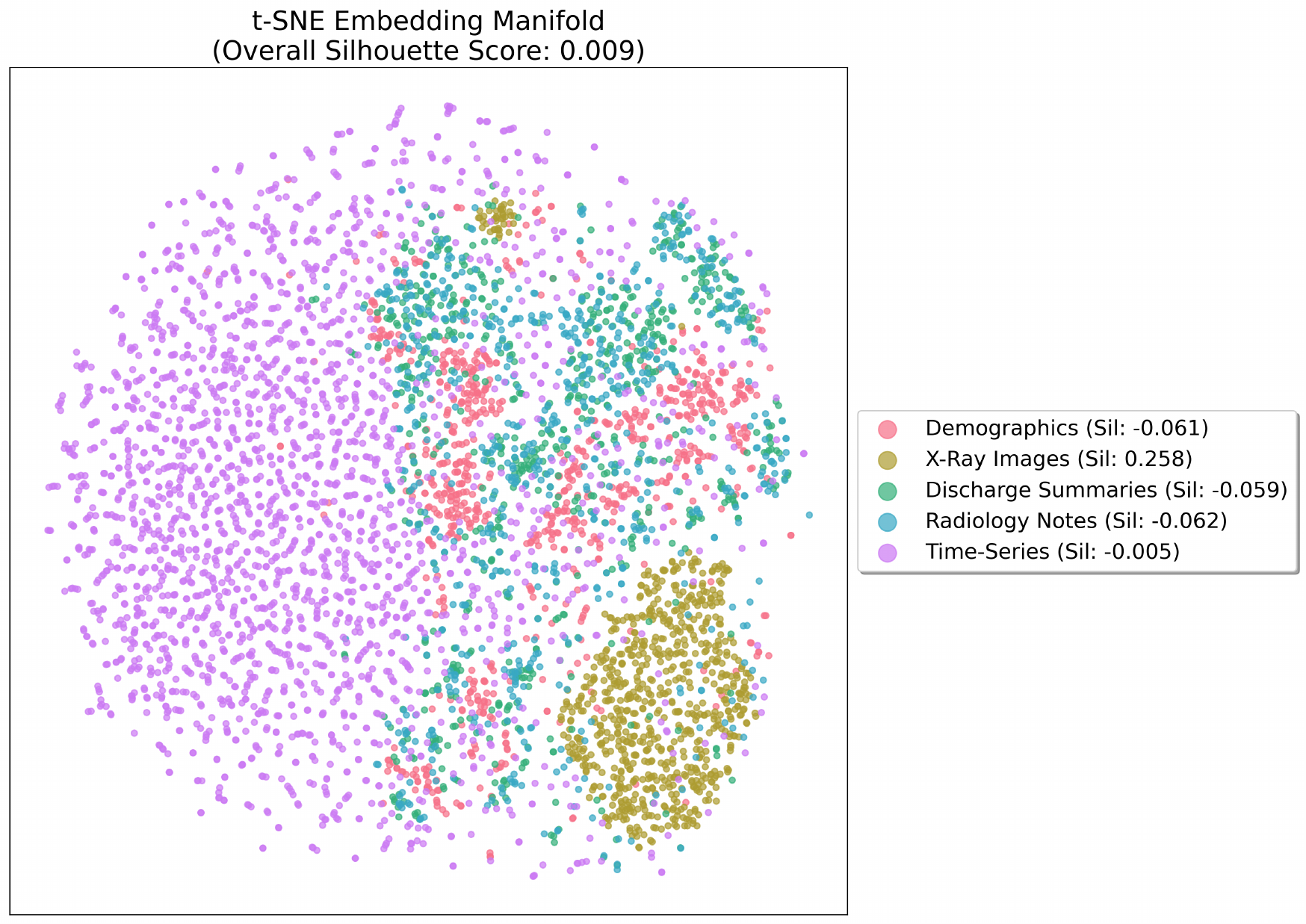}
        \caption{MIMIC-IV}
        \label{fig:mimic_tsne}
    \end{subfigure}\hfill
    \begin{subfigure}{0.48\textwidth}
        \centering
        \includegraphics[width=\linewidth]{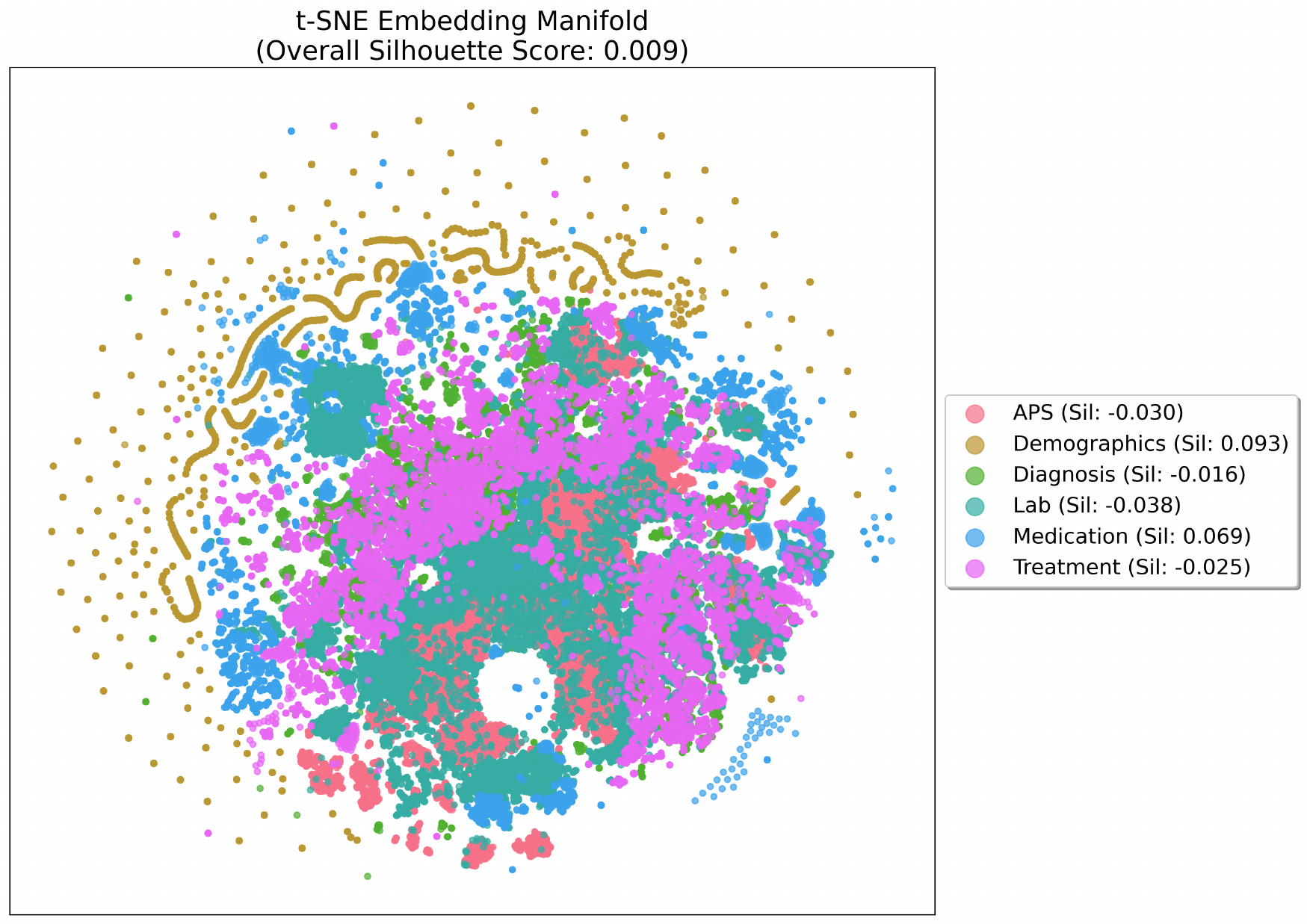}
        \caption{eICU}
        \label{fig:eicu_tsne}
    \end{subfigure}
    \caption{t-SNE visualizations of the pretrained MIMIC-IV (left) and eICU (right) embeddings in the latent space on the test set. Silhouette scores for each modality are computed and displayed within the legends. Through qualitative visual inspection of the latent manifolds and quantitative analysis of the silhouette scores, we determine that the modalities are well-integrated and harmonized in the shared latent space.} 
    \label{fig:mimic_eicu_latents}
\end{figure*}

We observe that the contrastive pre-training process yields well-integrated modality embeddings in the latent space for both the MIMIC-IV and eICU datasets in \ref{fig:mimic_eicu_latents}, corroborated by low silhouette scores of 0.009 overall on the latent modality representations, as well as low per-modality scores.  We provide additional visualizations of the contrastively pretrained latent embeddings in \ref{fig:combined_latents}.

To investigate semantic alignment with respect to the embeddings' actual context, we turn to retrieval metrics, as shown in \ref{tab:combined_retrieval}.  We observe that the contrastively pretrained embeddings are semantically aligned per our retrieval metrics, with large improvements over the random baseline, defined as a random retrieval from the total possible pool of patient modalities in the evaluation set.

\begin{table}[h]
\centering
\caption{Cross-modal test retrieval metrics for encoders contrastively pretrained on the MIMIC-IV and eICU datasets. We observe significant improvements over random baselines in our dataset, computed as a random retrieval from the corresponding test sets.}
\label{tab:combined_retrieval}
\begin{tabular}{llccc}
\toprule
\textbf{Dataset} & \textbf{Metric} & \textbf{Random Baseline} & \textbf{Masked Global} & \textbf{Improvement Factor} \\
\midrule
\textbf{MIMIC-IV}
& R@1  & 0.14\% & 5.18\% & \textbf{37.00$\times$} \\
& R@5  & 0.70\% & 5.69\% & \textbf{8.13$\times$} \\
& R@10 & 1.40\% & 8.41\% & \textbf{6.01$\times$} \\
\midrule
\textbf{eICU}
& R@1  & 0.003\% & 1.80\% & \textbf{600.00$\times$} \\
& R@5  & 0.017\% & 1.92\% & \textbf{112.94$\times$} \\
& R@10 & 0.034\% & 3.02\% & \textbf{89.12$\times$} \\
\bottomrule
\end{tabular}
\end{table}

\subsection{Sequential Modeling Provides Performance Boosts Over Static Architectures}

We next isolate the practical benefits of framing EHR data longitudinally rather than as static, aggregated snapshots.  Across both the MIMIC-IV and eICU datasets, transformer-based sequential models universally outperformed static multi-layer perceptrons (MLPs) and recurrent baselines (LSTMs).

This trend is particularly evident in MIMIC-IV phenotyping (Table \ref{table:mimic_static_and_seq_models_unified}). The static MLP initialized from scratch achieved a Spearman Correlation of 0.6002, whereas sequential architectures like Meta-Llama-3-8B achieved a coefficient of 0.7537 on encoders initialized from scratch. A similar performance trend is evident in the eICU Length of Stay (LoS) regression task (Table \ref{table:static_and_lm_models_unified}), where the static MLP maxed out at an AUPRC of 0.2451, while contrastively trained sequential models reached up to 0.5783.

While sequential architectures proved universally beneficial, the impact of our contrastive pre-training strategy was highly dataset-dependent, potentially revealing a more nuanced interaction between representation learning and modality missingness.

For the highly sparse MIMIC-IV dataset (where $86\%$ of the data are stays with only one out of five modalities present), combining contrastive pre-training with sequence modeling yielded dramatic performance leaps.  When initialized with contrastively learned embeddings, the Phi-3-Mini-4K-Instruct model yielded a $0.7652$ AUROC.  More impressively, DeepSeek-LLM-7B-Base saw its Mortality AUROC jump from $0.5720$ (scratch) to $0.8409$.  In this high-missingness dataset, forcing the model to align disparate modalities (e.g., projecting a missing lab result into the same space as an existing clinical note) during contrastive pre-training largely shields the downstream sequential model from missing-data noise, operating instead on a stable, modality-agnostic manifold.

Conversely, for the eICU tasks (where over $50\%$ of the data are patients with all six modalities present), the benefits of contrastive pre-training are less obvious.  Sequential models trained from scratch consistently outperformed their contrastively pretrained counterparts on both the Mortality and LoS tasks.  For example, Mistral-7B-v0.1 achieved the highest eICU Mortality AUROC of 0.9223 from scratch, which slightly degraded to 0.9107 under the contrastive setup.  We attribute this inversion to varying rates of modality "missingness" and the inherent "ceiling effect" of the tasks.  ICU patient stays have far fewer missing modalities overall (see Figures \ref{fig:eICU_pretraining_stats}-\ref{fig:eICU_los_data} for the full exploratory data analysis).  When clinical data is dense and highly complete, the inherent difficulty of the diagnostic task drops significantly, as evidenced by the baseline MLP from scratch achieving a highly competitive 0.9000 AUROC for eICU mortality. We conclude that sequential models yield performance gains across all tasks, and that our contrastive pre-training setup yields further gains on datasets with a higher rate of modality missingness. For additional calibration metrics of the best performing models, refer to \ref{table:mimic_calibration_metrics} and \ref{table:eicu_calibration_metrics}. For a detailed analysis on performance over different event sequence lengths, refer to \ref{fig:eicu_mortality_trajectory}-\ref{fig:MIMIC_phenotyping_base_trajectory}.

\begin{table}[htbp]
    \centering
    \caption{Performance of baseline static models and sequential transformer-based models on Mortality and Phenotyping tasks on the MIMIC-IV dataset. Best results for each metric are bolded, and we observe that the sequential models trained on contrastively learned embeddings consistently yield the highest performance across both classification tasks.}
    \resizebox{\textwidth}{!}{%
        \begin{tabular}{llcccc}
            \toprule
            & & \multicolumn{2}{c}{\textbf{Mortality}} & \multicolumn{2}{c}{\textbf{Phenotyping}} \\
            \cmidrule(lr){3-4} \cmidrule(lr){5-6}
            \textbf{Architecture} & \textbf{Initialization} & \textbf{AUROC} ($\uparrow$) & \textbf{AUPRC} ($\uparrow$) & \textbf{AUROC} ($\uparrow$) & \textbf{AUPRC} ($\uparrow$) \\
            \midrule
            Multi-layer Perceptron & Scratch & 0.6703 & 0.2189 & 0.5541 & 0.2211 \\
            LSTM & Scratch & 0.5575 & 0.1334 & 0.6880 & 0.3453 \\
            Multi-layer Perceptron & Contrastive & 0.4645 & 0.1501 & 0.5644 & 0.2451 \\
            LSTM & Contrastive & 0.5760 & 0.1653 & 0.5469 & 0.2410 \\
            \midrule
            BioMistral-7B & Scratch & 0.6364 & 0.2350 & 0.4881 & 0.2430 \\
            DeepSeek-LLM-7B-Base & Scratch & 0.5720 & 0.1741 & 0.4654 & 0.2098 \\
            Meditron-7B & Scratch & 0.5852 & 0.1880 & 0.5430 & 0.2803 \\
            Meta-Llama-3-8B & Scratch & 0.6477 & 0.2620 & 0.5098 & 0.2709 \\
            Phi-3-Mini-4K-Instruct & Scratch & 0.2803 & 0.1272 & 0.5192 & 0.2658 \\
            Mistral-7B-v0.1 & Scratch & 0.5985 & 0.2456 & 0.5159 & 0.2788 \\
            BioMistral-7B & Contrastive & 0.7500 & \textbf{0.4971} & \textbf{0.7362} & \textbf{0.5783} \\
            DeepSeek-LLM-7B-Base & Contrastive & \textbf{0.8409} & 0.3900 & 0.7162 & 0.4850 \\
            Meditron-7B & Contrastive & 0.6913 & 0.2147 & 0.7104 & 0.5009 \\
            Meta-Llama-3-8B & Contrastive & 0.7841 & 0.4553 & 0.7180 & 0.5148 \\
            Phi-3-Mini-4K-Instruct & Contrastive & 0.7652 & 0.3699 & 0.7270 & 0.5320 \\
            Mistral-7B-v0.1 & Contrastive & 0.6818 & 0.4358 & 0.7266 & 0.5203 \\
            \bottomrule
        \end{tabular}%
    }
    \label{table:mimic_static_and_seq_models_unified}
\end{table}

\begin{table}[htbp]
    \centering
    \caption{Performance of baseline static models and sequential transformer-based models on Mortality and Length of Stay tasks on the eICU dataset. Best results for each metric are bolded, and we observe that the sequential models with encoders trained from scratch consistently yields the highest performance across both regression and classification tasks.}
    \resizebox{\textwidth}{!}{%
        \begin{tabular}{llcccccc}
            \toprule
            & & \multicolumn{2}{c}{\textbf{Mortality}} & \multicolumn{4}{c}{\textbf{Length of Stay}} \\
            \cmidrule(lr){3-4} \cmidrule(lr){5-8}
            \textbf{Architecture} & \textbf{Initialization} & \textbf{AUROC} ($\uparrow$) & \textbf{AUPRC} ($\uparrow$) & \textbf{MSE} ($\downarrow$) & \textbf{MAE} ($\downarrow$) & \textbf{Pearson} ($\uparrow$) & \textbf{Spearman} ($\uparrow$) \\
            \midrule
            Multi-layer Perceptron & Scratch & 0.9000 & 0.5416 & 13356.1 & 81.97 & 0.5802 & 0.6002 \\
            LSTM & Scratch & 0.9000 & 0.5384 & 14799.3 & 84.68 & 0.5309 & 0.6009 \\
            Multi-layer Perceptron & Contrastive & 0.9006 & 0.5391 & 13331.2 & 81.88 & 0.5810 & 0.6016 \\
            LSTM & Contrastive & 0.8960 & 0.5385 & 14233.0 & 83.64 & 0.5470 & 0.6105 \\
            \midrule
            BioMistral-7B & Scratch & 0.9218 & 0.6411 & 9333.5 & 65.04 & 0.7333 & \textbf{0.7582} \\
            DeepSeek-LLM-7B-Base & Scratch & 0.9190 & \textbf{0.6453} & 9112.9 & 62.70 & 0.7399 & 0.7528 \\
            Meditron-7B & Scratch & 0.9207 & 0.6221 & 9229.9 & 66.66 & 0.7370 & 0.7513 \\
            Meta-Llama-3-8B & Scratch & 0.9213 & 0.6336 & \textbf{9089.2} & 62.98 & \textbf{0.7406} & 0.7537 \\
            Phi-3-Mini-4K-Instruct & Scratch & 0.9154 & 0.6218 & 9101.2 & \textbf{62.64} & 0.7403 & 0.7552 \\
            Mistral-7B-v0.1 & Scratch & \textbf{0.9223} & 0.6359 & 9145.0 & 65.08 & 0.7388 & 0.7543 \\
            BioMistral-7B & Contrastive & 0.8949 & 0.5717 & 9420.0 & 64.83 & 0.7299 & 0.7498 \\
            DeepSeek-LLM-7B-Base & Contrastive & 0.9053 & 0.6016 & 9232.2 & 63.80 & 0.7357 & 0.7531 \\
            Meditron-7B & Contrastive & 0.9116 & 0.6088 & 9426.8 & 66.72 & 0.7290 & 0.7493 \\
            Meta-Llama-3-8B & Contrastive & 0.9082 & 0.5977 & 9377.1 & 66.70 & 0.7312 & 0.7497 \\
            Phi-3-Mini-4K-Instruct & Contrastive & 0.9089 & 0.6045 & 9322.3 & 64.74 & 0.7327 & 0.7516 \\
            Mistral-7B-v0.1 & Contrastive & 0.9107 & 0.6020 & 9256.3 & 66.01 & 0.7352 & 0.7488 \\
            \bottomrule
        \end{tabular}%
    }
    \label{table:static_and_lm_models_unified}
\end{table}

\subsection{Interpreting Contrastive Sequential Models Reveal Critical Safety Considerations in Handling Missing Modalities}
\begin{figure*}[t]
\centering
\begin{subfigure}{0.49\textwidth}
\centering
\includegraphics[width=\linewidth]{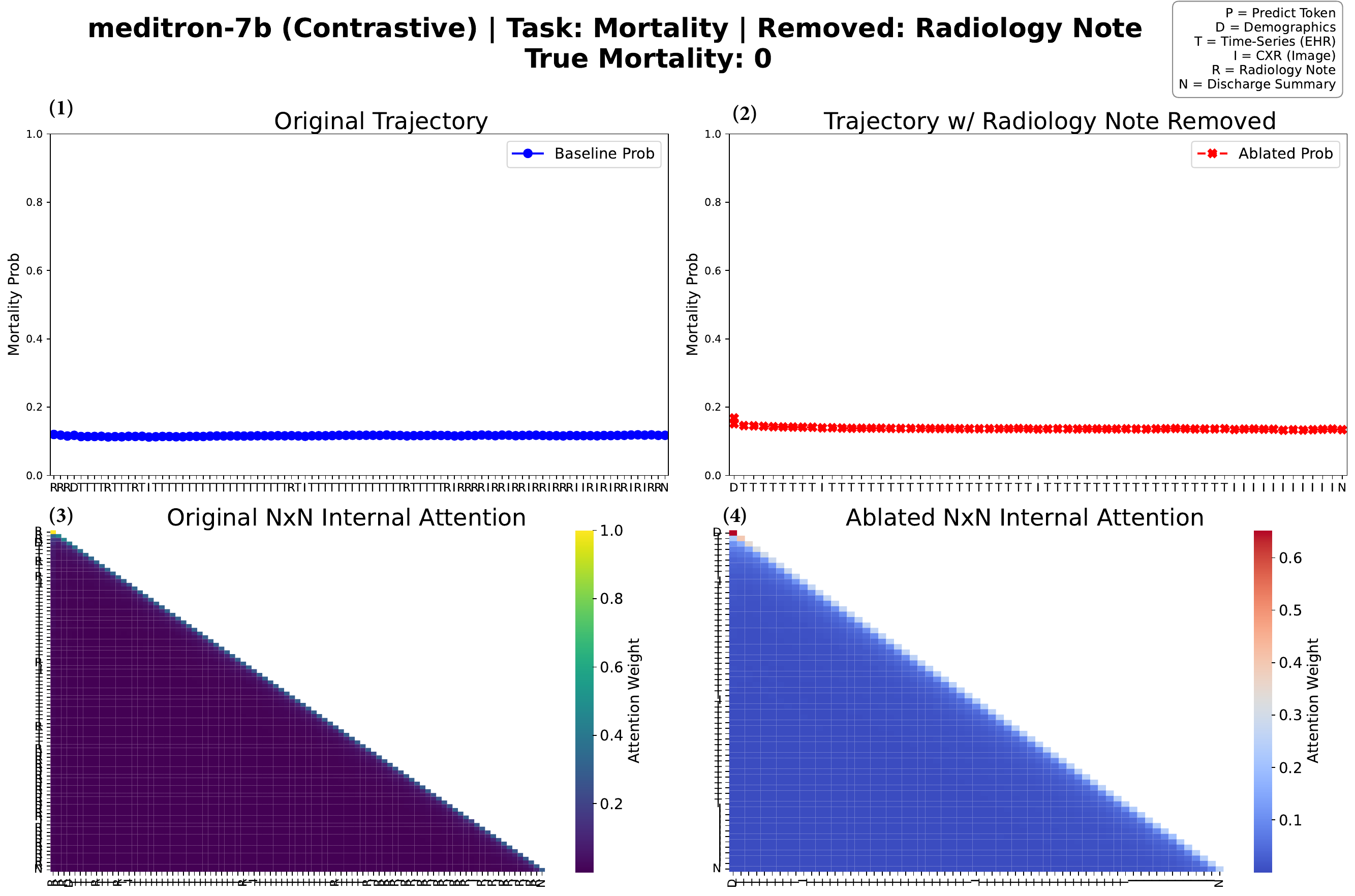}
\caption{Sequential architecture finetuned on contrastively learned embeddings.}\label{fig:contrastive_bias_main}\end{subfigure}
\hfill 
\begin{subfigure}{0.5\textwidth}\centering\includegraphics[width=\linewidth]{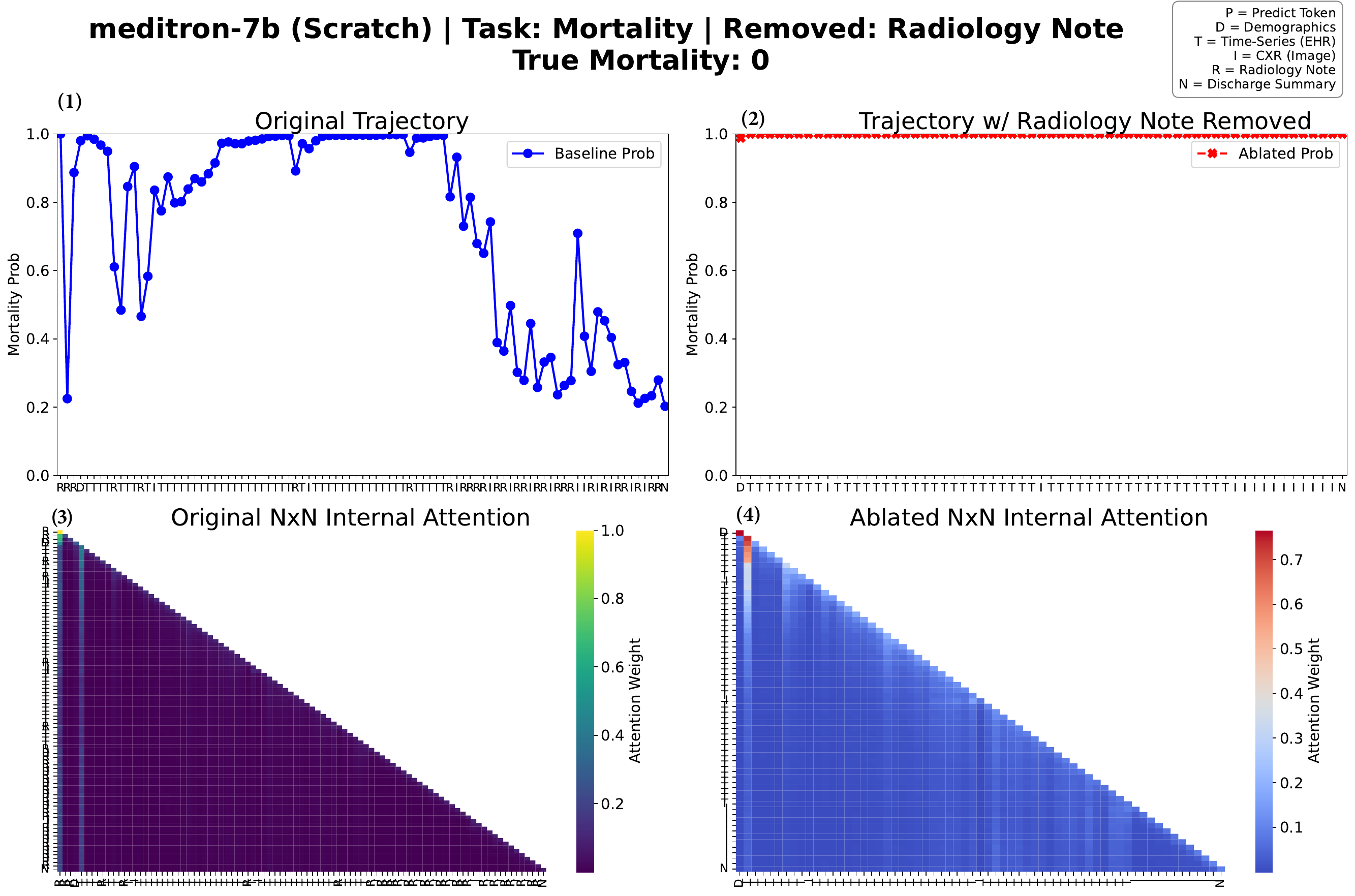}\caption{Sequential architecture finetuned on non-contrastive embeddings.}\label{fig:scratch_bias}\end{subfigure}
\caption{We experimentally remove the radiology note embeddings associated with a given MIMIC-IV stay. For the contrastively fine-tuned model, this does not result in a change in the prediction trajectory [\textbf{(a.1)} to \textbf{(a.2)}], nor divergent behavior in the internal attention [\textbf{(a.3) to \textbf{(a.4)}}]. Removing this modality changes the diagnosis incorrectly for the non-contrastive model, however [\textbf{(b.1)} to \textbf{(b.2)}].  Additionally, attention patterns reveal that the non-contrastive model previously gave high attention on the first radiology note throughout the sequence [\textbf{(b.3)} to \textbf{(b.4)}], and its removal may be the causal basis for the new, incorrect prediction.}
\label{fig:reliance}
\end{figure*}

Finally, we interpret the reasoning process of the sequential models under experimentally sparse patient stay trajectories.
We observed that the sequential model fine-tuned on standard, non-contrastive embeddings exhibits a collapse in its attention routing for certain patient stay trajectories.  We examine a stay in the MIMIC-IV dataset (Figure \ref{fig:reliance}), and observe that removing the radiology note modality embeddings does not appear to disrupt the sequential predictions on the sequential model fine-tuned on contrastively pre-trained embeddings.  However, for the model trained on non-contrastive embeddings, removing the radiology note embeddings causes the predictions to become incorrect.  We note that the heatmaps indicate increased attention on the first radiology note embedding, which is forcibly removed in the ablated sequence.  As such, the model must rely on the other modalities, which have not been aligned in the latent space via contrastive pre-training.  We posit that this behavior is directly mitigated by our Masked Global Alignment objective, which forces the encoders to learn robust, modality-invariant representations during pre-training.  Because the latent space is already geometrically aligned to handle missingness, the contrastive sequential decoder does not suffer in the same way as the non-contrastive model; instead, it safely routes attention through the available clinical history, preserving predictive accuracy in highly sparse environments.

\begin{figure*}[t]
    \centering
    % Subfigure A
    \begin{subfigure}{0.49\textwidth}
        \centering
        \includegraphics[width=\linewidth]{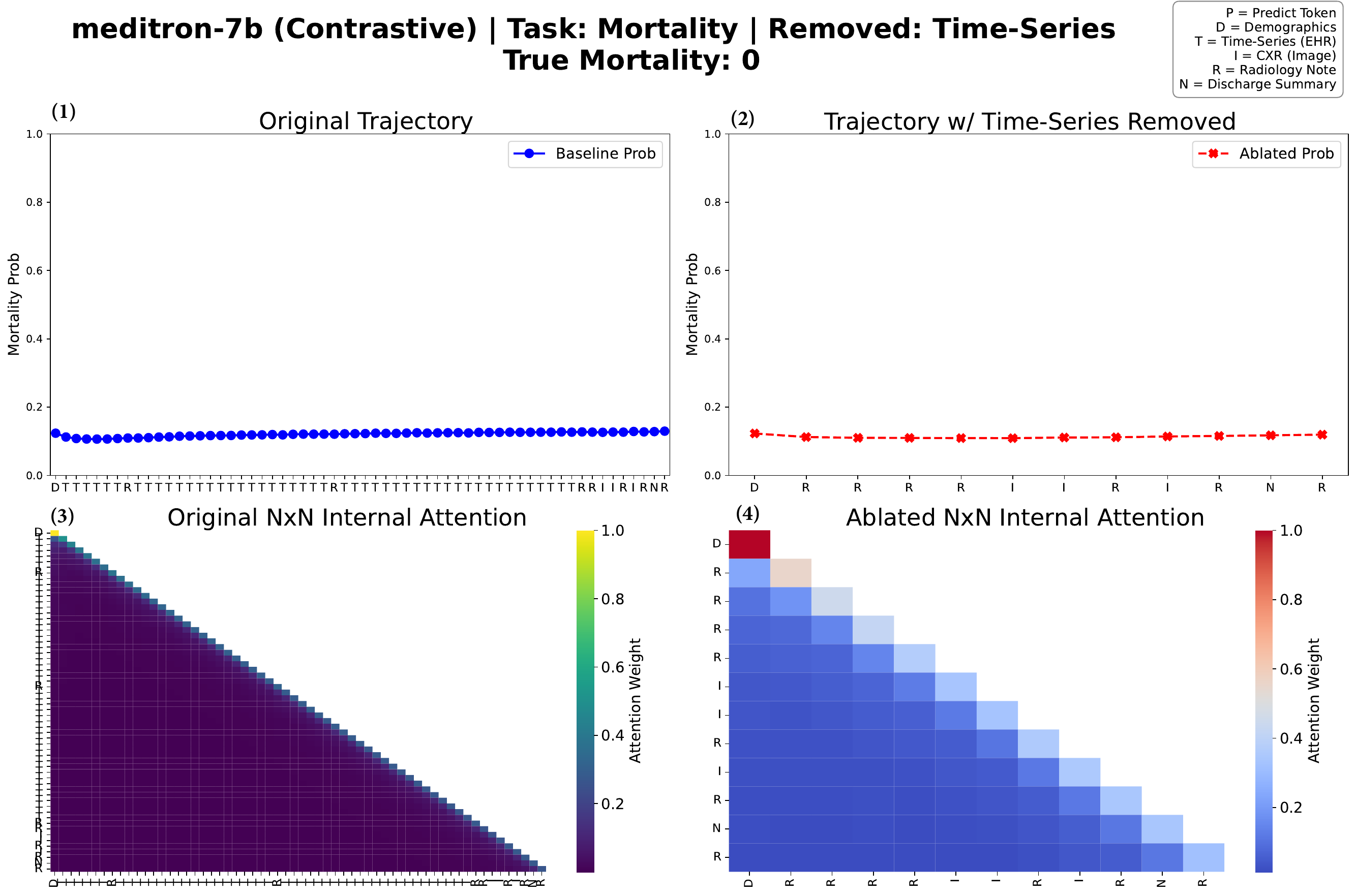}
        \caption{Sequential architecture finetuned on contrastively learned embeddings.} 
        \label{fig:contrastive_bias}
    \end{subfigure}\hfill
    % Subfigure B
    \begin{subfigure}{0.49\textwidth}
        \centering
        \includegraphics[width=\linewidth]{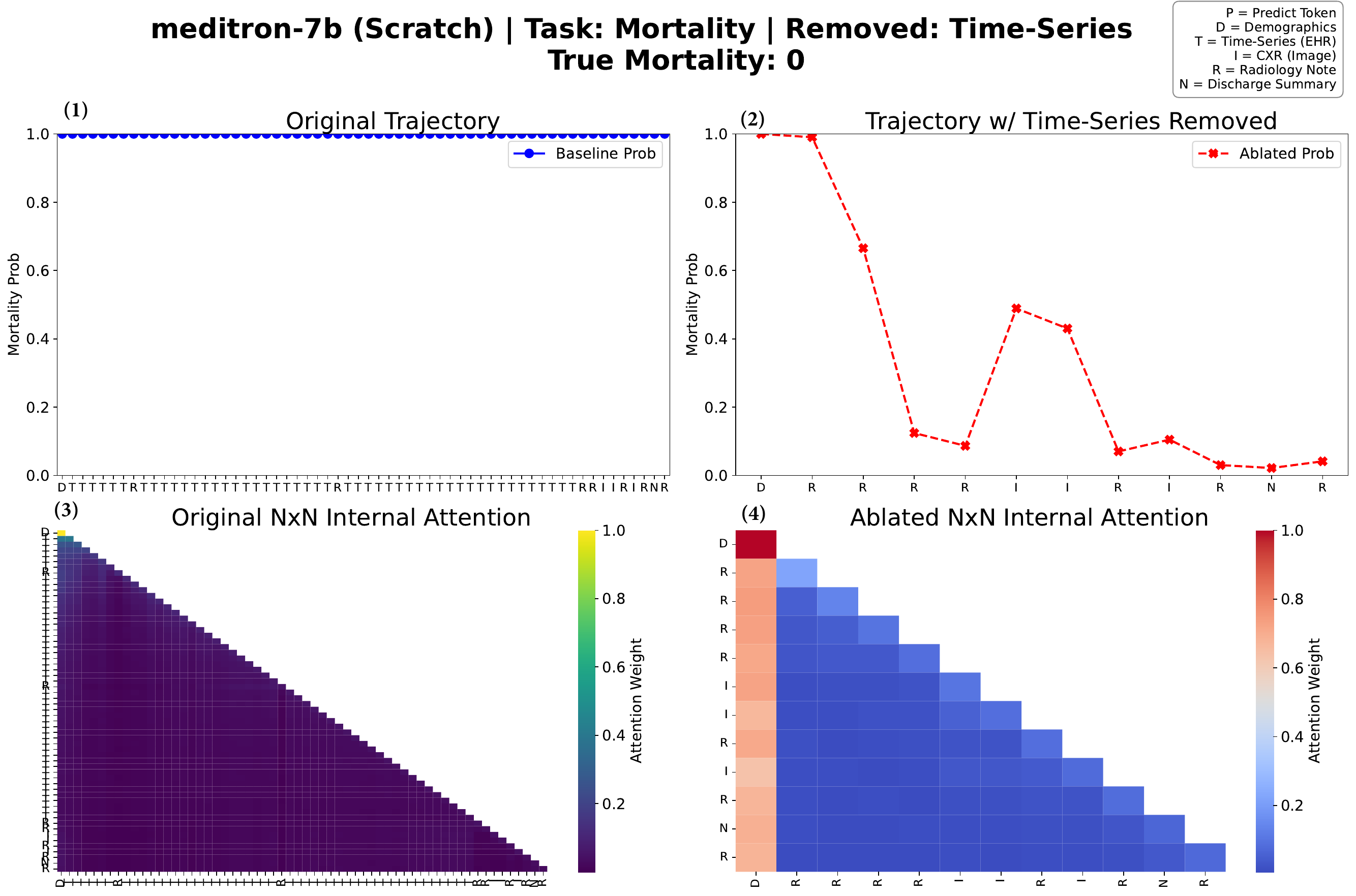}
        \caption{Sequential architecture finetuned on non-contrastive embeddings.} 
        \label{fig:scratch_bias_mimic}
    \end{subfigure}
    
    % Main overarching caption
    \caption{We experimentally remove the time-series embeddings associated with a given MIMIC-IV stay. For the contrastively fine-tuned model, this does not result in a change in the prediction trajectory [\textbf{(a.1)} to \textbf{(a.2)}], nor divergent behavior in the internal attention [\textbf{(a.3) to \textbf{(a.4)}}].  Interestingly, removing this modality corrects the diagnosis for the non-contrastive model [\textbf{(b.1)} to \textbf{(b.2)}].  However, internal attention patterns reveal that this change introduces a new, potentially dangerous reliance on demographic information [\textbf{(b.3)} to \textbf{(b.4)}], indicating the model reached the correct conclusion based on a hidden dependence on demographic data.}
    \label{fig:mimic_safety}
\end{figure*}

As illustrated in \ref{fig:mimic_safety}, we also identify concerning model reliance on demographic data when modalities are removed.  For the non-contrastive model, removal of the time-series modality embeddings causes a sudden vertical band of attention to form strictly over the Demographics token.  In other words, the downstream predictive tokens bypass the remaining clinical sequence entirely, redirecting the vast majority of their attention mass to the patient's demographic embedding.  This shift in attention reveals a concerning trend that non-contrastive sequential models can lose their clinical grounding and default to demographics as an ``attention sink." Consequently, the model's prediction is no longer driven by the patient's physiological trajectory but instead relies on the probability associated with their demographic priors.  In a clinical deployment scenario, if this baseline architecture lacks certain modalities, it is prone to masking its uncertainty by confidently outputting predictions that are heavily skewed by demographic variables, effectively treating demographics as a proxy for clinical severity.  Conversely, the model initialized with our contrastive embeddings demonstrates robust resilience against this demographic anchoring.  Under the exact same ablation conditions, the contrastive model once again maintains a distributed, clinically grounded attention pattern across the surviving historical sequence.  By profiling two distinct patient stay trajectories and their behavior under simulated missingness scenarios, we suggest that the benefits of our Masked Global Alignment pre-training strategy extend beyond numerical performance boosts to fairer, more robust model reasoning behavior, as revealed by the transparency of sequential models.

We defer further examples of patient stays across other tasks, models, and the eICU dataset as well as the raw data associated with these examples to \ref{sec:more_interp}.

\section{Discussion}
In this work, we introduced a missing modality-aware framework for multimodal clinical sequence modeling and interpretation.  We first developed a novel contrastive pre-training objective, aligning the modality-specific representations into a shared latent space.  We then make use of the integrated data to develop static and sequential diagnostic ML architectures, revealing the performance benefits of sequential modeling of clinical events.  Finally, we move beyond performance metrics and provide examples of dangerous demographic bias reliance in missing modality conditions.  In providing examples of safety-aware interpretation, we also identify the potential of our pre-training approach to mitigate potentially dangerous patterns in sequential models. 

We argue that the benefit of our approach is twofold: First, contrastive pre-training via our Masked Global Alignment strategy creates well-aligned latent representations of multimodal clinical events, conferring robustness to ``missingness" in modeling patient stay trajectories.  Second, autoregressive clinical sequence modeling offers a valuable lens into the black box of ML-aided diagnosis by allowing us to query each step of the decision updating process. 

Taken together, our framework argues for the fusion of missing-modality aware methods with inherently explainable architectures such as sequence models.  Traditional static models are both limited in performance ceiling in missing-modality settings and in interpretability.  In exposing and interpreting these internal mechanisms, we enable explicit auditing of exactly when the model formed its hypothesis and what specific piece of data contributed to the decision, which may more directly aid clinicians in the diagnostic process.  As healthcare increasingly incorporates automated decision-support systems, the underlying ML methodology must be robust to the noisy, asynchronous reality of clinical data.  In doing so, we may sharpen our understanding of how these models integrate sparse, high-dimensional patient data while providing practitioners with a human-interpretable aid that can be deployed in a fair, safety-aware manner.

\paragraph{Limitations}
Due to computational constraints, we restrict the architectures used to models of a narrow size family (3 billion to 8 billion parameters). Future work evaluating larger models may yield more complex, emergent behavior of autoregressive decoders on clinical sequence modeling. Furthermore, because standard Transformers scale quadratically, modeling lifelong patient trajectories remains computationally prohibitive. Adapting our contrastive sequence modeling framework to more efficient architectures like State Space Models (SSMs) or hybrid transfomer-SSM models presents a promising, practical avenue of work. Finally, while we interpret the reasoning process of clinical sequence models and observe that our contrastive pretraining method offers a potential solution to mitigate one type of bias in clinical decision making, it is beyond the scope of this work to develop strictly causal interventions on the patterns observed. A promising direction for future avenues of work include designing automated mechanistic harm mitigation strategies for handling missing modalities in sequential reasoning ML models.

\clearpage 
\bibliographystyle{unsrtnat}  
\bibliography{references}  

@article{Pollard2018,
  author   = {Pollard, Tom J. and Johnson, Alistair E. W. and Raffa, Jesse D. and others},
  title    = {The eICU Collaborative Research Database, a freely available multi-center database for critical care research},
  journal  = {Scientific Data},
  year     = {2018},
  month    = {sep},
  volume   = {5},
  number   = {1},
  pages    = {180178},
  issn     = {2052-4463},
  doi      = {10.1038/sdata.2018.178},
  url      = {https://doi.org/10.1038/sdata.2018.178}
}

@article{Johnson2023_MIMIC-IV,
  author = {Johnson, Alistair E. W. and Bulgarelli, Lucas and Shen, Lu and others},
  doi = {10.1038/s41597-022-01899-x},
  issue = {1},
  journal = {Scientific Data},
  language = {en},
  month = {1},
  publisher = {Springer Science and Business Media LLC},
  title = {MIMIC-IV, a freely accessible electronic health record dataset},
  url = {http://dx.doi.org/10.1038/s41597-022-01899-x},
  volume = {10},
  year = {2023},
}

@article{Johnson2019-0c,
  author = {Johnson, Alistair EW and Pollard, Tom J and Berkowitz, Seth J and others},
  journal = {Scientific Data},
  publisher = {Nature Publishing Group},
  title = {MIMIC-CXR, a de-identified publicly available database of chest radiographs with free-text reports},
  volume = {6},
  year = {2019},
}

@misc{wang2025moehealthmixtureexpertsframework,
      title={MoE-Health: A Mixture of Experts Framework for Robust Multimodal Healthcare Prediction}, 
      author={Wang, Xiaoyang and Yang, Christopher C.},
      year={2025},
      eprint={2508.21793},
      archivePrefix={arXiv},
      primaryClass={cs.LG},
      url={https://arxiv.org/abs/2508.21793}, 
}

@article{10.1145/3746456,
  author = {Geraghty, Jack and Hines, Andrew and Golpayegani, Fatemeh},
  title = {Learning to Associate: Multimodal Inference with Fully Missing Modalities},
  year = {2025},
  publisher = {Association for Computing Machinery},
  volume = {16},
  number = {5},
  url = {https://doi.org/10.1145/3746456},
  doi = {10.1145/3746456},
  journal = {ACM Trans. Intell. Syst. Technol.}
}

@InProceedings{pmlr-v250-mordacq24a,
  title =    {ADAPT: Multimodal Learning for Detecting Physiological Changes under Missing Modalities},
  author =       {Mordacq, Julie and Milecki, Leo and Vakalopoulou, Maria and others},
  booktitle =    {Proceedings of The 7nd International Conference on Medical Imaging with Deep Learning},
  pages =    {1040--1055},
  year =     {2024},
  volume =   {250},
  series =   {Proceedings of Machine Learning Research},
  url =      {https://proceedings.mlr.press/v250/mordacq24a.html}
}

@misc{li2025simmlmsimpleframeworkmultimodal,
      title={SimMLM: A Simple Framework for Multi-modal Learning with Missing Modality}, 
      author={Li, Sijie and Chen, Chen and Han, Jungong},
      year={2025},
      eprint={2507.19264},
      archivePrefix={arXiv},
      primaryClass={cs.CV},
      url={https://arxiv.org/abs/2507.19264}, 
}

@INPROCEEDINGS{10204754,
  author={Wang, Hu and Chen, Yuanhong and Ma, Congbo and others},
  booktitle={2023 IEEE/CVF Conference on Computer Vision and Pattern Recognition (CVPR)}, 
  title={Multi-Modal Learning with Missing Modality via Shared-Specific Feature Modelling}, 
  year={2023},
  pages={15878-15887},
  doi={10.1109/CVPR52729.2023.01524}
}

@article{LIU2023104466,
  title = {Attention-based multimodal fusion with contrast for robust clinical prediction in the face of missing modalities},
  journal = {Journal of Biomedical Informatics},
  volume = {145},
  pages = {104466},
  year = {2023},
  doi = {https://doi.org/10.1016/j.jbi.2023.104466},
  author = {Liu, Jinghui and Capurro, Daniel and Nguyen, Anthony and Verspoor, Karin}
}

@misc{tölle2025arbitrarydataimagesfusion,
      title={Arbitrary Data as Images: Fusion of Patient Data Across Modalities and Irregular Intervals with Vision Transformers}, 
      author={Tölle, Malte and Scharaf, Mohamad and Fischer, Samantha and others},
      year={2025},
      eprint={2501.18237},
      archivePrefix={arXiv},
      primaryClass={cs.CV},
      url={https://arxiv.org/abs/2501.18237}, 
}

@misc{lee2023learningmissingmodalelectronic,
      title={Learning Missing Modal Electronic Health Records with Unified Multi-modal Data Embedding and Modality-Aware Attention}, 
      author={Lee, Kwanhyung and Lee, Soojeong and Hahn, Sangchul and others},
      year={2023},
      eprint={2305.02504},
      archivePrefix={arXiv},
      primaryClass={cs.LG},
      url={https://arxiv.org/abs/2305.02504}, 
}

@misc{wang2023missingmodalityenabledmultimodalfusion,
      title={Missing-modality Enabled Multi-modal Fusion Architecture for Medical Data}, 
      author={Wang, Muyu and Fan, Shiyu and Li, Yichen and Chen, Hui},
      year={2023},
      eprint={2309.15529},
      archivePrefix={arXiv},
      primaryClass={eess.IV},
      url={https://arxiv.org/abs/2309.15529}, 
}

@article{Yao_Yin_Cheung_Liu_Qin_2024, 
  title={DrFuse: Learning Disentangled Representation for Clinical Multi-Modal Fusion with Missing Modality and Modal Inconsistency}, 
  volume={38}, 
  url={https://ojs.aaai.org/index.php/AAAI/article/view/29578}, 
  DOI={10.1609/aaai.v38i15.29578}, 
  number={15}, 
  journal={Proceedings of the AAAI Conference on Artificial Intelligence}, 
  author={Yao, Wenfang and Yin, Kejing and Cheung, William K. and others}, 
  year={2024}, 
  month={Mar.}, 
  pages={16416-16424} 
}

@misc{golovanevsky2025picmepipelinecontrastivemodality,
      title={PiCME: Pipeline for Contrastive Modality Evaluation and Encoding in the MIMIC Dataset}, 
      author={Golovanevsky, Michal and Mahableshwarkar, Pranav and Eickhoff, Carsten and Singh, Ritambhara},
      year={2025},
      eprint={2507.03165},
      archivePrefix={arXiv},
      primaryClass={cs.LG},
      url={https://arxiv.org/abs/2507.03165}, 
}

@misc{sadanandan2026multimodaldeeplearningearly,
      title={Multimodal Deep Learning for Early Prediction of Patient Deterioration in the ICU: Integrating Time-Series EHR Data with Clinical Notes}, 
      author={Sadanandan, Binesh},
      year={2026},
      eprint={2603.14719},
      archivePrefix={arXiv},
      primaryClass={cs.LG},
      url={https://arxiv.org/abs/2603.14719}, 
}

@article{Sheikhalishahi_2020,
   title={Benchmarking machine learning models on multi-centre eICU critical care dataset},
   volume={15},
   url={http://dx.doi.org/10.1371/journal.pone.0235424},
   DOI={10.1371/journal.pone.0235424},
   number={7},
   journal={PLOS ONE},
   author={Sheikhalishahi, Seyedmostafa and Balaraman, Vevake and Osmani, Venet},
   year={2020},
   month=jul, 
   pages={e0235424} 
}

@article{Abuhamad2026,
  author = {Abuhamad, Husam and Zainudin, Suhaila and Abu Bakar, Azuraliza},
  title = {Integrative multimodal hybrid data fusion for mortality prediction},
  journal = {Scientific Reports},
  volume = {16},
  number = {1},
  pages = {5803},
  year = {2026},
  month = {jan},
  doi = {10.1038/s41598-026-36296-6},
  url = {https://doi.org/10.1038/s41598-026-36296-6}
}

@article{WANG2024106672,
  title = {Multimodal fusion network for ICU patient outcome prediction},
  journal = {Neural Networks},
  volume = {180},
  pages = {106672},
  year = {2024},
  doi = {https://doi.org/10.1016/j.neunet.2024.106672},
  url = {https://www.sciencedirect.com/science/article/pii/S0893608024005963},
  author = {Wang, Chutong and Yang, Xuebing and Sun, Mengxuan and others}
}

@article{SHE202423,
  title = {Interpretable machine learning models for predicting 90-day death in patients in the intensive care unit with epilepsy},
  journal = {Seizure: European Journal of Epilepsy},
  volume = {114},
  pages = {23-32},
  year = {2024},
  doi = {https://doi.org/10.1016/j.seizure.2023.11.017},
  url = {https://www.sciencedirect.com/science/article/pii/S1059131123003047},
  author = {She, Yingfang and Zhou, Liemin and Li, Yide}
}

@inproceedings{Rocheteau_2021, 
   title={Temporal pointwise convolutional networks for length of stay prediction in the intensive care unit},
   url={http://dx.doi.org/10.1145/3450439.3451860},
   DOI={10.1145/3450439.3451860},
   booktitle={Proceedings of the Conference on Health, Inference, and Learning},
   author={Rocheteau, Emma and Liò, Pietro and Hyland, Stephanie},
   year={2021},
   month=apr, 
   pages={58–68} 
}

@Article{info:doi/10.2196/74142,
  author="Naliyatthaliyazchayil, Parvati and Muthyala, Raajitha and Gichoya, Judy Wawira and others",
  title="Evaluating the Reasoning Capabilities of Large Language Models for Medical Coding and Hospital Readmission Risk Stratification: Zero-Shot Prompting Approach",
  journal="J Med Internet Res",
  year="2025",
  volume="27",
  pages="e74142",
  doi="10.2196/74142",
  url="https://doi.org/10.2196/74142"
}

@misc{liao2025ehrr1reasoningenhancedfoundationallanguage,
      title={EHR-R1: A Reasoning-Enhanced Foundational Language Model for Electronic Health Record Analysis}, 
      author={Liao, Yusheng and Wu, Chaoyi and Liu, Junwei and others},
      year={2025},
      eprint={2510.25628},
      archivePrefix={arXiv},
      primaryClass={cs.CL},
      url={https://arxiv.org/abs/2510.25628}, 
}

@article{Qiu2025,
  author = {Qiu, Pengcheng and Wu, Chaoyi and Liu, Shuyu and others},
  title = {Quantifying the reasoning abilities of LLMs on clinical cases},
  journal = {Nature Communications},
  volume = {16},
  number = {1},
  pages = {9799},
  year = {2025},
  month = {nov},
  doi = {10.1038/s41467-025-64769-1},
  url = {https://doi.org/10.1038/s41467-025-64769-1}
}

@misc{zhu2025causaldebiasingmedicalmultimodal,
      title={Causal Debiasing Medical Multimodal Representation Learning with Missing Modalities}, 
      author={Zhu, Xiaoguang and Sun, Lianlong and Liu, Yang and others},
      year={2025},
      eprint={2509.05615},
      archivePrefix={arXiv},
      primaryClass={cs.LG},
      url={https://arxiv.org/abs/2509.05615}, 
}

@Article{diagnostics11122242,
  AUTHOR = {Wu, Jingyi and Lin, Yu and Li, Pengfei and others},
  TITLE = {Predicting Prolonged Length of ICU Stay through Machine Learning},
  JOURNAL = {Diagnostics},
  VOLUME = {11},
  YEAR = {2021},
  NUMBER = {12},
  URL = {https://www.mdpi.com/2075-4418/11/12/2242},
  DOI = {10.3390/diagnostics11122242}
}

@misc{gong2026embracingaleatoricuncertaintymedical,
      title={Embracing Aleatoric Uncertainty in Medical Multimodal Learning with Missing Modalities}, 
      author={Gong, Linxiao and Liu, Yang and Sun, Lianlong and others},
      year={2026},
      eprint={2601.21950},
      archivePrefix={arXiv},
      primaryClass={cs.LG},
      url={https://arxiv.org/abs/2601.21950}, 
}

@misc{liventsev2024intensivecarebigsequence,
      title={Intensive Care as One Big Sequence Modeling Problem}, 
      author={Liventsev, Vadim and Fritz, Tobias},
      year={2024},
      eprint={2402.17501},
      archivePrefix={arXiv},
      primaryClass={cs.LG},
      url={https://arxiv.org/abs/2402.17501}, 
}

@inproceedings{10.1145/3777577.3777712,
  author = {Zheng, Yi and Zhao, Fei and Liu, Xiaohua and others},
  title = {A Multimodal Deep Learning Framework for Predicting Cardiovascular Deterioration Based on MIMIC-IV Dataset},
  year = {2026},
  url = {https://doi.org/10.1145/3777577.3777712},
  doi = {10.1145/3777577.3777712},
  booktitle = {Proceedings of the 2025 6th International Symposium on Artificial Intelligence for Medical Sciences},
  pages = {837–842}
}

@article{Lin2024,
  author = {Lin, Jiaxi and Yang, Jin and Yin, Minyue and others},
  title = {Development and Validation of Multimodal Models to Predict the 30-Day Mortality of ICU Patients Based on Clinical Parameters and Chest X-Rays},
  journal = {Journal of Imaging Informatics in Medicine},
  volume = {37},
  number = {4},
  pages = {1312--1322},
  year = {2024},
  doi = {10.1007/s10278-024-01066-1},
  url = {https://doi.org/10.1007/s10278-024-01066-1}
}

@INPROCEEDINGS{11095208,
  author={Chen, Wanyi and Zhao, Zihua and Yao, Jiangchao and others},
  booktitle={2025 IEEE/CVF Conference on Computer Vision and Pattern Recognition (CVPR)}, 
  title={Multi-modal Medical Diagnosis via Large-small Model Collaboration}, 
  year={2025},
  pages={30763-30773},
  doi={10.1109/CVPR52734.2025.02865}
}

@article{Soenksen2022,
  author = {Soenksen, Luis R. and Ma, Yu and Zeng, Cynthia and others},
  title = {Integrated multimodal artificial intelligence framework for healthcare applications},
  journal = {npj Digital Medicine},
  volume = {5},
  number = {1},
  pages = {149},
  year = {2022},
  doi = {10.1038/s41746-022-00689-4},
  url = {https://doi.org/10.1038/s41746-022-00689-4}
}

@article{Renc2024,
  author = {Renc, Pawel and Jia, Yugang and Samir, Anthony E. and others},
  title = {Zero shot health trajectory prediction using transformer},
  journal = {npj Digital Medicine},
  volume = {7},
  number = {1},
  pages = {256},
  year = {2024},
  doi = {10.1038/s41746-024-01235-0},
  url = {https://doi.org/10.1038/s41746-024-01235-0}
}

@ARTICLE{9964038,
  author={Li, Yikuan and Mamouei, Mohammad and Salimi-Khorshidi, Gholamreza and others},
  journal={IEEE Journal of Biomedical and Health Informatics}, 
  title={Hi-BEHRT: Hierarchical Transformer-Based Model for Accurate Prediction of Clinical Events Using Multimodal Longitudinal Electronic Health Records}, 
  year={2023},
  volume={27},
  number={2},
  pages={1106-1117},
  doi={10.1109/JBHI.2022.3224727}
}

@article {Bornet2023.06.01.23290824,
	author = {Bornet, Alban and Proios, Dimitrios and Yazdani, Anthony and others},
	title = {Comparing neural language models for medical concept representation and patient trajectory prediction},
	year = {2024},
	doi = {10.1101/2023.06.01.23290824},
	URL = {https://www.medrxiv.org/content/early/2024/10/22/2023.06.01.23290824},
	journal = {medRxiv}
}

@misc{wang2026revisitingperformanceclaimschest,
      title={Revisiting Performance Claims for Chest X-Ray Models Using Clinical Context}, 
      author={Andrew Wang and Jiashuo Zhang and Michael Oberst},
      year={2026},
      eprint={2509.19671},
      archivePrefix={arXiv},
      primaryClass={cs.LG},
      url={https://arxiv.org/abs/2509.19671}, 
}

@misc{huang2020clinicalbertmodelingclinicalnotes,
      title={ClinicalBERT: Modeling Clinical Notes and Predicting Hospital Readmission}, 
      author={Huang, Kexin and Altosaar, Jaan and Ranganath, Rajesh},
      year={2020},
      eprint={1904.05342},
      archivePrefix={arXiv},
      primaryClass={cs.CL},
      url={https://arxiv.org/abs/1904.05342}, 
}

@misc{chen2023meditron70bscalingmedicalpretraining,
      title={MEDITRON-70B: Scaling Medical Pretraining for Large Language Models}, 
      author={Chen, Zeming and Cano, Alejandro Hernández and Romanou, Angelika and others},
      year={2023},
      eprint={2311.16079},
      archivePrefix={arXiv},
      primaryClass={cs.CL},
      url={https://arxiv.org/abs/2311.16079}, 
}

@misc{grattafiori2024llama3herdmodels,
      title={The Llama 3 Herd of Models}, 
      author={Grattafiori, Aaron and Dubey, Abhimanyu and Jauhri, Abhinav and others},
      year={2024},
      eprint={2407.21783},
      archivePrefix={arXiv},
      primaryClass={cs.AI},
      url={https://arxiv.org/abs/2407.21783}, 
}

@misc{jiang2023mistral7b,
      title={Mistral 7B}, 
      author={Jiang, Albert Q. and Sablayrolles, Alexandre and Mensch, Arthur and others},
      year={2023},
      eprint={2310.06825},
      archivePrefix={arXiv},
      primaryClass={cs.CL},
      url={https://arxiv.org/abs/2310.06825}, 
}

@misc{labrak2024biomistralcollectionopensourcepretrained,
      title={BioMistral: A Collection of Open-Source Pretrained Large Language Models for Medical Domains}, 
      author={Labrak, Yanis and Bazoge, Adrien and Morin, Emmanuel and others},
      year={2024},
      eprint={2402.10373},
      archivePrefix={arXiv},
      primaryClass={cs.CL},
      url={https://arxiv.org/abs/2402.10373}, 
}

@misc{deepseekai2024deepseekllmscalingopensource,
      title={DeepSeek LLM: Scaling Open-Source Language Models with Longtermism}, 
      author={DeepSeek-AI and Bi, Xiao and Chen, Deli and others},
      year={2024},
      eprint={2401.02954},
      archivePrefix={arXiv},
      primaryClass={cs.CL},
      url={https://arxiv.org/abs/2401.02954}, 
}

@misc{abdin2024phi3technicalreporthighly,
      title={Phi-3 Technical Report: A Highly Capable Language Model Locally on Your Phone}, 
      author={Abdin, Marah and Aneja, Jyoti and Awadalla, Hany and others},
      year={2024},
      eprint={2404.14219},
      archivePrefix={arXiv},
      primaryClass={cs.CL},
      url={https://arxiv.org/abs/2404.14219}, 
}

@misc{wornow2025contextcluesevaluatinglong,
      title={Context Clues: Evaluating Long Context Models for Clinical Prediction Tasks on EHRs}, 
      author={Wornow, Michael and Bedi, Suhana and Hernandez, Miguel Angel Fuentes and others},
      year={2025},
      eprint={2412.16178},
      archivePrefix={arXiv},
      primaryClass={cs.LG},
      url={https://arxiv.org/abs/2412.16178}, 
}

@misc{hu2021loralowrankadaptationlarge,
      title={LoRA: Low-Rank Adaptation of Large Language Models}, 
      author={Hu, Edward J. and Shen, Yelong and Wallis, Phillip and others},
      year={2021},
      eprint={2106.09685},
      archivePrefix={arXiv},
      primaryClass={cs.CL},
      url={https://arxiv.org/abs/2106.09685}, 
}

@misc{vaswani2023attentionneed,
      title={Attention Is All You Need}, 
      author={Ashish Vaswani and Noam Shazeer and Niki Parmar and Jakob Uszkoreit and Llion Jones and Aidan N. Gomez and Lukasz Kaiser and Illia Polosukhin},
      year={2023},
      eprint={1706.03762},
      archivePrefix={arXiv},
      primaryClass={cs.CL},
      url={https://arxiv.org/abs/1706.03762}, 
}

@article{lee2024fill,
  title = {Fill in the Blank, MIMIC the Procedure: Tracking Patient Record for Medical Data Reconstruction},
  author = {Lee, Sujung and Seo, Daechul and Ko, Taehoon},
  journal = {Studies in health technology and informatics},
  volume = {316},
  pages = {1594--1595},
  year = {2024},
  doi = {10.3233/SHTI240726},
  pmid = {39176513}
}

@article{Poette2026,
  title = {Benchmarking imputation strategies for missing time-series data in critical care using real-world-inspired scenarios},
  author = {Poette, Michael and Mouysset, Sandrine and Ruiz, Daniel and Pey, Vincent and Alliot, Jean-Marc and Minville, Vincent},
  journal = {Scientific Reports},
  volume = {16},
  number = {1},
  pages = {8116},
  year = {2026},
  doi = {10.1038/s41598-026-39035-z},
  url = {https://doi.org/10.1038/s41598-026-39035-z},
  issn = {2045-2322}
}

@misc{chen2020simpleframeworkcontrastivelearning,
      title={A Simple Framework for Contrastive Learning of Visual Representations}, 
      author={Ting Chen and Simon Kornblith and Mohammad Norouzi and Geoffrey Hinton},
      year={2020},
      eprint={2002.05709},
      archivePrefix={arXiv},
      primaryClass={cs.LG},
      url={https://arxiv.org/abs/2002.05709}, 
}

@article{Thapa2026,
  title = {A multimodal sleep foundation model for disease prediction},
  author = {Thapa, Rahul and Kjaer, Magnus Ruud and He, Bryan and Covert, Ian and Moore IV, Hyatt and Hanif, Umaer and Ganjoo, Gauri and Westover, M. Brandon and Jennum, Poul and Brink-Kjaer, Andreas and Mignot, Emmanuel and Zou, James},
  journal = {Nature Medicine},
  volume = {32},
  number = {2},
  pages = {752--762},
  year = {2026},
  doi = {10.1038/s41591-025-04133-4},
  url = {https://doi.org/10.1038/s41591-025-04133-4},
  issn = {1546-170X}
}

\clearpage 
\appendix
\renewcommand{\thefigure}{A.\arabic{figure}}
\setcounter{figure}{0}

\renewcommand{\thetable}{A.\arabic{table}}
\setcounter{table}{0}

\section{Clinical Dataset Preprocessing}
\label{sec:data_preprocessing}

\subsection{Contrastive Pretraining Data}
\subsubsection{Cohort Selection and Data Linkage}
\textbf{MIMIC-IV Cohort:} The MIMIC cohort focuses on intensive care unit (ICU) admissions where multimodal data streams intersect. We performed an inner join between the MIMIC-CXR metadata and the MIMIC-IV \texttt{icustays} table on \texttt{subject\_id}. To ensure temporal consistency, we strictly aligned data availability with the ICU admission window: chest X-rays were filtered for images acquired where the \texttt{StudyDateTime} fell strictly between hospital admission (\texttt{intime}) and discharge (\texttt{outtime}). To maintain visual consistency, we restricted the imaging dataset to Anteroposterior (AP) views only. In cases where multiple AP views existed for a single ICU stay, the most temporally recent X-ray (\texttt{StudyDateTime}) was selected. The final pretraining MIMIC cohort includes 16,716 unique ICU stays possessing at least one valid modality.

\textbf{eICU Cohort:} For the eICU cohort, we extracted data from the \texttt{patient}, \texttt{diagnosis}, \texttt{treatment}, \texttt{medication}, \texttt{lab}, and \texttt{apacheApsVar} tables. In preparation for downstream predictive tasks (which are anchored 12 hours post-admission), we applied rigorous filtering criteria to the \texttt{patient} table. Patient age strings (e.g., ``> 89'') were parsed into integers, and we excluded patients younger than 18 or older than 89 years. Length of stay (LoS) was calculated by subtracting the \texttt{hospitaladmitoffset} from the \texttt{hospitaldischargeoffset}. We excluded visits lasting longer than 720 hours (30 days) and visits shorter than 12 hours. By anchoring the relational joins on the laboratory table (\texttt{patientunitstayid}), we established a final pretraining eICU cohort of 195,691 unique ICU stays.

\subsubsection{Demographics and Static Features}
\textbf{MIMIC-IV:} We extracted eight static features from the \texttt{core} module: admission type, admission location, insurance, language, marital status, ethnicity, gender, and age group. Missing continuous values were imputed using the dataset mean, and missing categorical variables were imputed with the most frequent value. 

\textbf{eICU:} We extracted age, gender, and ethnicity directly from the \texttt{patient} table. 

\textbf{Common Processing:} Across both datasets, categorical variables were one-hot encoded, ignoring unknown categories. Continuous variables were standardized using z-score normalization fit exclusively on the training split to prevent data leakage. 

\subsubsection{Physiological Signals and Laboratory Measurements}
\textbf{MIMIC-IV:} Physiological signals and laboratory measurements in the EHR time-series data were processed using a fixed discretization window (\texttt{timestep=1.0} hours). Measurements that fell within the same hour block were aggregated. To strictly respect causality and temporal flow, missing values within a sequence were imputed using a ``carry-forward'' (previous value) strategy. If a variable was entirely missing for a patient's entire trajectory, it was imputed with a pre-computed global normal value.

\textbf{eICU:} We filtered for laboratory result types with at least 10,000 instances across the cohort to reduce sparse feature dimensions. For duplicate lab result types within a single patient stay, the total values were summed. We pivoted the data to form a feature matrix where each column represents a specific lab test, and entirely missing laboratory records for a stay were filled with zeros. Acute physiology scores (e.g., motor, verbal, meds, urine) were additionally extracted from the \texttt{apacheApsVar} table and retained as continuous features.

\subsubsection{Clinical Text and Medical Codes}
\textbf{MIMIC-IV:} We aggregated Discharge Summaries (\texttt{discharge.csv}) and Radiology Reports (\texttt{radiology.csv}) by grouping on \texttt{subject\_id} and \texttt{hadm\_id}, concatenating all unique texts per stay. Texts were converted to lowercase and specific de-identification artifacts (headers containing ``name'', ``unit no'', ``admission date'', ``discharge date'', and ``date birth'') were removed to reduce noise. Broken unicode characters (e.g., \texttt{\textbackslash x95}) were explicitly stripped via regex. Crucially, standard punctuation and stop-words were retained to preserve clinical semantics (e.g., negation and uncertainty). Texts were tokenized using the \texttt{ClinicalBERT} tokenizer \cite{huang2020clinicalbertmodelingclinicalnotes} with a maximum sequence length of 512 tokens.

\textbf{eICU:} Because the eICU dataset relies heavily on structured medical codes rather than free text, we aggregated \texttt{diagnosis}, \texttt{treatment}, and \texttt{medication} strings. For each table, strings were grouped by \texttt{patientunitstayid} and concatenated using a pipe delimiter (\texttt{|}). We extracted the unique set of clinical events per stay and generated one-hot encoded dummy variables. To ensure computational feasibility and robust representation, we calculated the total frequency of each dummy variable across the cohort and dropped any clinical event present in fewer than 5\% of total visits.

\subsubsection{Medical Imaging (MIMIC-IV Only)}
MIMIC-CXR images were resized to $256 \times 256$ pixels and center-cropped to $224 \times 224$. Pixel values were normalized using standard ImageNet mean and standard deviation. We applied the following augmentations to the pretraining set:
\begin{itemize}
    \item Random Resized Crop (scale $0.8 \text{-} 1.0$).
    \item Random Horizontal Flip ($p=0.5$).
    \item Random Affine transformation (rotation $\pm 10^{\circ}$, translation $\le 5\%$, scaling $0.95 \text{-} 1.05$).
    \item Color Jitter (brightness 0.2, contrast 0.2).
\end{itemize}

\subsection{Contrastive Pretraining Data}
\subsubsection{Cohort Isolation and Target Definition}
To prevent representation leakage, the fine-tuning cohort was strictly sampled from the pool of patient stays that were \textit{unseen} during the pretraining phase (i.e., the pretraining validation and test splits). 

For the \textbf{eICU dataset}, we anchored the cohort on the presence of valid laboratory measurements and applied the clinical inclusion criteria (age $\ge$ 18 and < 90, Length of Stay between 12 and 720 hours). This yielded an unseen fine-tuning pool of 58,538 stays. We defined two targets: Length of Stay (LoS, continuous regression in hours) and In-Hospital Mortality (binary classification).

For the \textbf{MIMIC-IV dataset}, we defined two distinct task cohorts. The \textit{In-Hospital Mortality} (IHM) task included 16,716 stays (yielding 11,696 training, 2,511 validation, and 2,509 testing samples after our hybrid modality-aware stratification). The \textit{Phenotyping} task, framed as a 25-class multi-label classification problem (e.g., Sepsis, Acute Myocardial Infarction, Shock), utilized a cohort of 7,381 stays (5,171 training, 1,103 validation, and 1,107 testing samples).

\clearpage
\begin{figure*}
	\centering
	\includegraphics[width=0.75\linewidth]{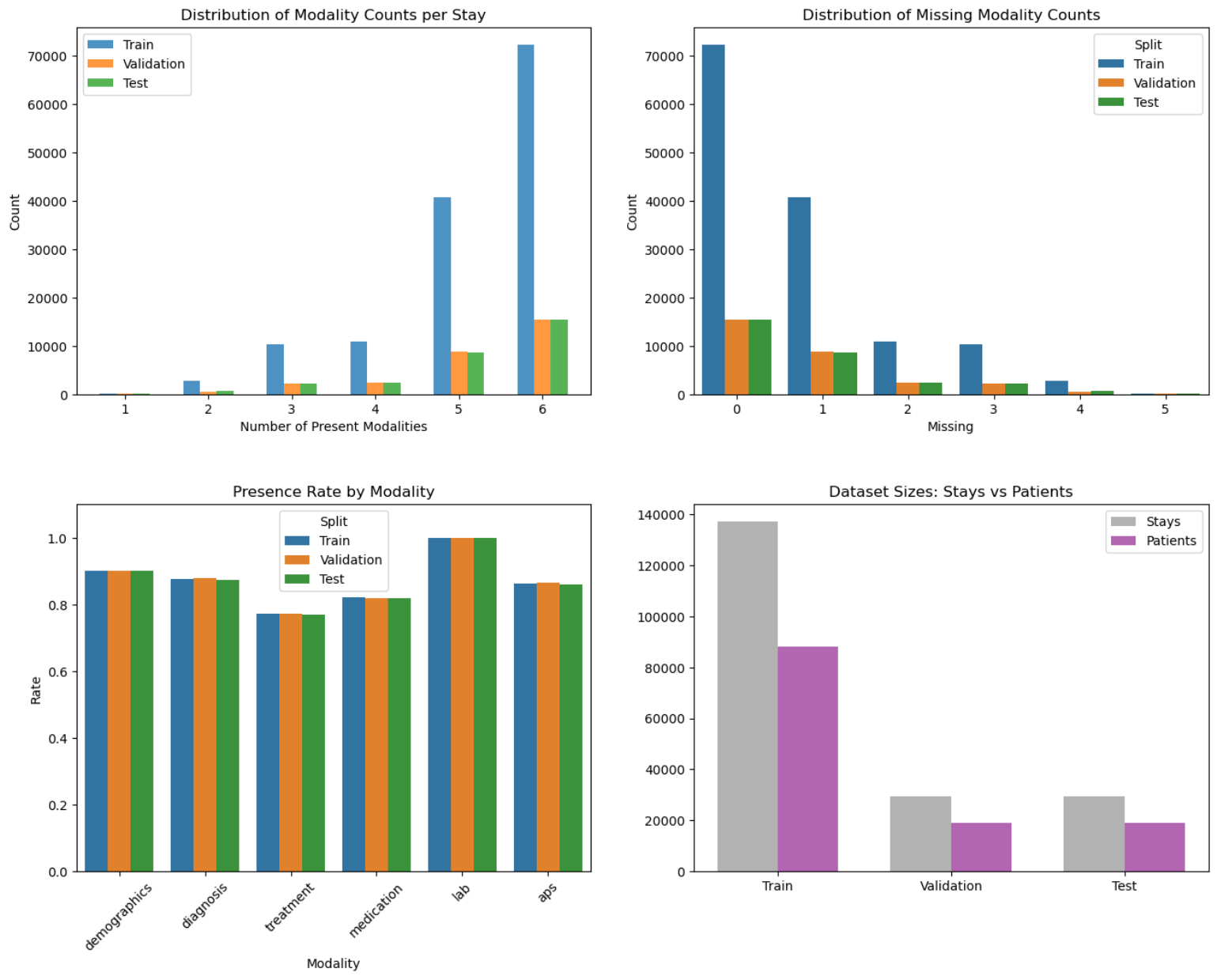}
	\caption{Overview of eICU pretraining data. We display the number of present (top left) and missing (top right) modalities per stay alongside the rate of modality presence (bottom left) across train/val/test splits. We also provide the number of patient and orphaned stays across splits (bottom right).} 
	\label{fig:eICU_pretraining_stats}
\end{figure*}

\begin{figure*}
	\centering
	\includegraphics[width=0.75\linewidth]{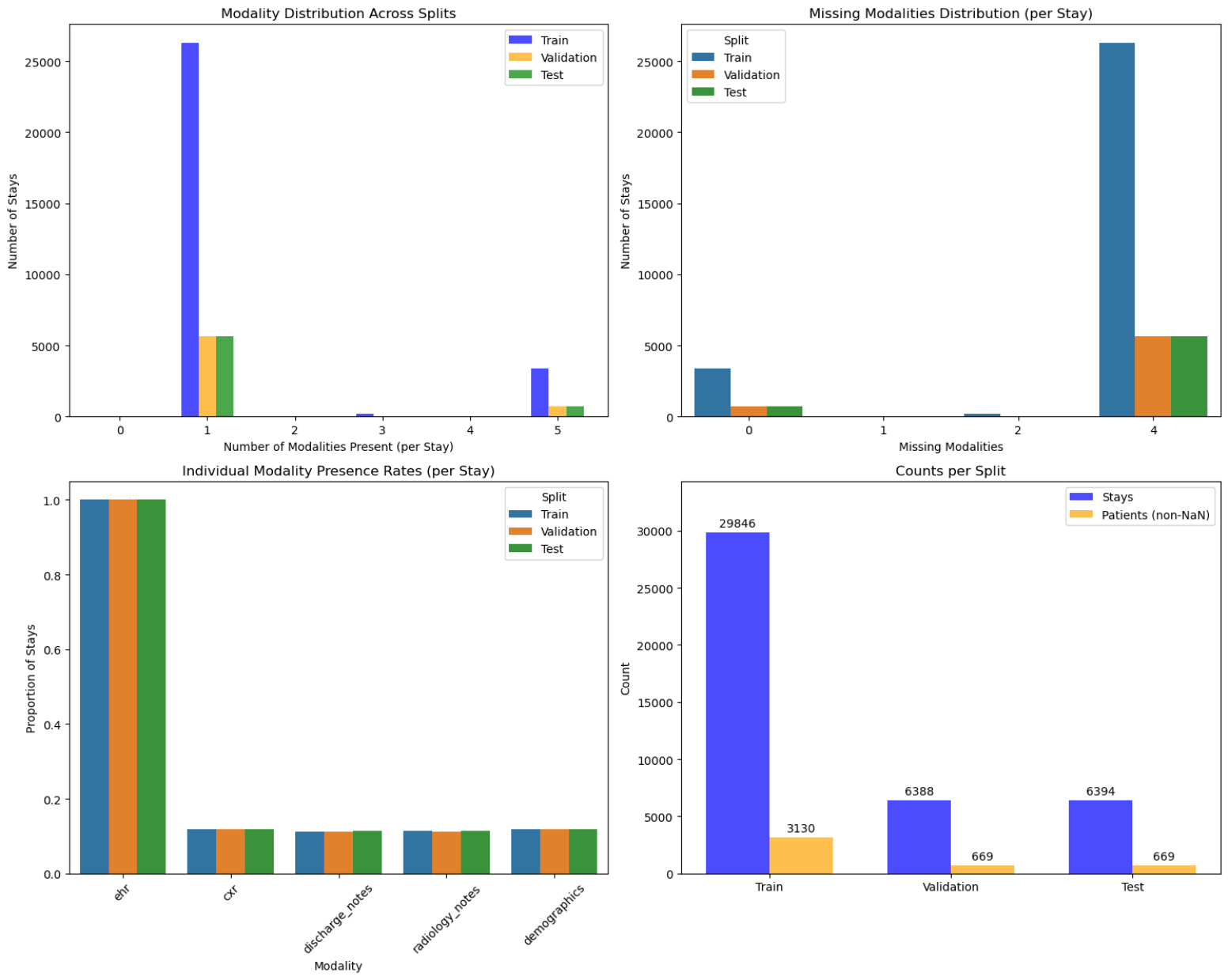}
	\caption{Overview of MIMIC-IV pretraining data. We display the number of present (top left) and missing (top right) modalities per stay alongside the rate of modality presence (bottom left) across train/val/test splits. We also provide the number of patient and orphaned stays across splits (bottom right).} 
	\label{fig:mimic_pretraining}
\end{figure*}

\begin{figure*}
	\centering
	\includegraphics[width=0.75\linewidth]{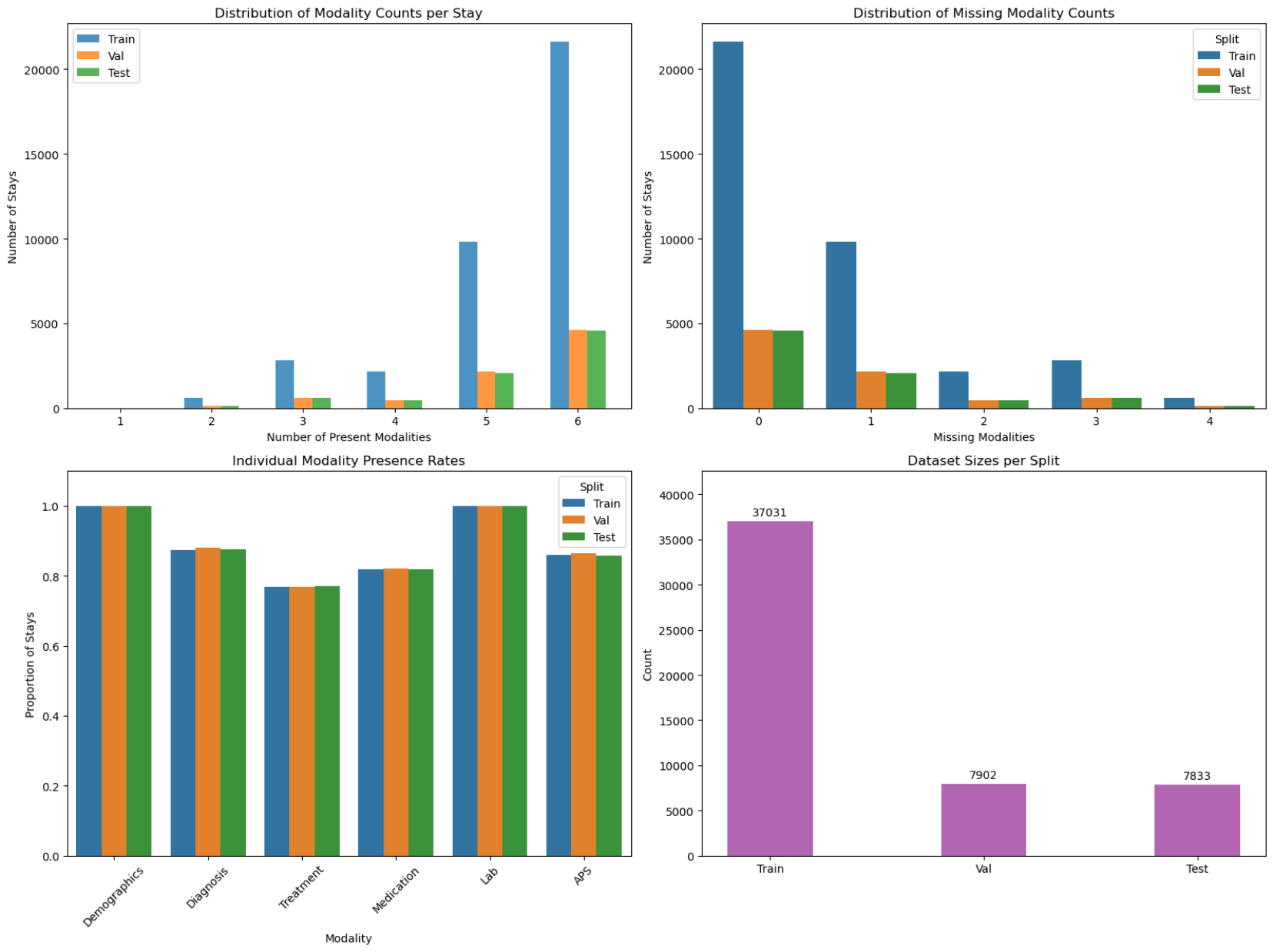}
	\caption{Overview of static eICU finetuning data. We display the number of present (top left) and missing (top right) modalities per stay alongside the rate of modality presence (bottom left) across train/val/test splits.  We also provide the total number of stays across splits (bottom right).} 
	\label{fig:eicu_static_finetuning}
\end{figure*}

\begin{figure*}
	\centering
	\includegraphics[width=0.75\linewidth]{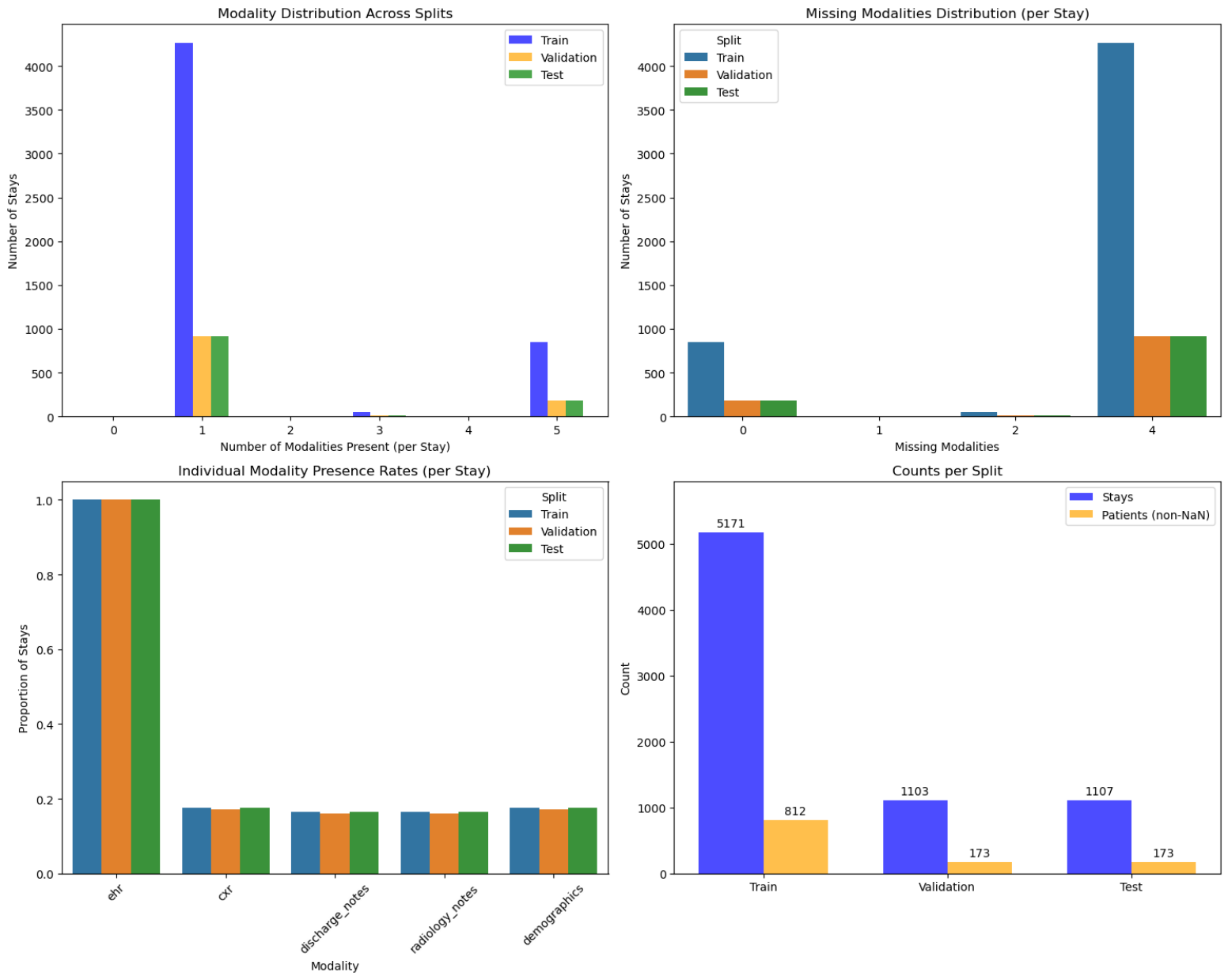}
	\caption{Overview of static MIMIC In-Hospital Mortality finetuning data. We display the number of present (top left) and missing (top right) modalities per stay alongside the rate of modality presence (bottom left) across train/val/test splits.  We also provide the total number of patient and orphaned stays across splits (bottom right).} 
	\label{fig:mimic_static_IHM_data}
\end{figure*}

\begin{figure*}
	\centering
	\includegraphics[width=0.75\linewidth]{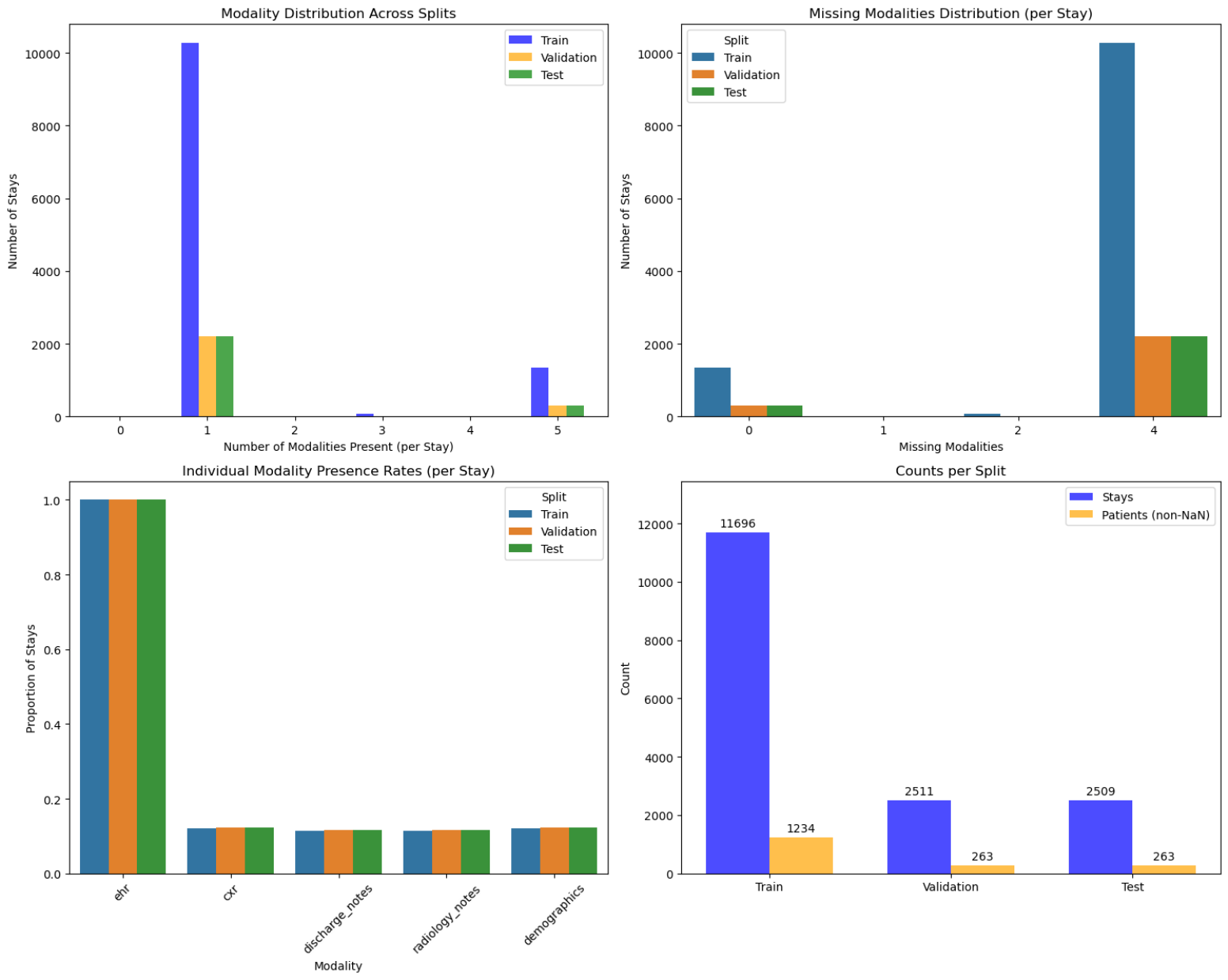}
	\caption{Overview of static MIMIC phenotyping finetuning data. We display the number of present (top left) and missing (top right) modalities per stay alongside the rate of modality presence (bottom left) across train/val/test splits.  We also provide the total number of patient and orphaned stays across splits (bottom right).} 
	\label{fig:mimic_static_pheno_data}
\end{figure*}

\begin{figure*}
	\centering
	\includegraphics[width=0.75\linewidth]{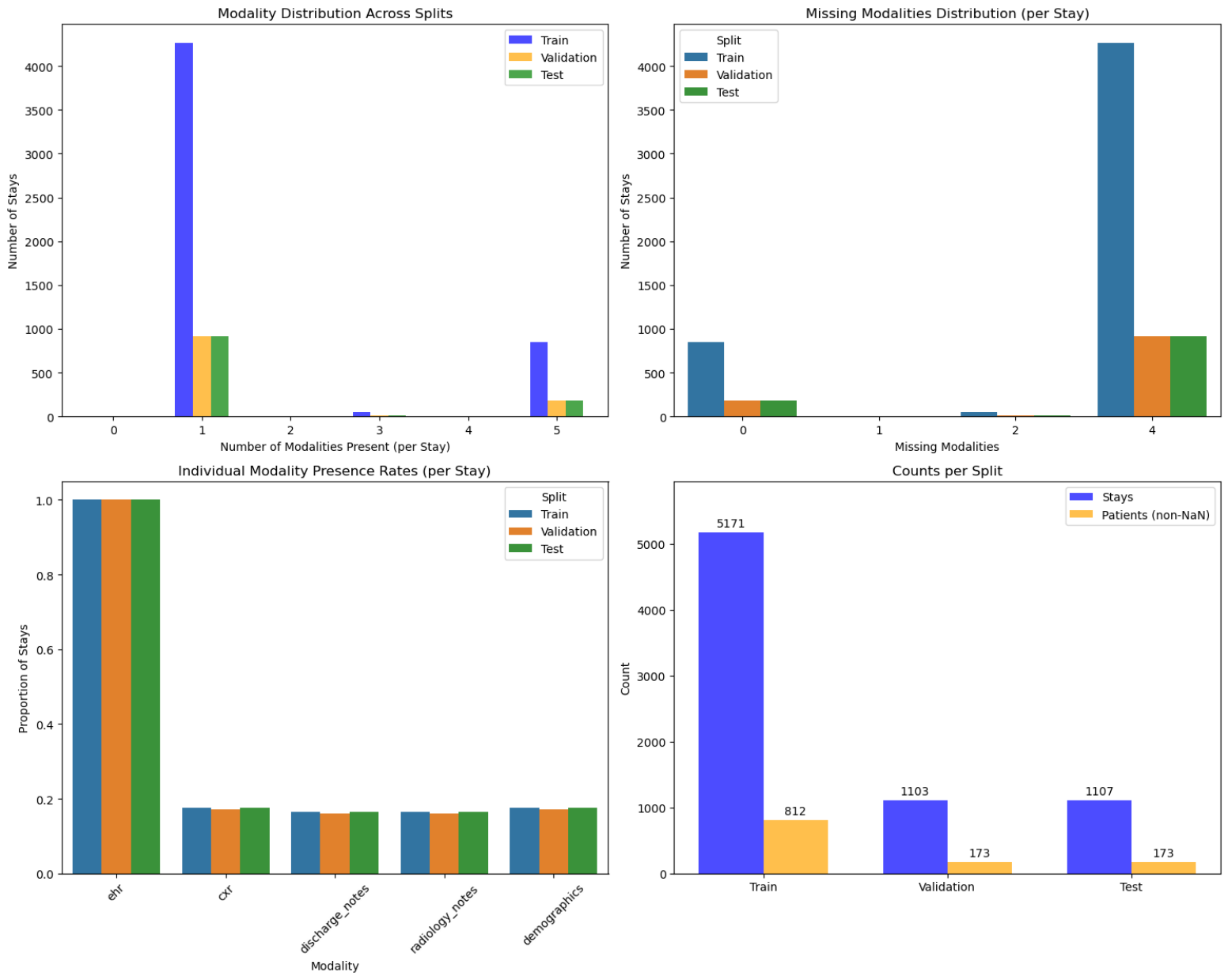}
	\caption{Overview of sequential MIMIC In-Hospital Mortality finetuning  data. We display the number of present (top left) and missing (top right) modalities per stay alongside the rate of modality presence (bottom left) across train/val/test splits.  We also provide the total number of patient and orphaned stays across splits (bottom right).} 
	\label{fig:mimic_LM_IHM_data}
\end{figure*}

\begin{figure*}
	\centering
	\includegraphics[width=0.75\linewidth]{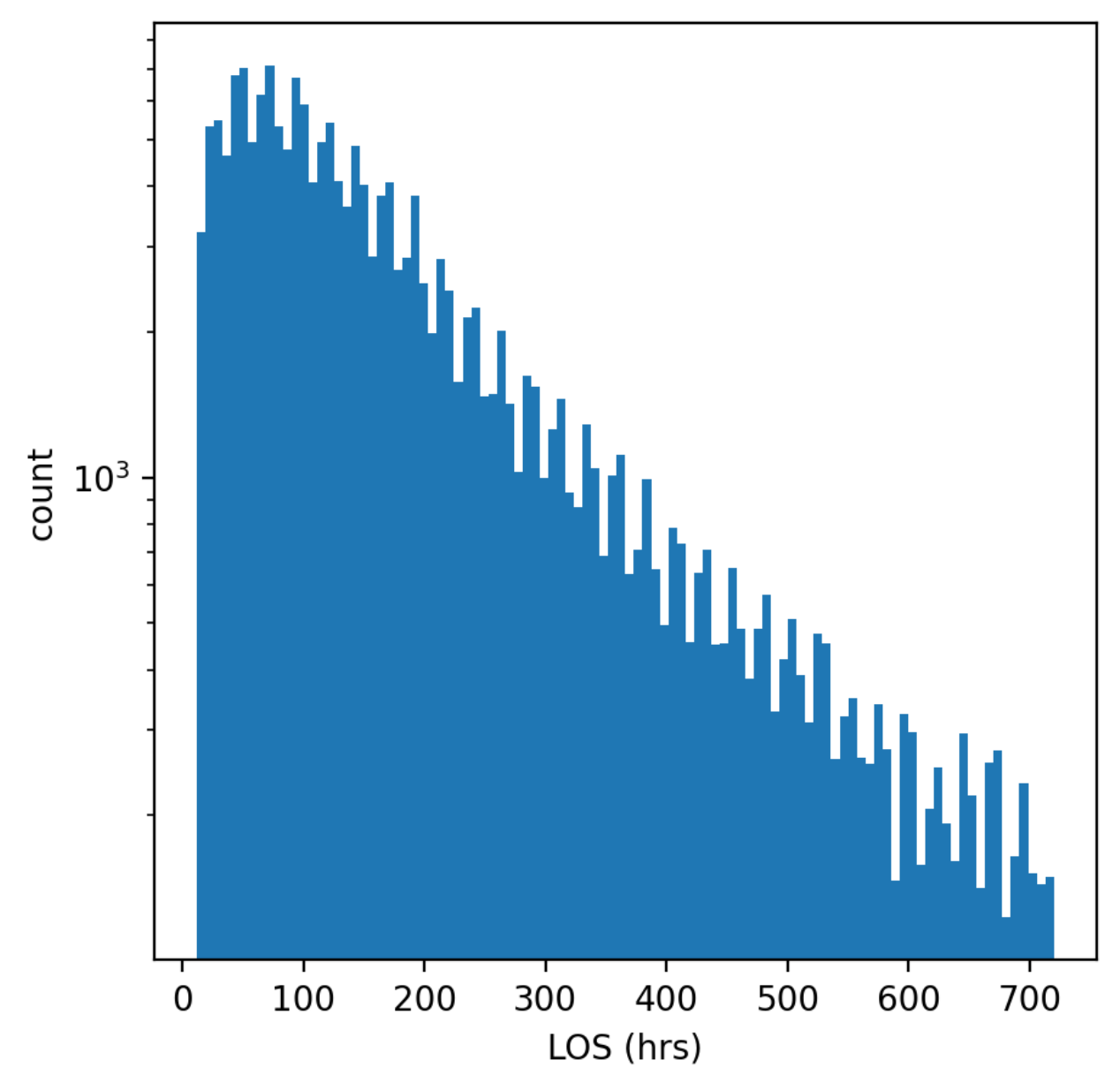}
	\caption{Distribution of length of stays in the eICU finetuning dataset.} 
	\label{fig:eICU_los_data}
\end{figure*}

\clearpage
\section{Architectural/Training Setup}
\label{sec:architectures}

\subsection{Modality-Specific Encoders}
Given the structural differences between our two datasets, we employed dataset-specific encoder architectures to map all inputs to the shared latent dimension ($D=256$ for MIMIC-IV, and sweeping $D \in \{128, 256\}$ for eICU).

\begin{itemize}
    \item \textbf{MIMIC-IV (Heterogeneous Domains):} We utilized domain-specific neural architectures. Clinical notes (Discharge and Radiology) were processed using a pre-trained ClinicalBERT backbone. Chest X-rays were encoded via a Convolutional Neural Network (CNN). The discretized EHR time-series were encoded using a Long Short-Term Memory (LSTM) network. To ensure recurrent stability, the LSTM weights were initialized using exact orthogonal initialization for recurrent matrices and Xavier uniform initialization for input matrices. Static demographics were processed via a standard Multi-Layer Perceptron (MLP).
    \item \textbf{eICU (Tabular Domains):} Because eICU consists entirely of structured tabular records, we utilized modality-agnostic MLPs for all six subsets (Demographics, Diagnosis, Treatment, Medication, Lab, and APS). To handle severe outliers in continuous variables at the dataloader level, raw values were scaled using a sign-preserved log-transformation: $f(x) = \text{sgn}(x) \log(1 + |x|)$. To prevent variance collapse from highly sparse representations, these MLPs substitute standard batch normalization with Layer Normalization, structured as: Linear $\to$ LayerNorm $\to$ ReLU $\to$ Dropout($0.3$) $\to$ Linear $\to$ LayerNorm.
\end{itemize}

\subsection{Training Setup and Hyperparameter Optimization}Models were implemented in PyTorch and trained on NVIDIA RTX 3090 GPUs.\begin{itemize}\item \textbf{MIMIC-IV Configurations:} We optimized the network using Adam with a batch size of 128 for 100 epochs. To identify the optimal convergence paradigm, we swept learning rates across $\{1\times 10^{-3}, 1\times 10^{-4}, 1\times 10^{-5}, 1\times 10^{-6}\}$ and weight decay across $\{0.005, 0.01, 0.05\}$. To simulate downstream real-world missingness and improve generalizability, we implemented stochastic modality dropout during training: a randomly selected target modality was artificially masked with probability $p$ (swept up to $0.25$).\item \textbf{eICU Configurations:} Models were optimized using Adam for 100 epochs, sweeping over batch sizes of $\{128, 256\}$. Due to the density of the tabular features, we swept learning rates $\{1\times 10^{-3}, 1\times 10^{-4}, 5\times 10^{-5}, 1\times 10^{-6}\}$ alongside tighter weight decay parameters $\{0, 1\times 10^{-3}, 1\times 10^{-4}, 1\times 10^{-5}\}$ to prevent overfitting.\end{itemize}In both pipelines, the optimal checkpoint was selected based on the minimization of the global Masked Centroid validation loss, governed by an early stopping patience of 10 epochs.

\textbf{Parameter-Efficient Fine-Tuning (PEFT):} To optimize memory usage and training stability, the LLM backbones were loaded in 4-bit NormalFloat (NF4) precision. We applied Low-Rank Adaptation (LoRA) \cite{hu2021loralowrankadaptationlarge} to the pre-trained LLMs, injecting trainable decomposition matrices (rank $r \in \{8, 16\}$, $\alpha \in \{16, 32\}$, dropout $\in \{0.1, 0.3\}$) into all linear projection layers of the self-attention mechanism (\texttt{q\_proj, k\_proj, v\_proj, o\_proj}) and the feed-forward network (\texttt{gate\_proj, up\_proj, down\_proj}).

\subsection{Task-Specific Optimization and Loss Functions}
All models were optimized using the 8-bit AdamW optimizer with gradient checkpointing and Flash Attention 2 enabled to maximize throughput. To ensure robustness to modality missingness in the sequential models, we applied stochastic modality dropout ($p=0.20$) exclusively during training.

The objective functions were tailored to the downstream tasks:
\begin{itemize}
    \item \textbf{Length of Stay (eICU):} Modeled as a continuous regression task. Due to the right-skewed nature of ICU stays, targets were transformed via $\log(1 + y)$. The network was optimized using Mean Squared Error (MSE) loss.
    \item \textbf{Phenotyping (MIMIC-IV):} Modeled as 25-class multi-label classification using weighted BCEWithLogitsLoss, where class-specific positive weights were dynamically calculated based on the training distribution.
    \item \textbf{In-Hospital Mortality (both datasets):} Modeled as binary classification and optimized using Binary Cross-Entropy with Logits (BCEWithLogitsLoss), with a positive weight factor applied to counteract severe class imbalance.
\end{itemize}

\clearpage

\section{Additional Figures and Results}
\label{sec:additional}    

\begin{figure*}[htbp]
    \centering
    \begin{subfigure}{\linewidth}
        \centering
        \includegraphics[width=0.75\linewidth]{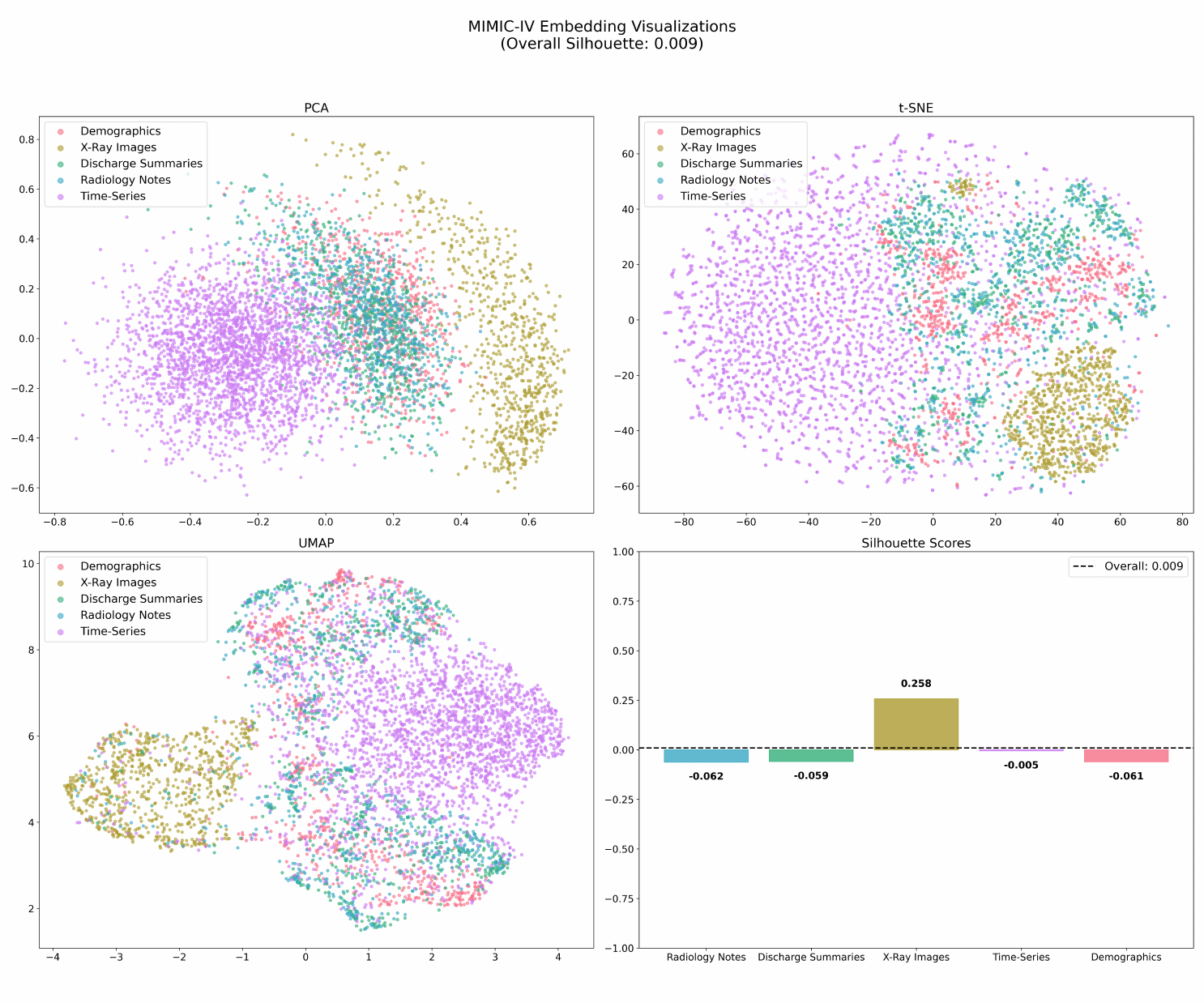}
        \caption{MIMIC-IV}
        \label{fig:mimic_latents}
    \end{subfigure}
    
    \vspace{1.5em} % Adds breathing room between the top and bottom plots
    
    \begin{subfigure}{\linewidth}
        \centering
        \includegraphics[width=0.75\linewidth]{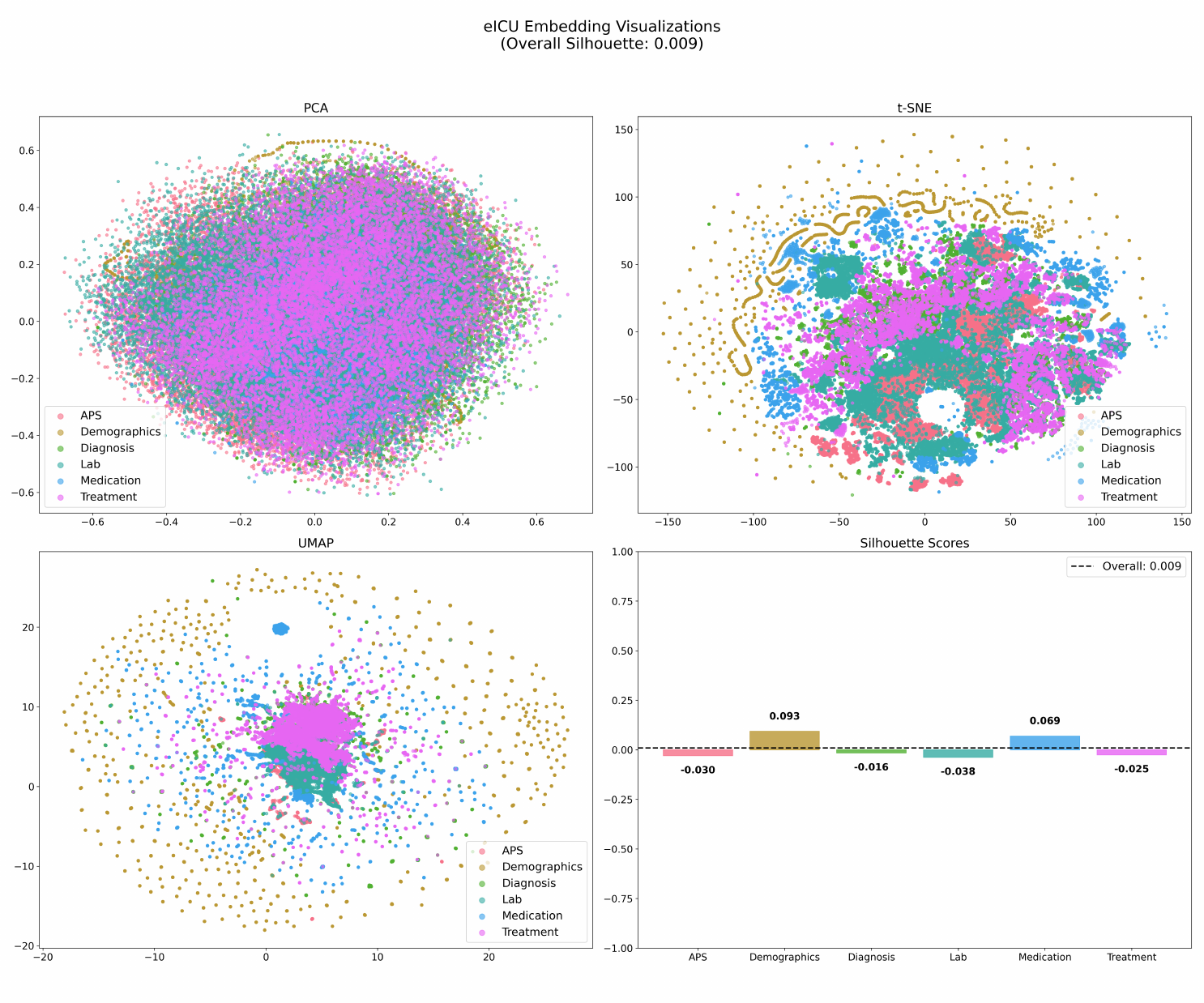}
        \caption{eICU}
        \label{fig:eICU_latents}
    \end{subfigure}
    
    \caption{Latent space embeddings for the MIMIC-IV (top) and eICU (bottom) datasets. We observe the same trends of qualitatively and quantitatively well integrated latent modality representations.} 
    \label{fig:combined_latents}
\end{figure*}

\begin{figure*}[htbp]
    \centering
    \begin{subfigure}{0.515\textwidth}
        \centering
        \includegraphics[width=\linewidth]{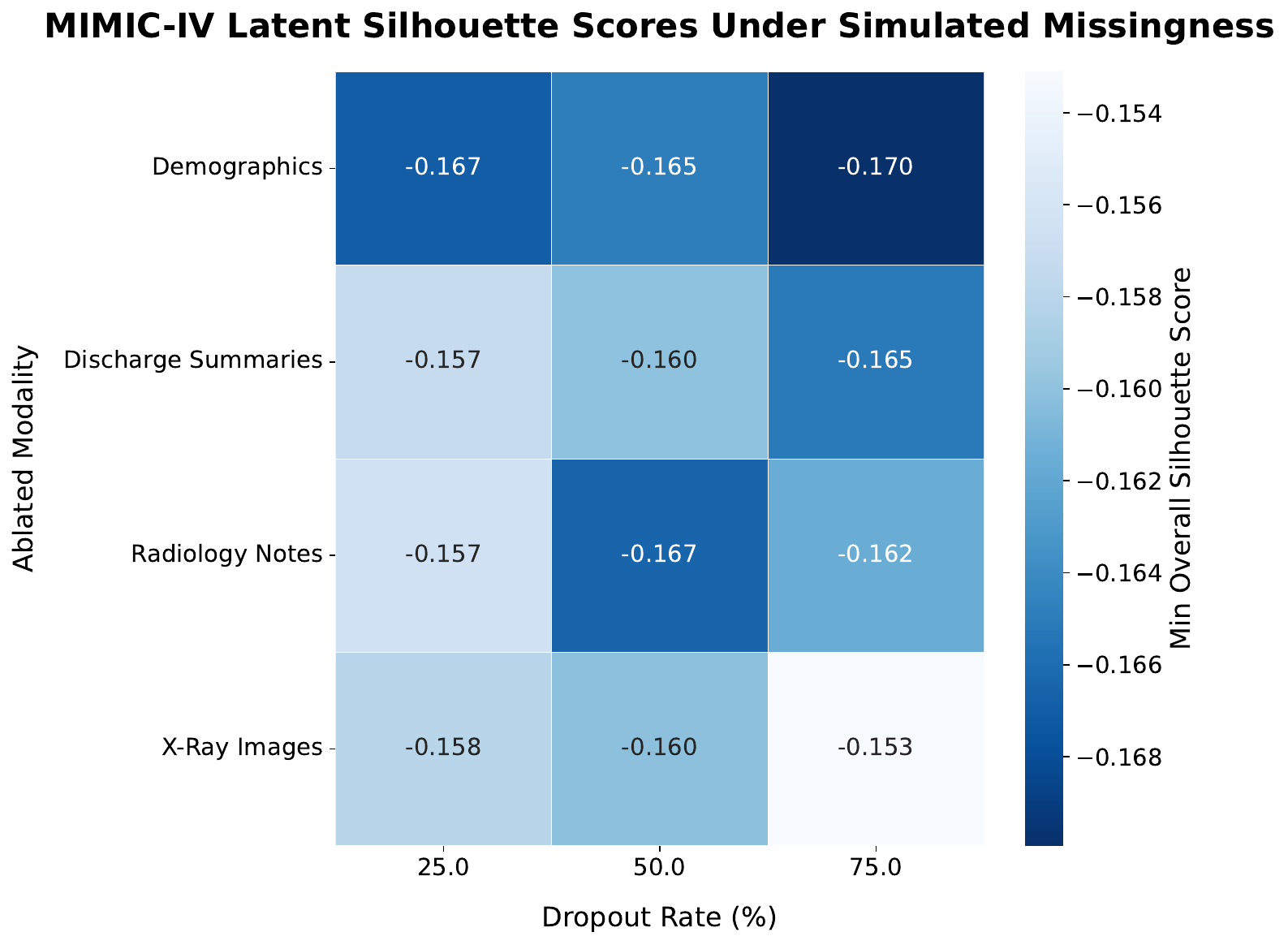}
        \caption{MIMIC-IV}
        \label{fig:mimic_ablations}
    \end{subfigure}\hfill
    \begin{subfigure}{0.4775\textwidth}
        \centering
        \includegraphics[width=\linewidth]{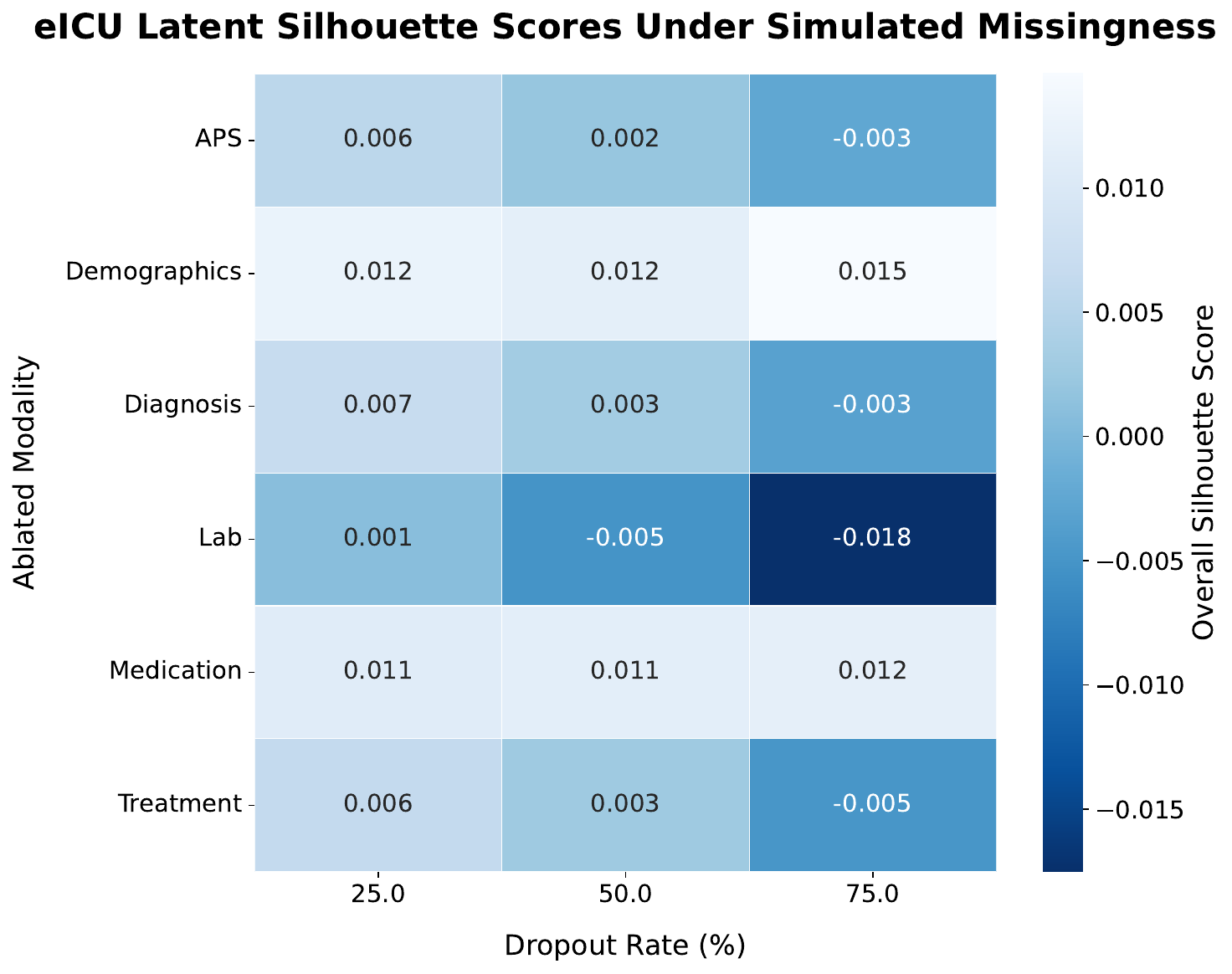}
        \caption{eICU}
        \label{fig:eicu_ablations}
    \end{subfigure}
    \caption{Latent space degradation under simulated modality missingness during pre-training for the MIMIC-IV (a) and eICU (b) datasets.  We simulate the missingness scenarios by artificially dropping a given proportion of a specific modality from the dataset.  The heatmaps display the overall Silhouette scores for models trained with varying dropout rates (x-axis) applied to individual modalities (y-axis).   A lower Silhouette score indicates a more severe disruption of the harmonized latent space.  As expected, higher dropout rates generally lead to worse clustering.} 
    \label{fig:pretraining_ablations}
\end{figure*}

\begin{table}[htbp]
    \centering
    \caption{Calibration performance (Adaptive Calibration Error and Brier Skill Score) of baseline static models and sequential transformer-based models on Mortality and Phenotyping tasks on the MIMIC-IV dataset. ACE measures the absolute difference between confidence and accuracy ($\downarrow$), while BSS measures probabilistic forecasting skill compared to a naive prevalence baseline ($\uparrow$). Best results for each metric are bolded. The results highlight a distinct architectural dynamic: while simpler static models (e.g., LSTMs) natively achieve lower ACE scores, indicating predicted probabilities that map closely to raw accuracy, sequential LLMs dominate in BSS when initialized with contrastively learned embeddings. Notably, contrastive pretraining consistently transitions the LLMs from negative to highly positive BSS scores across both tasks. This demonstrates that although large language models are traditionally prone to overconfidence, contrastive alignment effectively regularizes their latent space, enabling superior probabilistic forecasting and fundamental class separation.}
    \resizebox{\textwidth}{!}{%
        \begin{tabular}{llcccc}
            \toprule
            & & \multicolumn{2}{c}{\textbf{Mortality}} & \multicolumn{2}{c}{\textbf{Phenotyping}} \\
            \cmidrule(lr){3-4} \cmidrule(lr){5-6}
            \textbf{Architecture} & \textbf{Initialization} & \textbf{ACE} ($\downarrow$) & \textbf{BSS} ($\uparrow$) & \textbf{ACE} ($\downarrow$) & \textbf{BSS} ($\uparrow$) \\
            \midrule
            Multi-layer Perceptron & Scratch & 0.0723 & -0.0003 & 0.0360 & 0.0017 \\
            LSTM & Scratch & \textbf{0.0328} & -0.0013 & 0.0288 & 0.0795 \\
            Multi-layer Perceptron & Contrastive & 0.0378 & 0.0112 & 0.0288 & 0.0105 \\
            LSTM & Contrastive & 0.0455 & -0.0008 & \textbf{0.0222} & 0.0141 \\
            \midrule
            BioMistral-7B & Scratch & 0.1277 & -0.0439 & 0.1447 & -0.0512 \\
            DeepSeek-LLM-7B-Base & Scratch & 0.1366 & -0.1015 & 0.1363 & -0.0390 \\
            Meditron-7B & Scratch & 0.1544 & -0.2715 & 0.1450 & -0.0390 \\
            Meta-Llama-3-8B & Scratch & 0.1175 & -0.0971 & 0.1565 & -0.0385 \\
            Phi-3-Mini-4K-Instruct & Scratch & 0.2012 & -0.0671 & 0.1424 & -0.0392 \\
            Mistral-7B-v0.1 & Scratch & 0.1824 & -0.1694 & 0.1500 & -0.0387 \\
            BioMistral-7B & Contrastive & 0.1780 & 0.0361 & 0.1346 & 0.0886 \\
            DeepSeek-LLM-7B-Base & Contrastive & 0.1351 & \textbf{0.0524} & 0.1434 & 0.0518 \\
            Meditron-7B & Contrastive & 0.0809 & -0.0565 & 0.1464 & 0.0549 \\
            Meta-Llama-3-8B & Contrastive & 0.1466 & 0.0078 & 0.1319 & 0.0814 \\
            Phi-3-Mini-4K-Instruct & Contrastive & 0.0999 & 0.0059 & 0.1350 & \textbf{0.1033} \\
            Mistral-7B-v0.1 & Contrastive & 0.1164 & 0.0232 & 0.1385 & 0.0974 \\
            \bottomrule
        \end{tabular}%
    }
    \label{table:mimic_calibration_metrics}
\end{table}

\begin{table}[htbp]
    \centering
    \caption{Calibration performance (Adaptive Calibration Error and Brier Skill Score) of baseline static models and sequential transformer-based models on the Mortality task on the eICU dataset. ACE measures the absolute difference between confidence and accuracy ($\downarrow$), while BSS measures probabilistic forecasting skill compared to a naive prevalence baseline ($\uparrow$). Best results for each metric are bolded. The results highlight a distinct architectural dynamic: while simpler static models (e.g., MLPs) natively achieve lower ACE scores—indicating predicted probabilities that map closely to raw accuracy—sequential LLMs dominate in overall probabilistic forecasting skill (BSS). However, unlike on MIMIC-IV, end-to-end (Scratch) training natively achieves the highest BSS scores on eICU, surpassing the contrastively aligned models. This suggests that for this specific dataset and task, direct end-to-end optimization leads to superior probabilistic separation.}
    \resizebox{0.75\textwidth}{!}{%
        \begin{tabular}{llcc}
            \toprule
            & & \multicolumn{2}{c}{\textbf{Mortality}} \\
            \cmidrule(lr){3-4}
            \textbf{Architecture} & \textbf{Initialization} & \textbf{ACE} ($\downarrow$) & \textbf{BSS} ($\uparrow$) \\
            \midrule
            Multi-layer Perceptron & Scratch & \textbf{0.0059} & 0.3178 \\
            LSTM & Scratch & 0.0077 & 0.3107 \\
            Multi-layer Perceptron & Contrastive & 0.0082 & 0.3156 \\
            LSTM & Contrastive & 0.0076 & 0.3079 \\
            \midrule
            BioMistral-7B & Scratch & 0.0126 & 0.3606 \\
            DeepSeek-LLM-7B-Base & Scratch & 0.0096 & \textbf{0.3662} \\
            Meditron-7B & Scratch & 0.0132 & 0.3449 \\
            Meta-Llama-3-8B & Scratch & 0.0140 & 0.3612 \\
            Phi-3-Mini-4K-Instruct & Scratch & 0.0130 & 0.3253 \\
            Mistral-7B-v0.1 & Scratch & 0.0119 & 0.3495 \\
            BioMistral-7B & Contrastive & 0.0164 & 0.2930 \\
            DeepSeek-LLM-7B-Base & Contrastive & 0.0218 & 0.3033 \\
            Meditron-7B & Contrastive & 0.0190 & 0.3192 \\
            Meta-Llama-3-8B & Contrastive & 0.0170 & 0.3013 \\
            Phi-3-Mini-4K-Instruct & Contrastive & 0.0160 & 0.3179 \\
            Mistral-7B-v0.1 & Contrastive & 0.0200 & 0.3246 \\
            \bottomrule
        \end{tabular}%
    }
    \label{table:eicu_calibration_metrics}
\end{table}

\begin{figure}[htbp]
    \centering
    % Row 1: E2E Baseline
    \begin{subfigure}{0.32\textwidth}
        \centering
        \includegraphics[width=\linewidth]{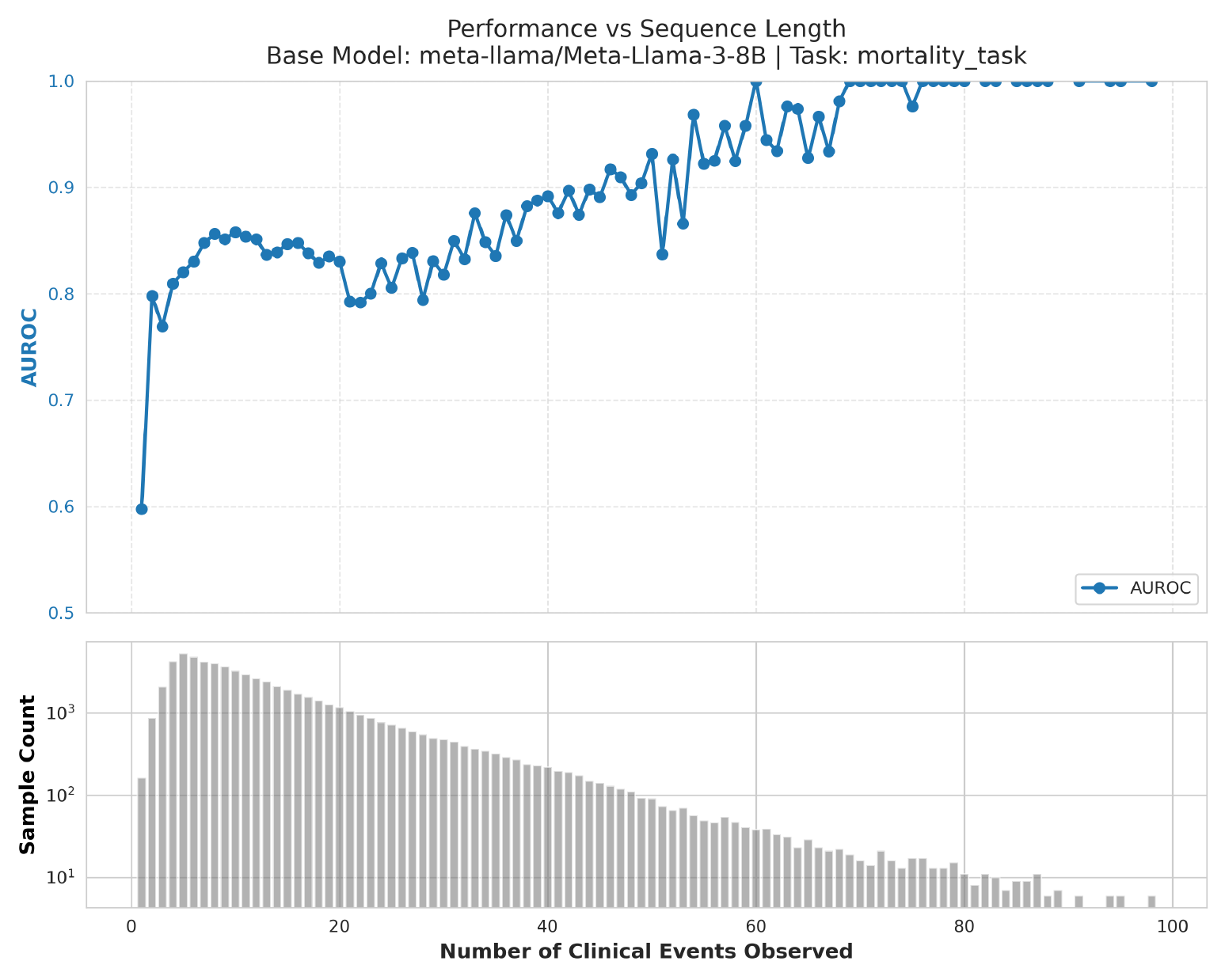}
        \caption{Non-Contrastive (Llama-3)}
    \end{subfigure}\hfill
    \begin{subfigure}{0.32\textwidth}
        \centering
        \includegraphics[width=\linewidth]{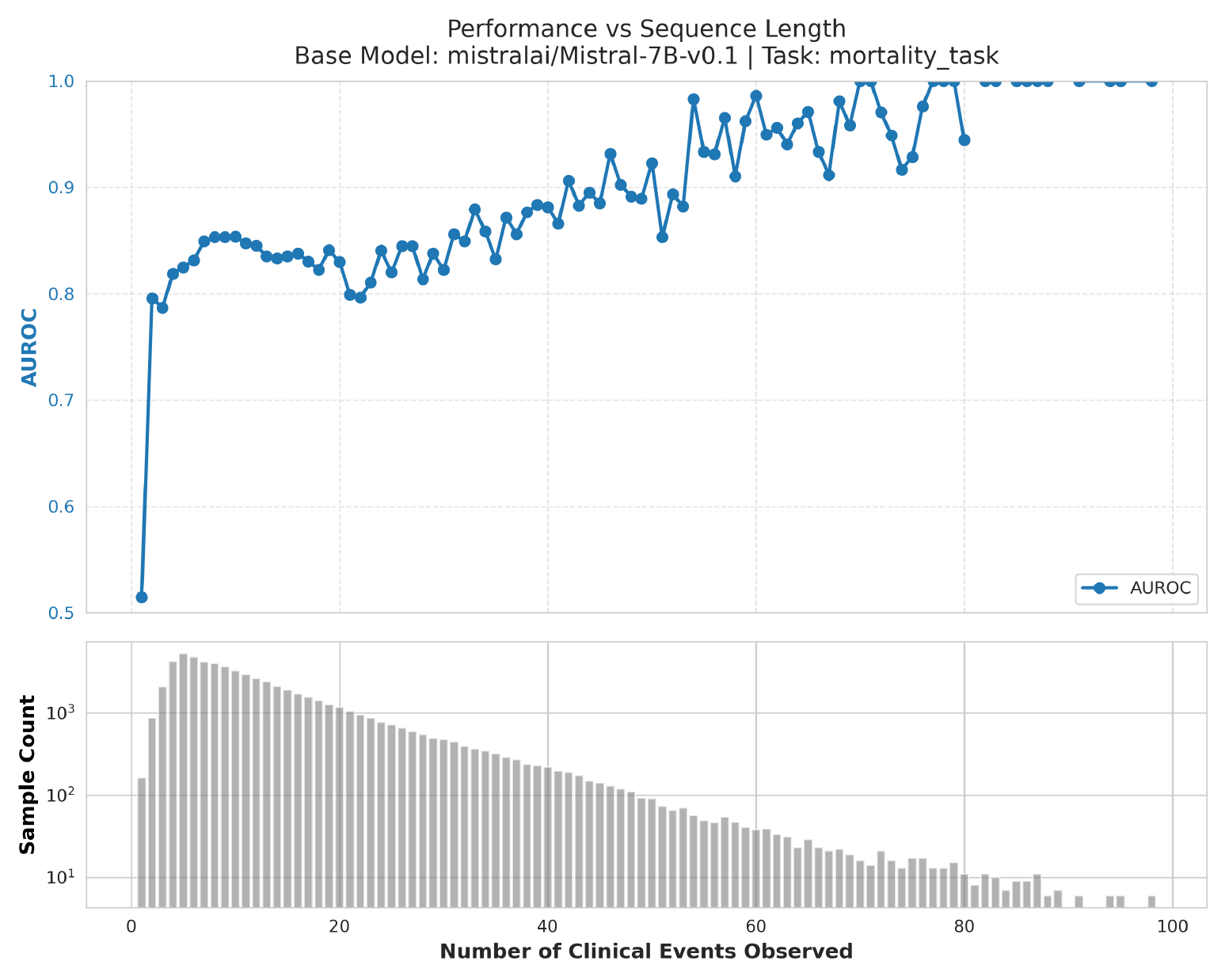}
        \caption{Non-Contrastive (Mistral)}
    \end{subfigure}\hfill
    \begin{subfigure}{0.32\textwidth}
        \centering
        \includegraphics[width=\linewidth]{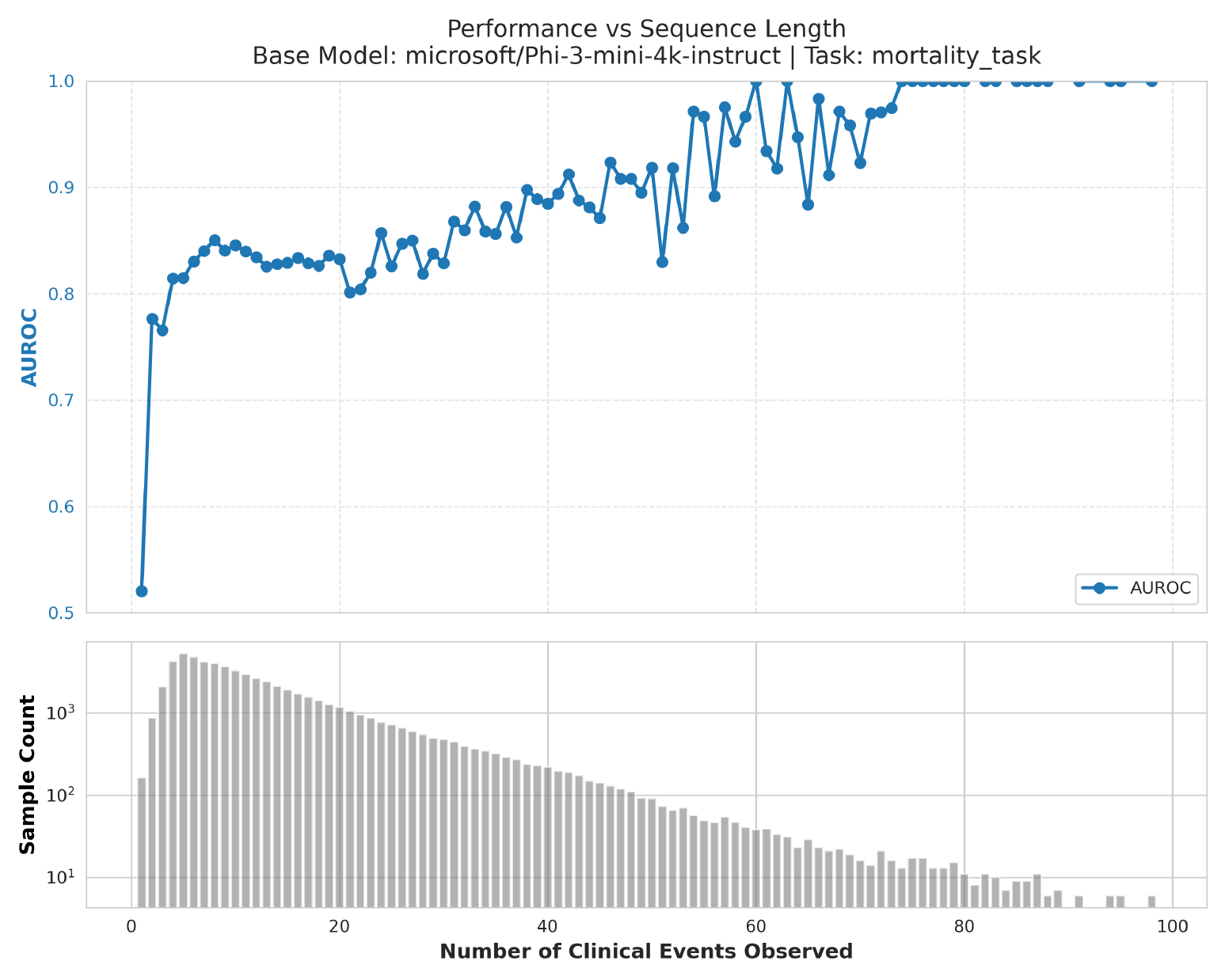}
        \caption{Non-Contrastive (Phi-3)}
    \end{subfigure}
    
    \vspace{1em}
    
    % Row 2: Contrastive
    \begin{subfigure}{0.32\textwidth}
        \centering
        \includegraphics[width=\linewidth]{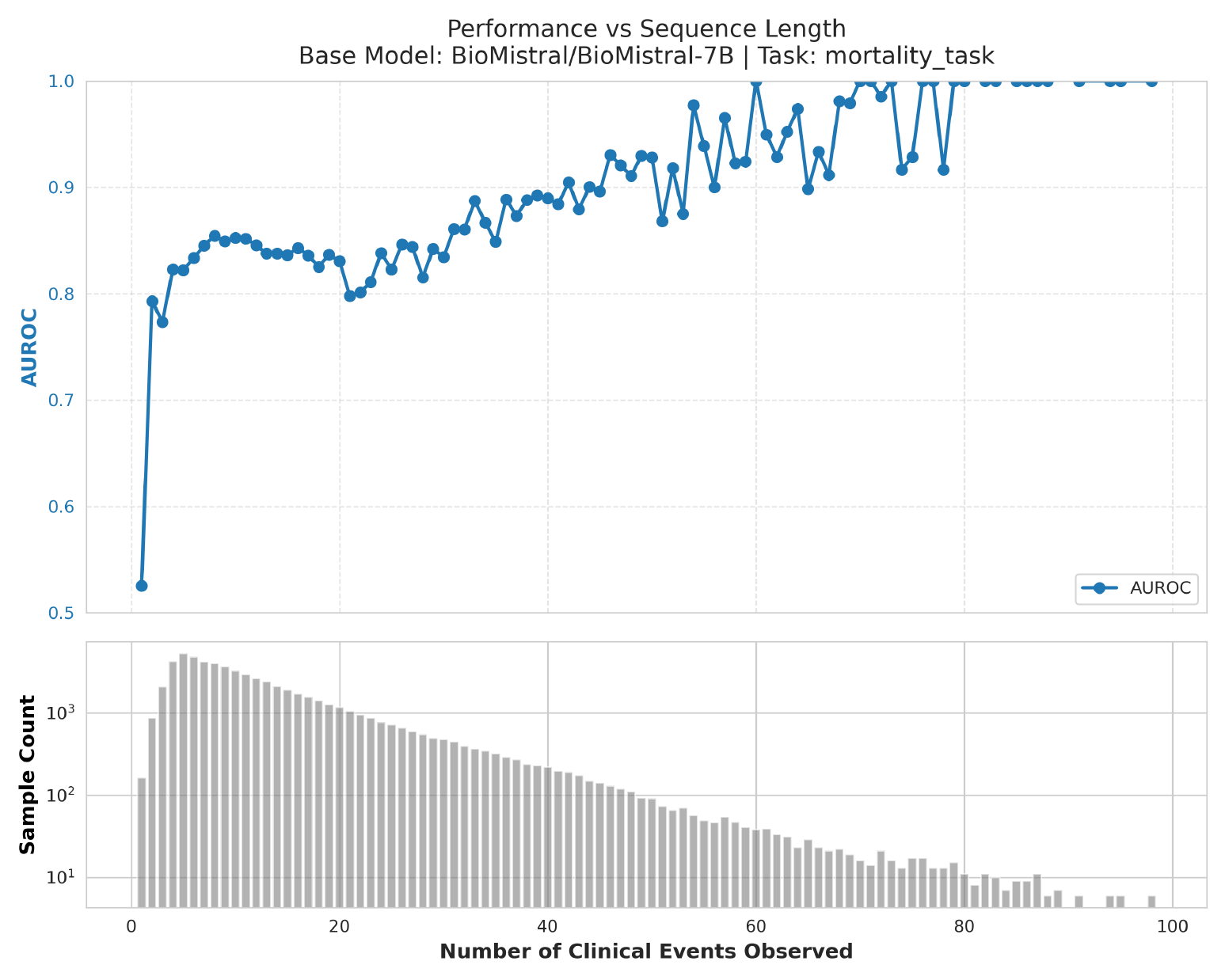}
        \caption{Non-Contrastive (BioMistral)}
    \end{subfigure}\hfill
    \begin{subfigure}{0.32\textwidth}
        \centering
        \includegraphics[width=\linewidth]{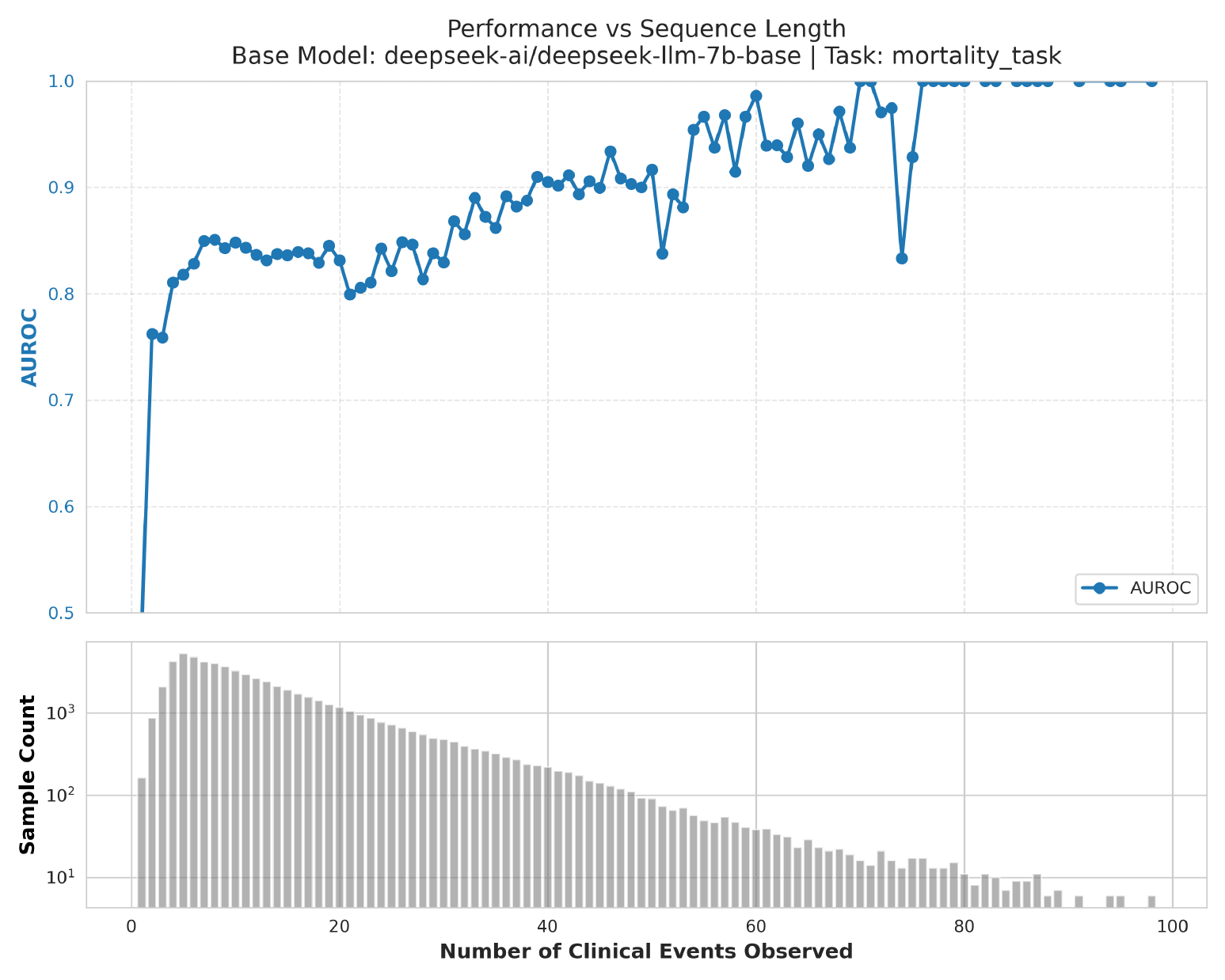}
        \caption{Non-Contrastive (Deepseek)}
    \end{subfigure}\hfill
    \begin{subfigure}{0.32\textwidth}
        \centering
        \includegraphics[width=\linewidth]{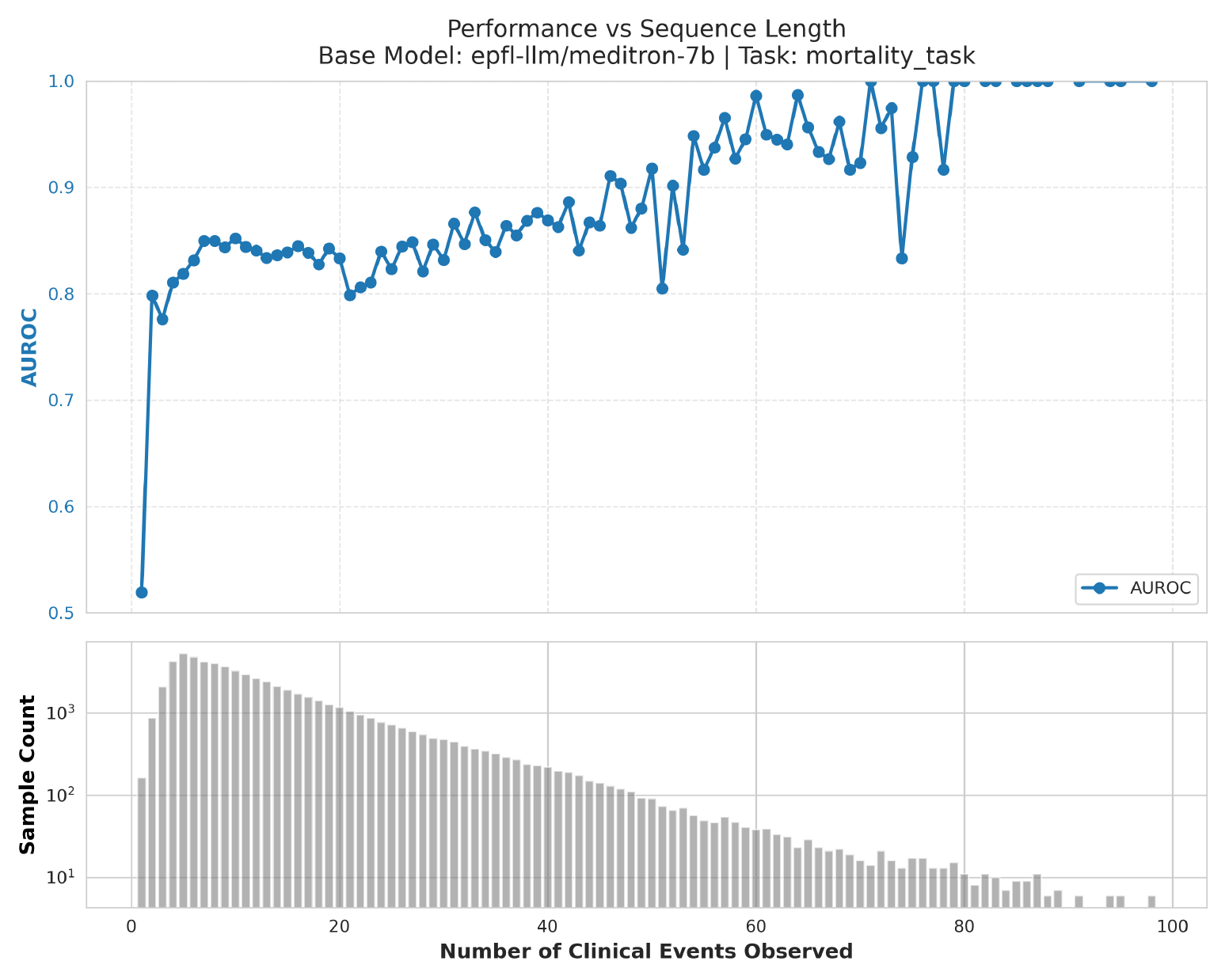}
        \caption{Non-Contrastive (Meditron)}
    \end{subfigure}
    
    \caption{Mortality performance plots for non-contrastive sequential models evaluated on eICU. The AUROC performance is plotted against the patient stays with the associated number of events in their sequences, along with a visualization of the patient stay lengths.}
    \label{fig:eicu_mortality_base_trajectory}
\end{figure}

\begin{figure}[htbp]
    \centering
    % Row 1: E2E Baseline
    \begin{subfigure}{0.32\textwidth}
        \centering
        \includegraphics[width=\linewidth]{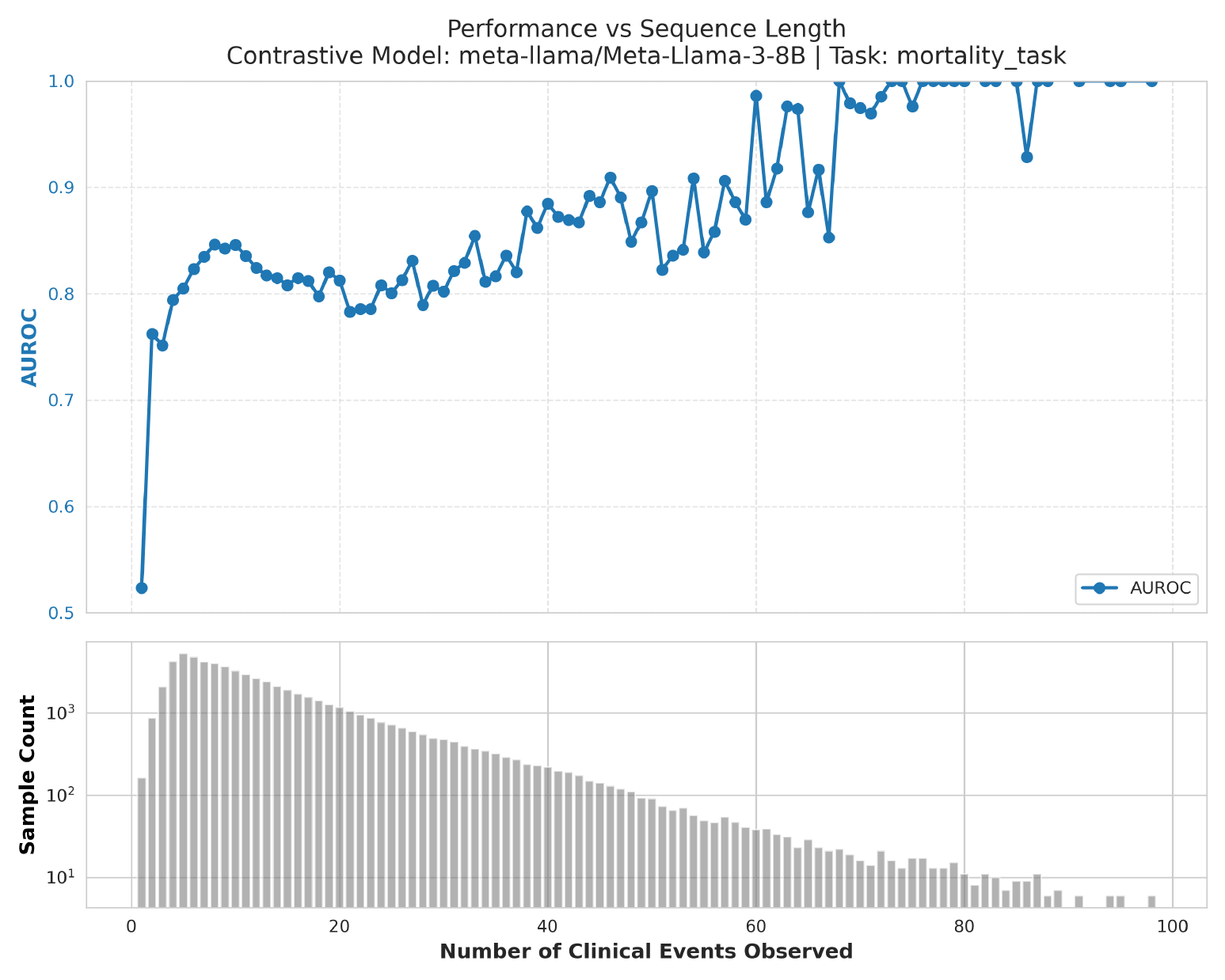}
        \caption{Contrastive (Llama-3)}
    \end{subfigure}\hfill
    \begin{subfigure}{0.32\textwidth}
        \centering
        \includegraphics[width=\linewidth]{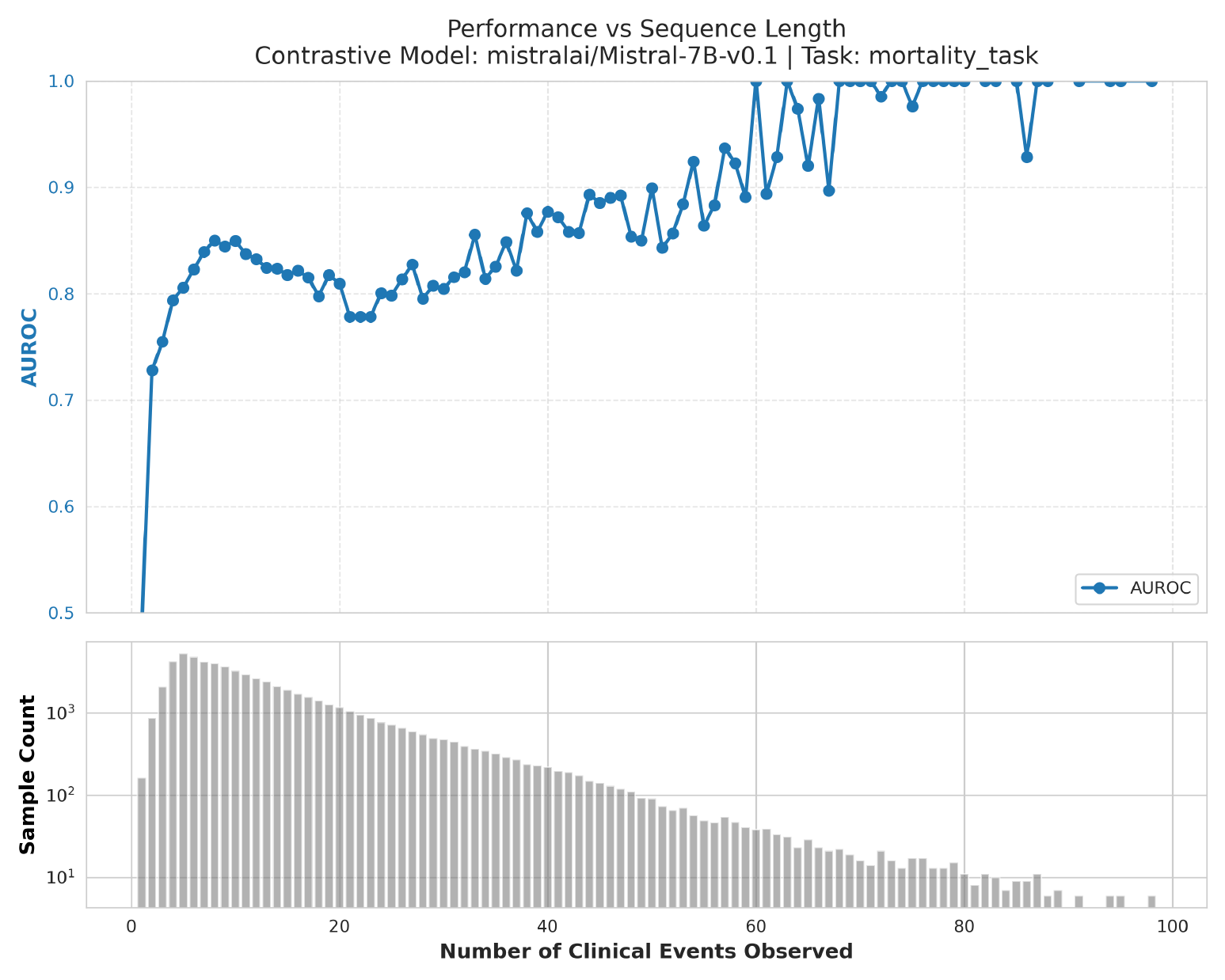}
        \caption{Contrastive (Mistral)}
    \end{subfigure}\hfill
    \begin{subfigure}{0.32\textwidth}
        \centering
        \includegraphics[width=\linewidth]{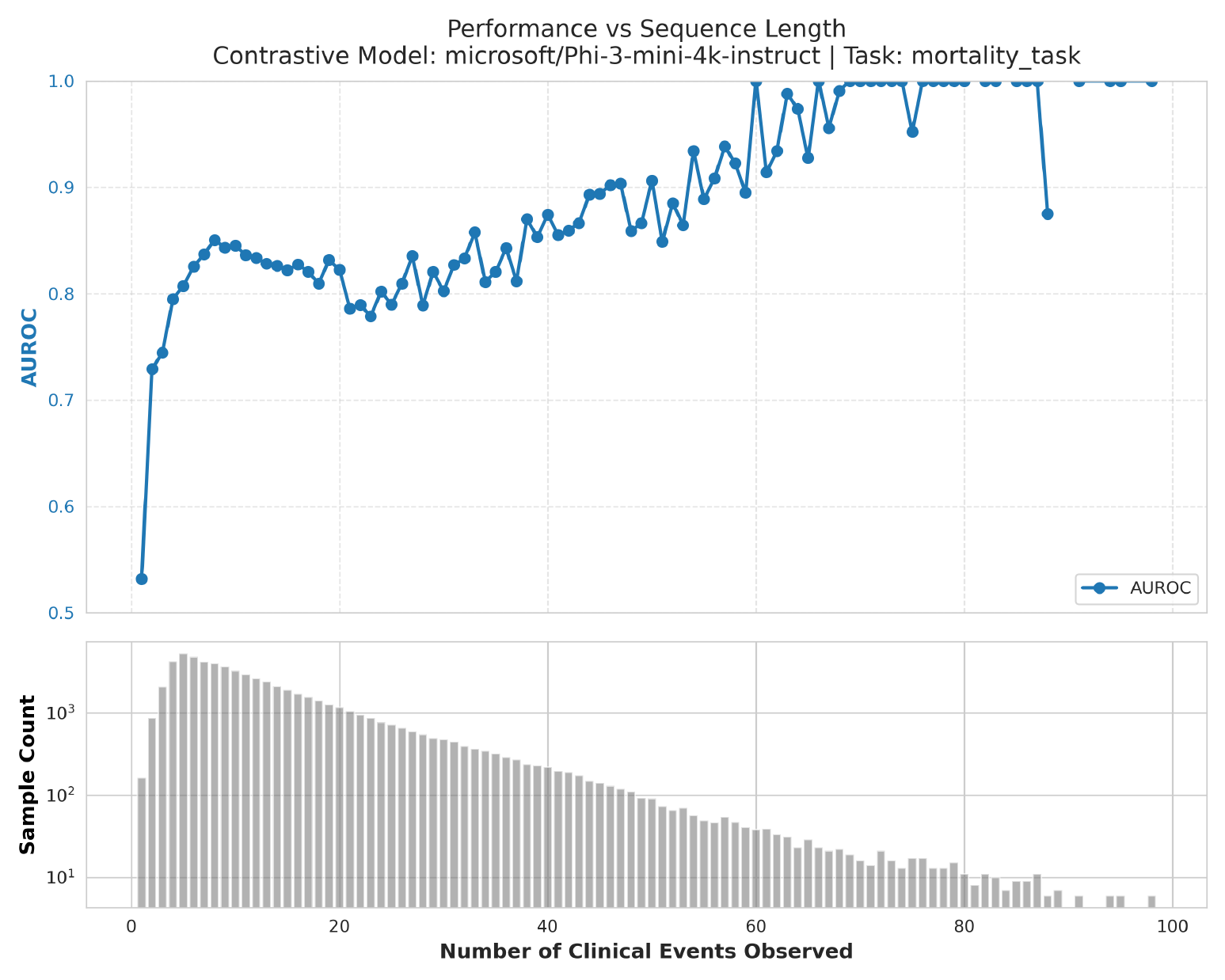}
        \caption{Contrastive (Phi-3)}
    \end{subfigure}
    
    \vspace{1em}
    
    % Row 2: Contrastive
    \begin{subfigure}{0.32\textwidth}
        \centering
        \includegraphics[width=\linewidth]{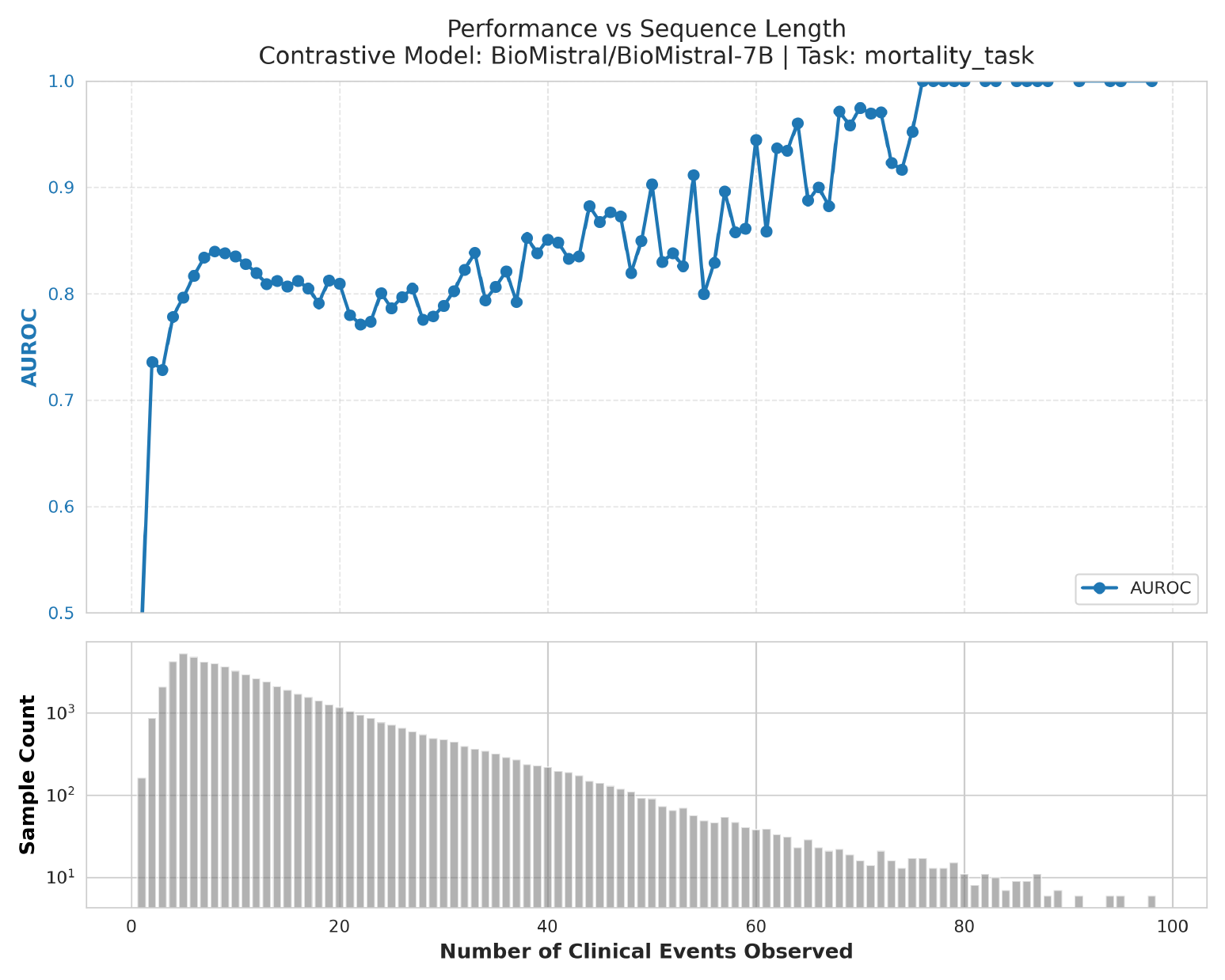}
        \caption{Contrastive (BioMistral)}
    \end{subfigure}\hfill
    \begin{subfigure}{0.32\textwidth}
        \centering
        \includegraphics[width=\linewidth]{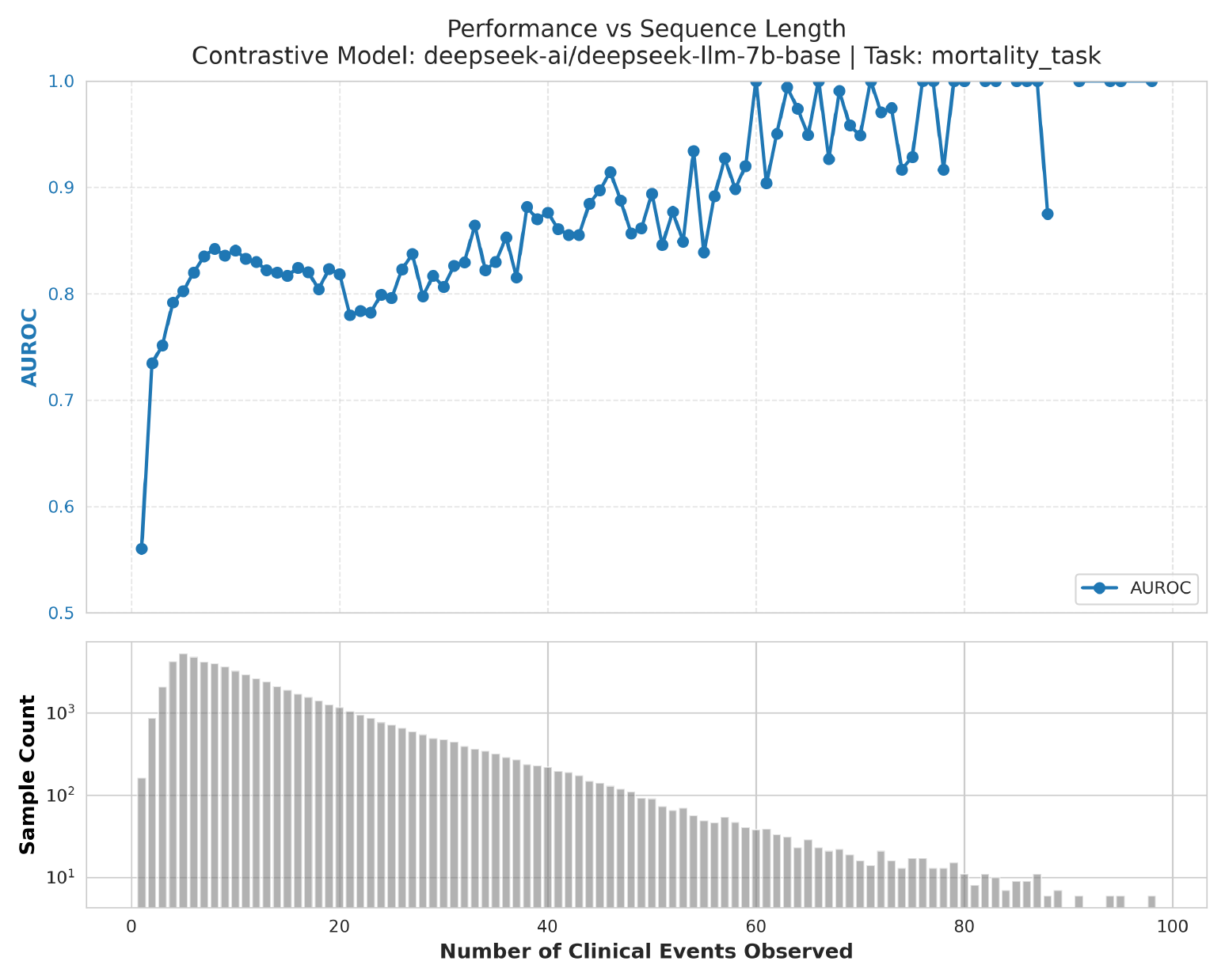}
        \caption{Contrastive (Deepseek)}
    \end{subfigure}\hfill
    \begin{subfigure}{0.32\textwidth}
        \centering
        \includegraphics[width=\linewidth]{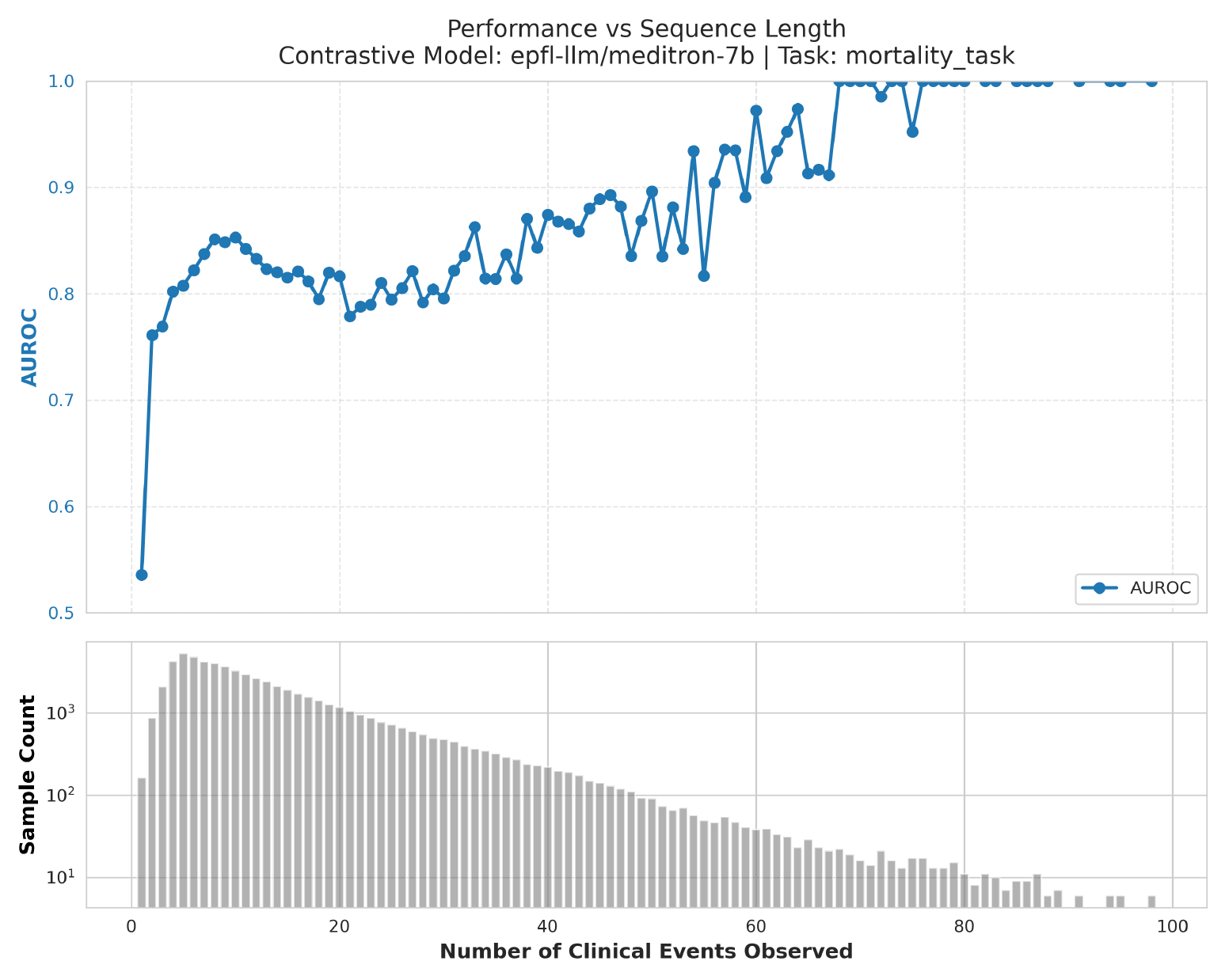}
        \caption{Contrastive (Meditron)}
    \end{subfigure}
    
    \caption{Mortality performance plots for contrastive sequential models evaluated on eICU. The AUROC performance is plotted against the patient stays with the associated number of events in their sequences, along with a visualization of the patient stay lengths.}
    \label{fig:eicu_mortality_trajectory}
\end{figure}

\begin{figure}[htbp]
    \centering
    % Row 1: E2E Baseline
    \begin{subfigure}{0.32\textwidth}
        \centering
        \includegraphics[width=\linewidth]{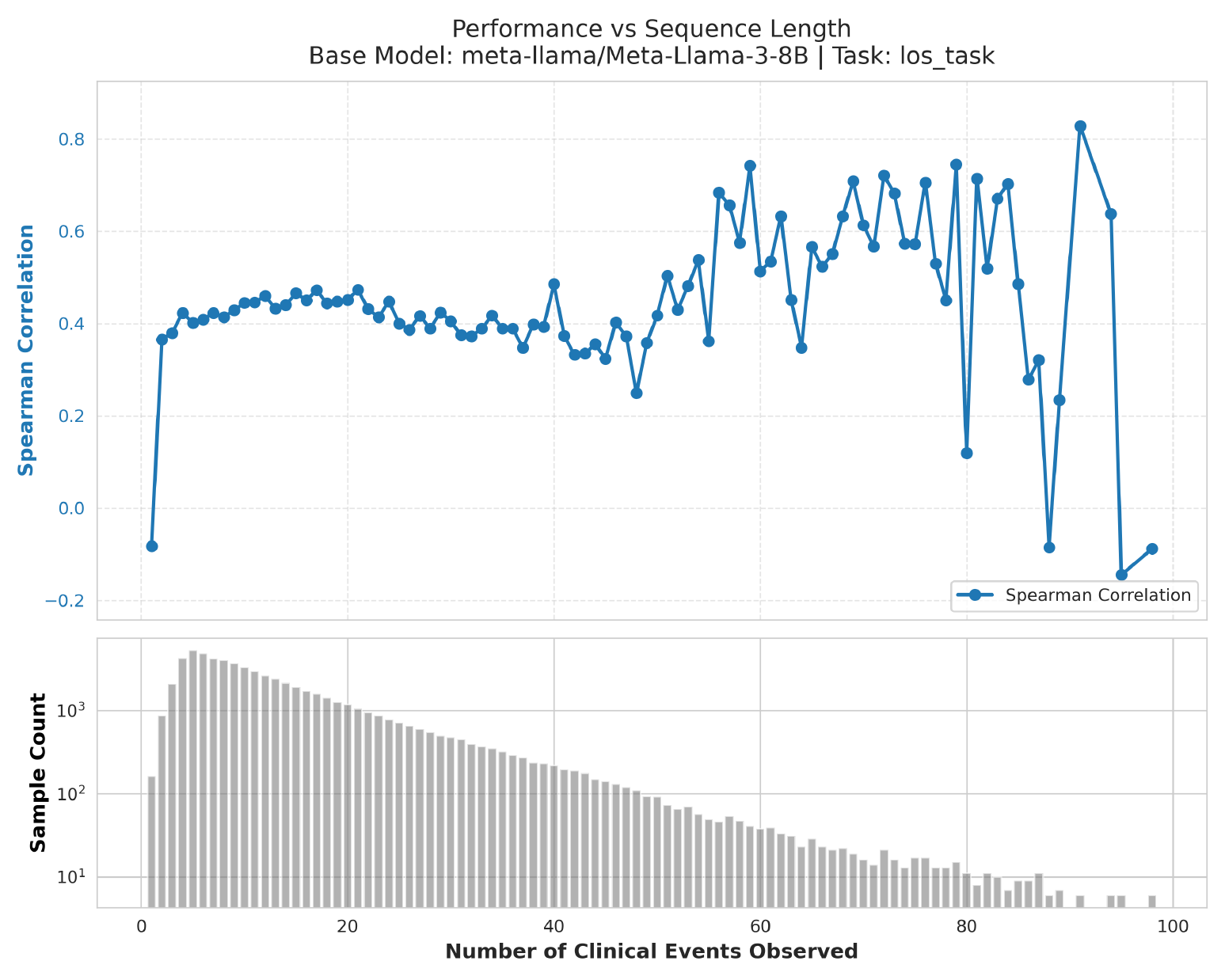}
        \caption{Non-Contrastive (Llama-3)}
    \end{subfigure}\hfill
    \begin{subfigure}{0.32\textwidth}
        \centering
        \includegraphics[width=\linewidth]{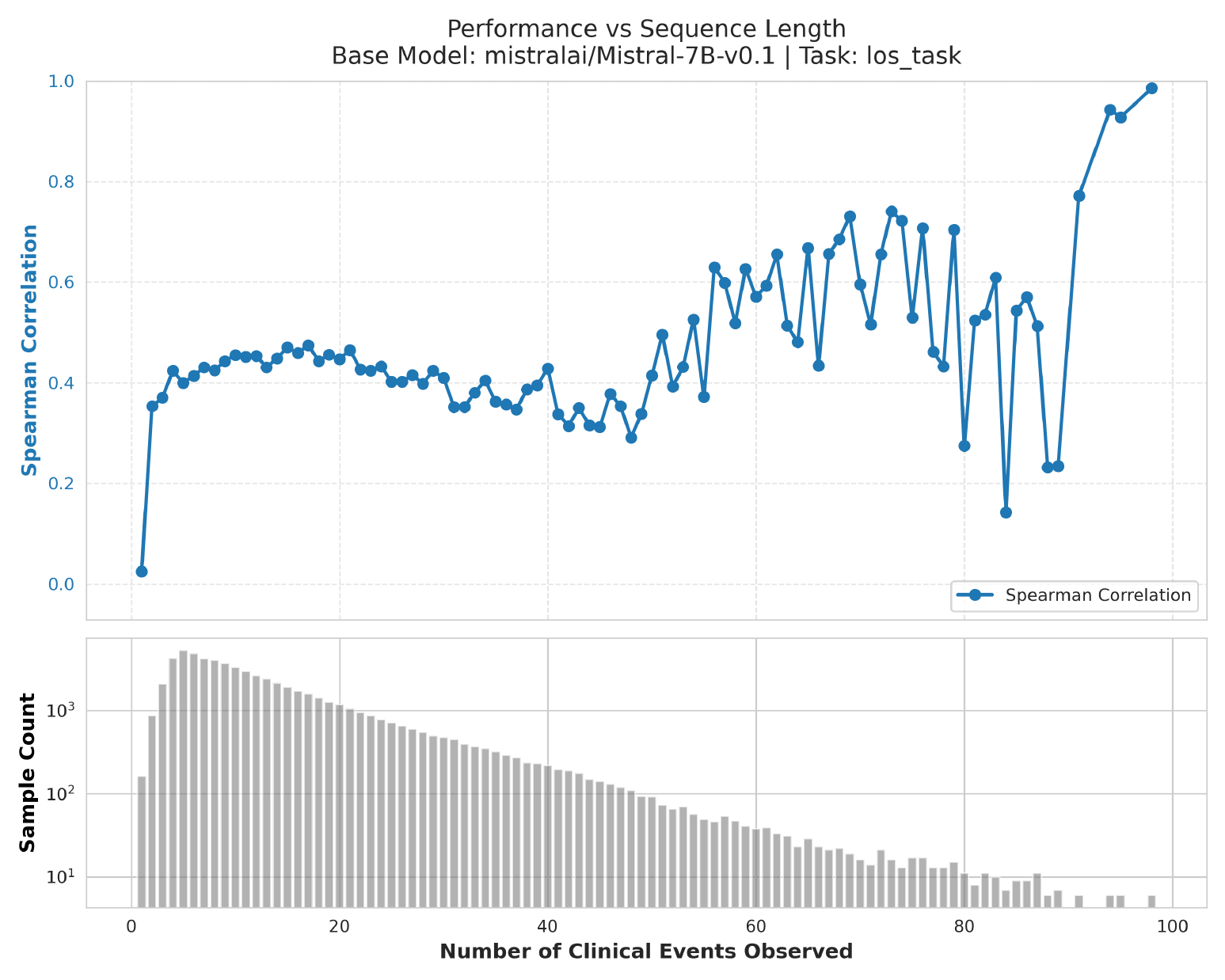}
        \caption{Non-Contrastive (Mistral)}
    \end{subfigure}\hfill
    \begin{subfigure}{0.32\textwidth}
        \centering
        \includegraphics[width=\linewidth]{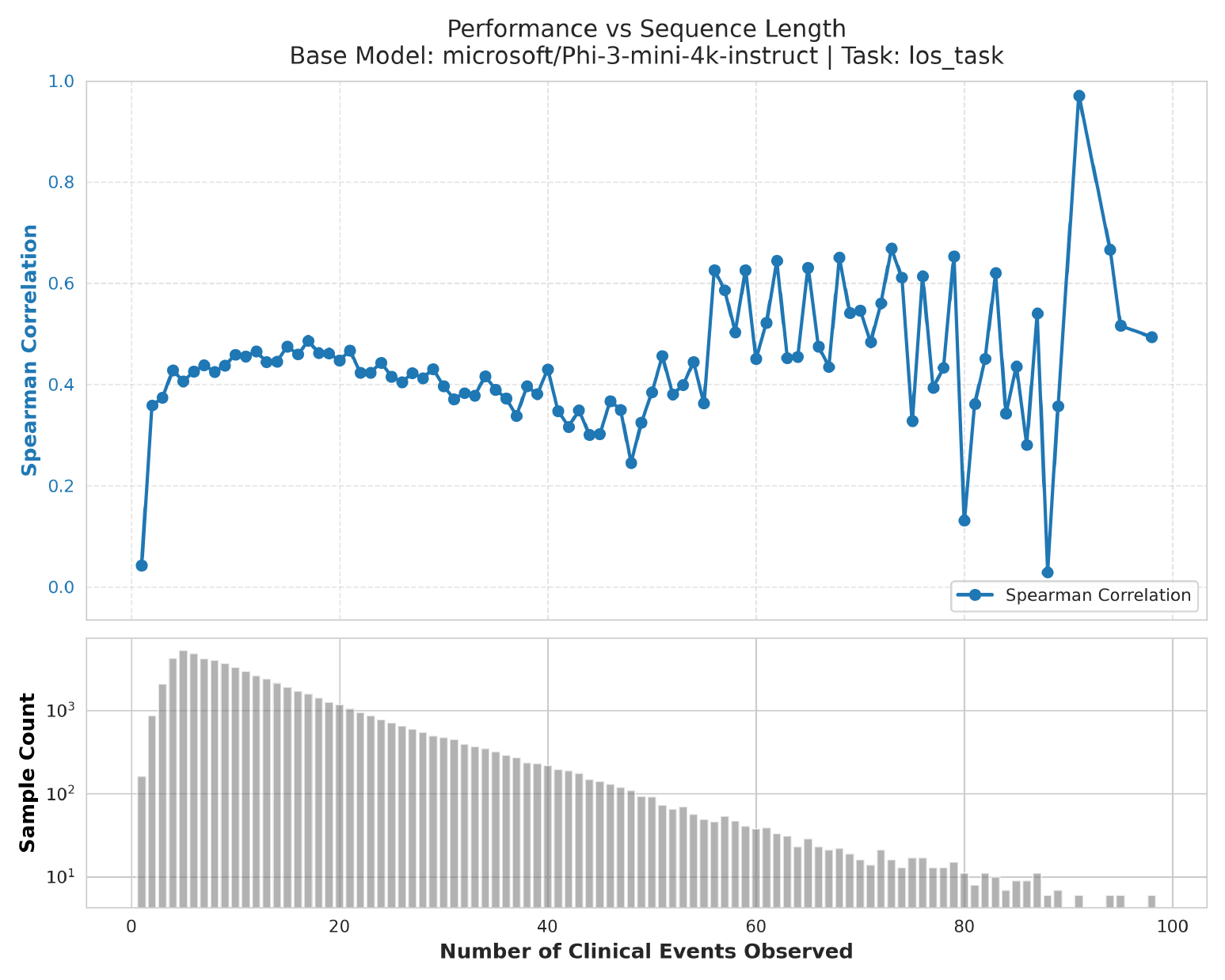}
        \caption{Non-Contrastive (Phi-3)}
    \end{subfigure}
    
    \vspace{1em}
    
    % Row 2: Contrastive
    \begin{subfigure}{0.32\textwidth}
        \centering
        \includegraphics[width=\linewidth]{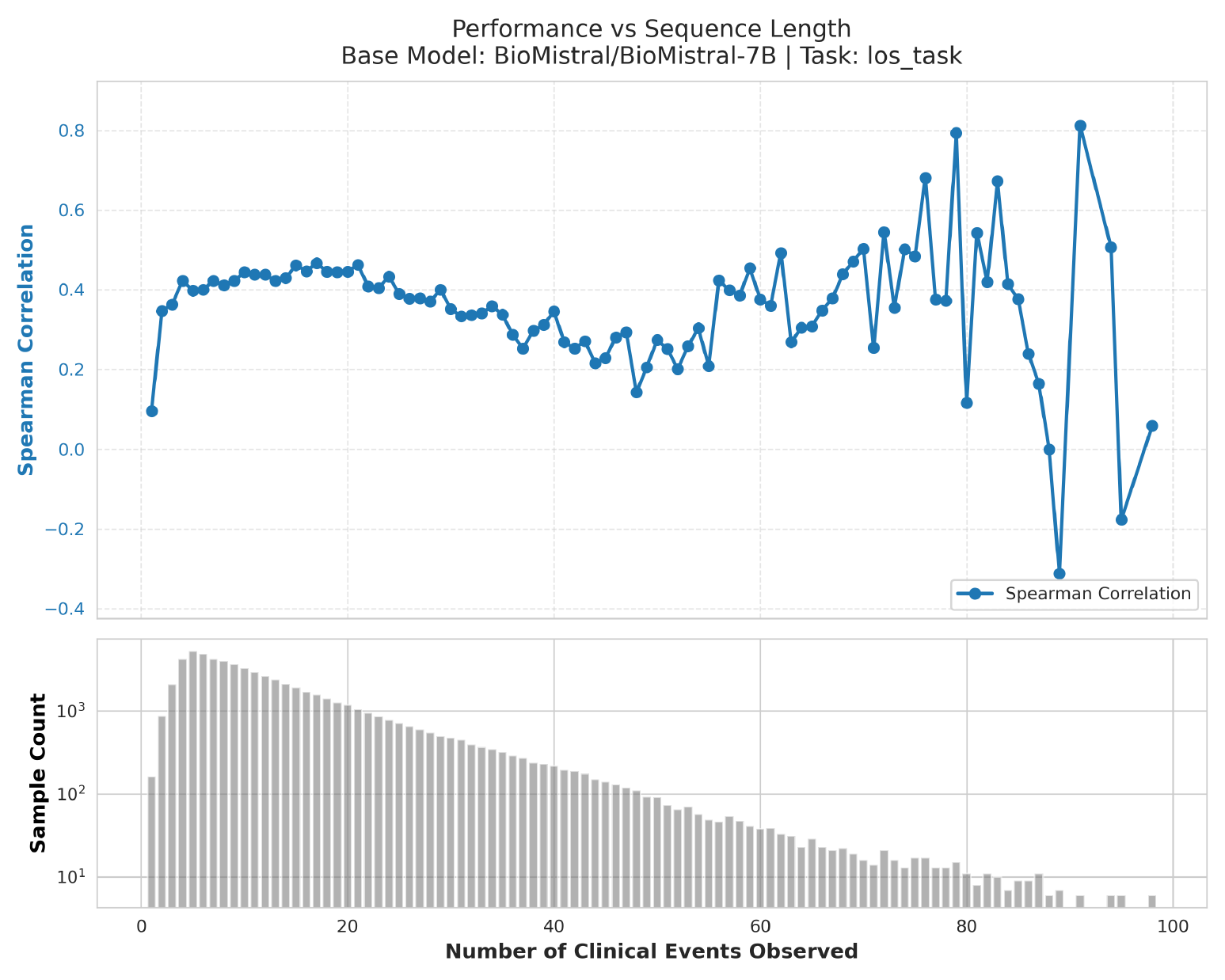}
        \caption{Non-Contrastive (BioMistral)}
    \end{subfigure}\hfill
    \begin{subfigure}{0.32\textwidth}
        \centering
        \includegraphics[width=\linewidth]{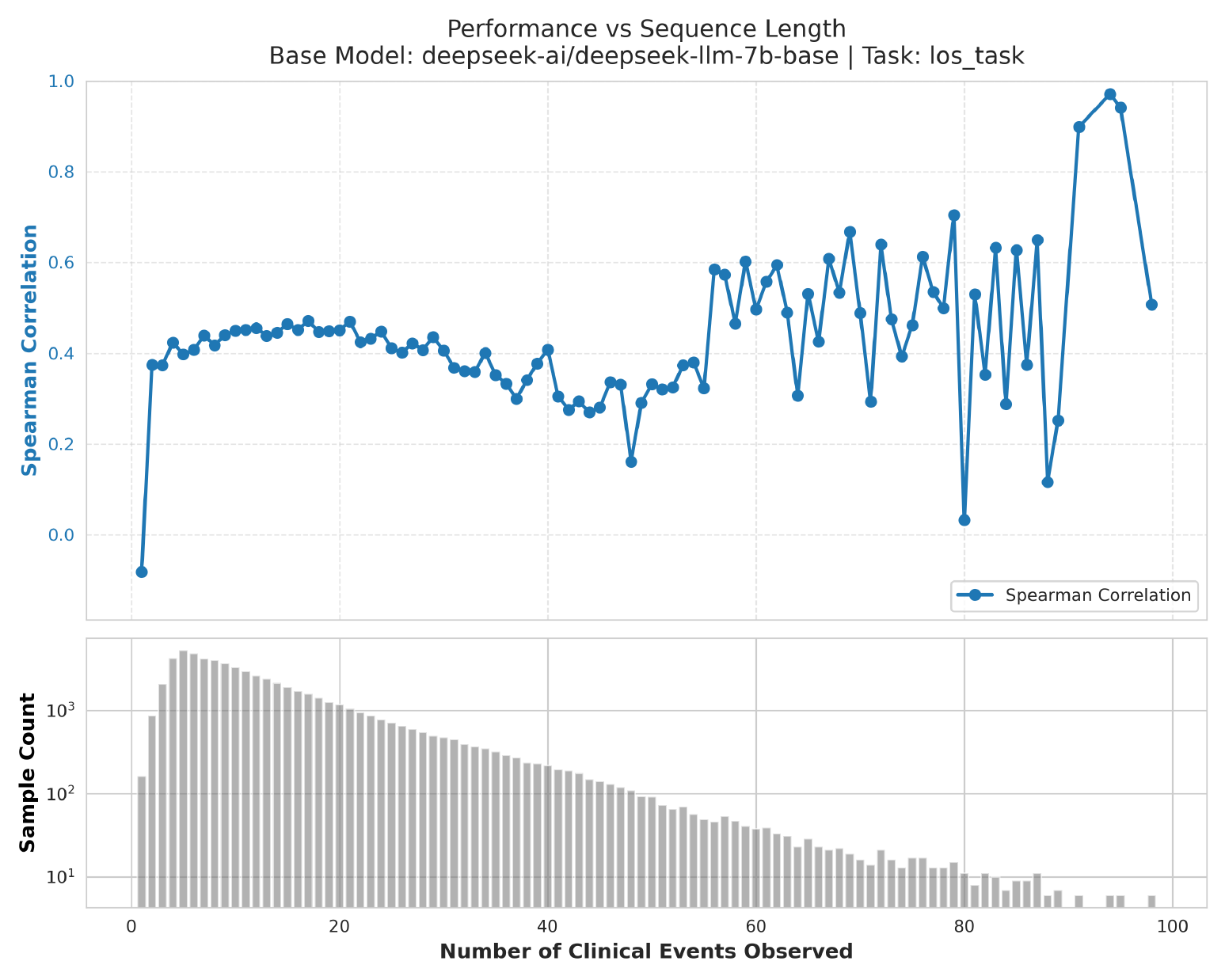}
        \caption{Non-Contrastive (Deepseek)}
    \end{subfigure}\hfill
    \begin{subfigure}{0.32\textwidth}
        \centering
        \includegraphics[width=\linewidth]{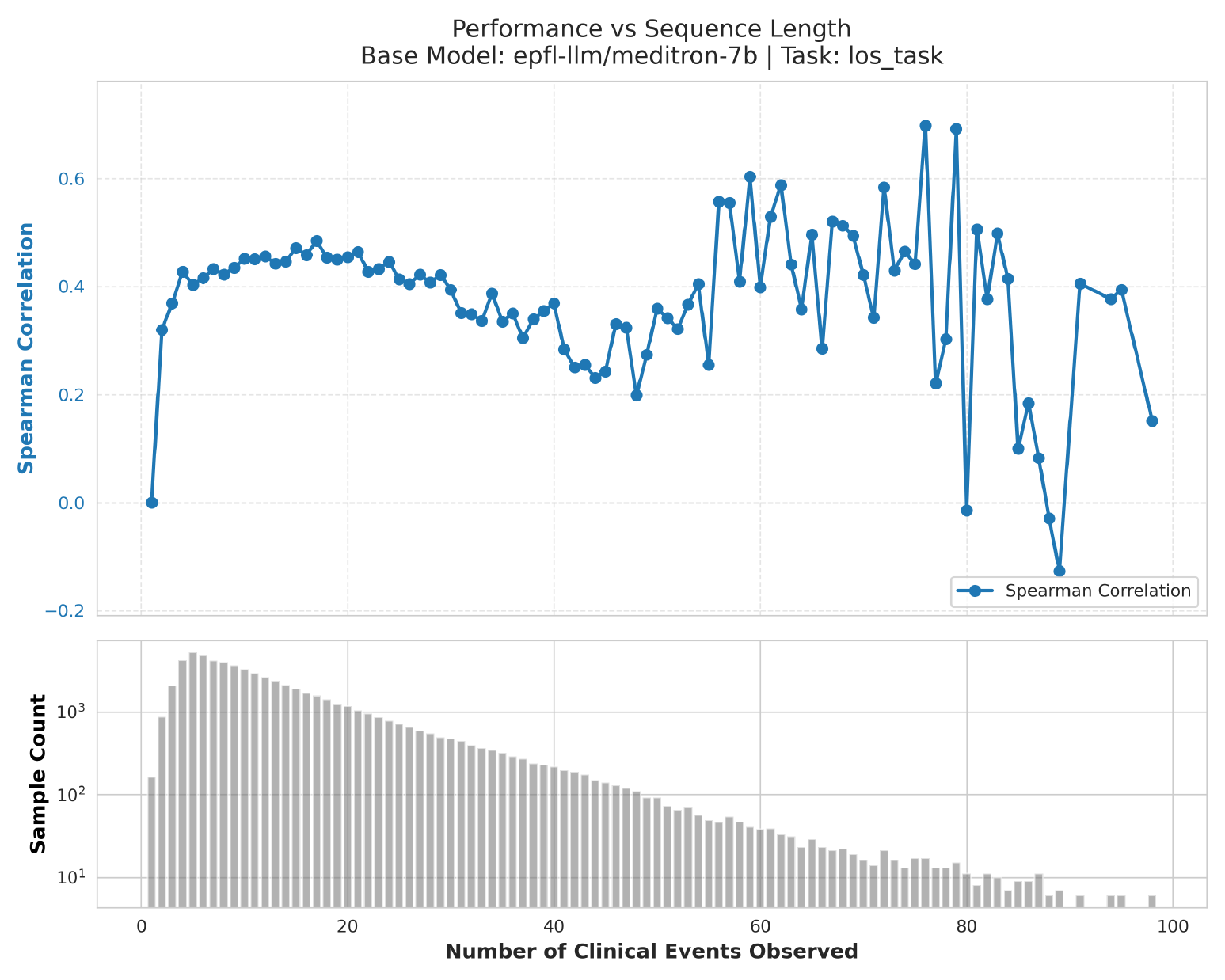}
        \caption{Non-Contrastive (Meditron)}
    \end{subfigure}
    
    \caption{Length of Stay performance plots for non-contrastive sequential models evaluated on eICU. The Spearman Correlation Coefficient is plotted against the patient stays with the associated number of events in their sequences, along with a visualization of the patient stay lengths.}
    \label{fig:eicu_los_base_trajectory}
\end{figure}

\begin{figure}[htbp]
    \centering
    % Row 1: E2E Baseline
    \begin{subfigure}{0.32\textwidth}
        \centering
        \includegraphics[width=\linewidth]{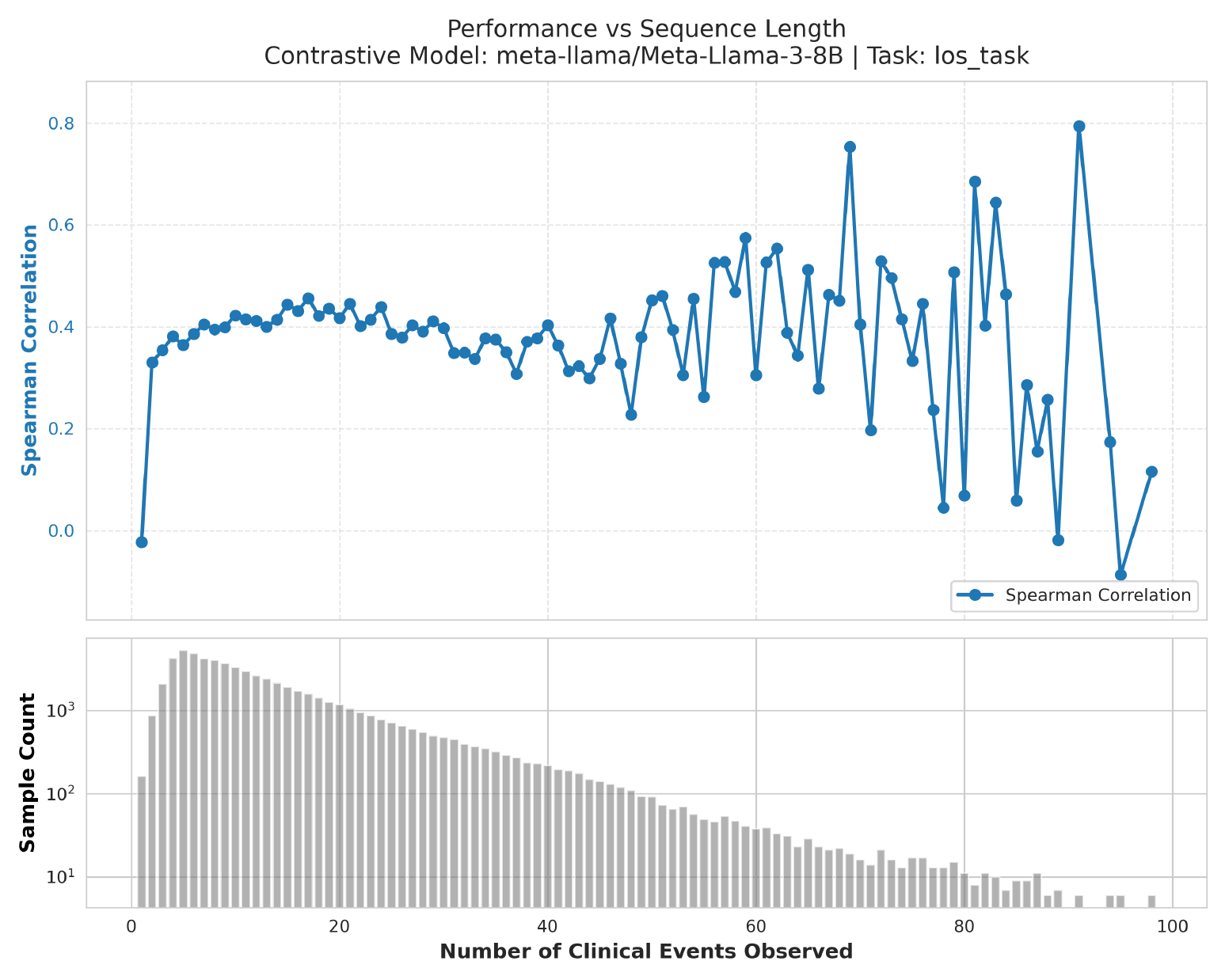}
        \caption{Contrastive (Llama-3)}
    \end{subfigure}\hfill
    \begin{subfigure}{0.32\textwidth}
        \centering
        \includegraphics[width=\linewidth]{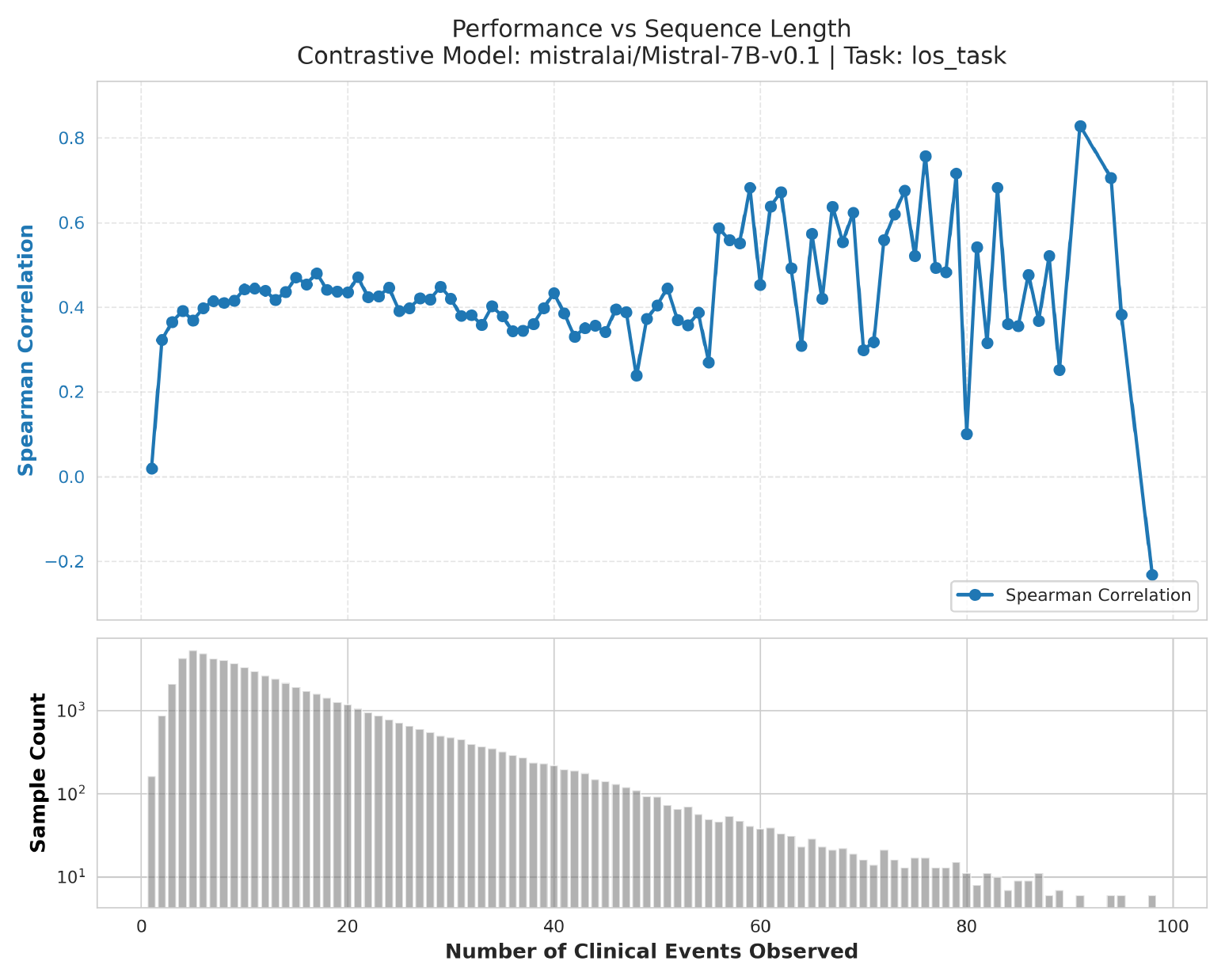}
        \caption{Contrastive (Mistral)}
    \end{subfigure}\hfill
    \begin{subfigure}{0.32\textwidth}
        \centering
        \includegraphics[width=\linewidth]{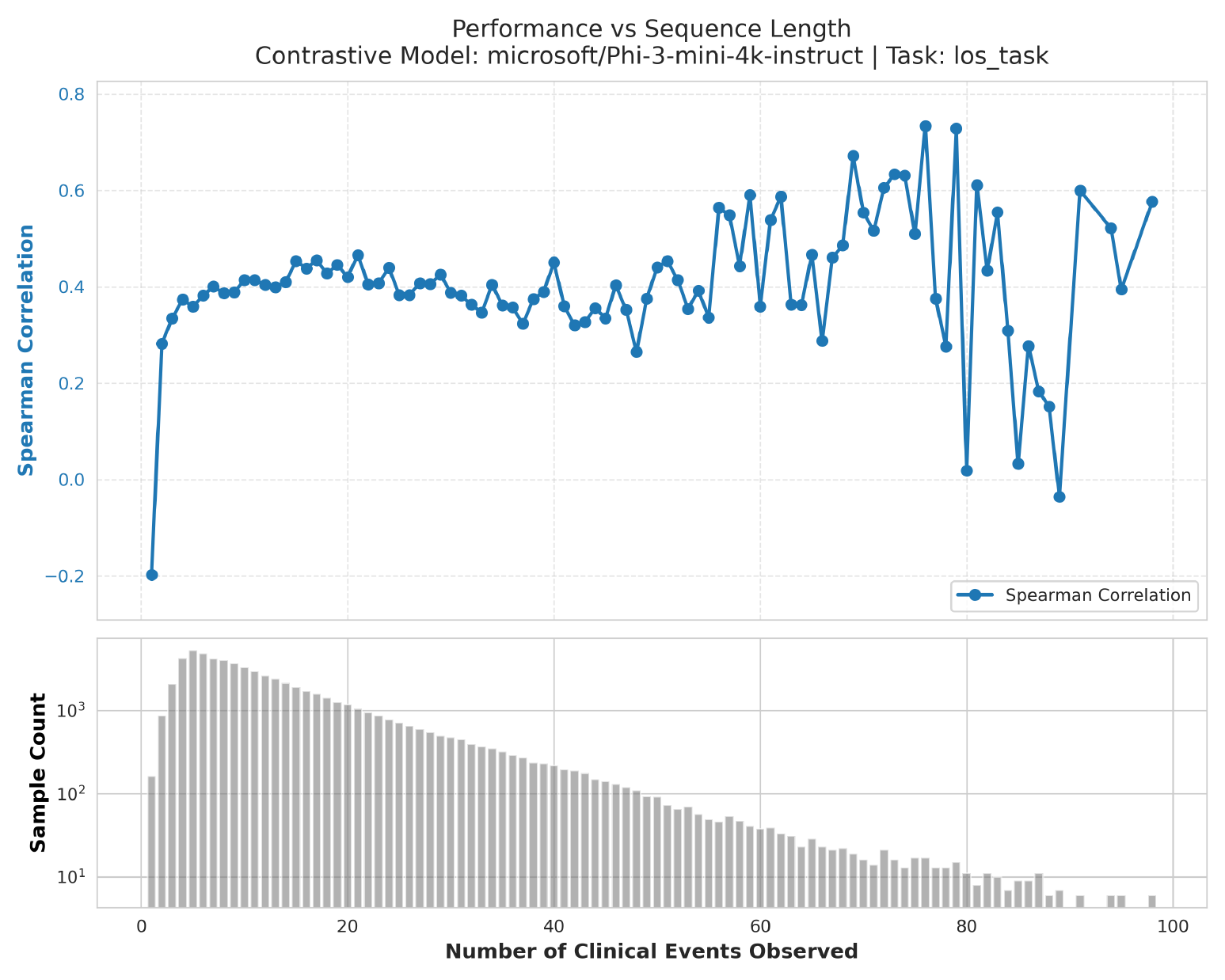}
        \caption{Contrastive (Phi-3)}
    \end{subfigure}
    
    \vspace{1em}
    
    % Row 2: Contrastive
    \begin{subfigure}{0.32\textwidth}
        \centering
        \includegraphics[width=\linewidth]{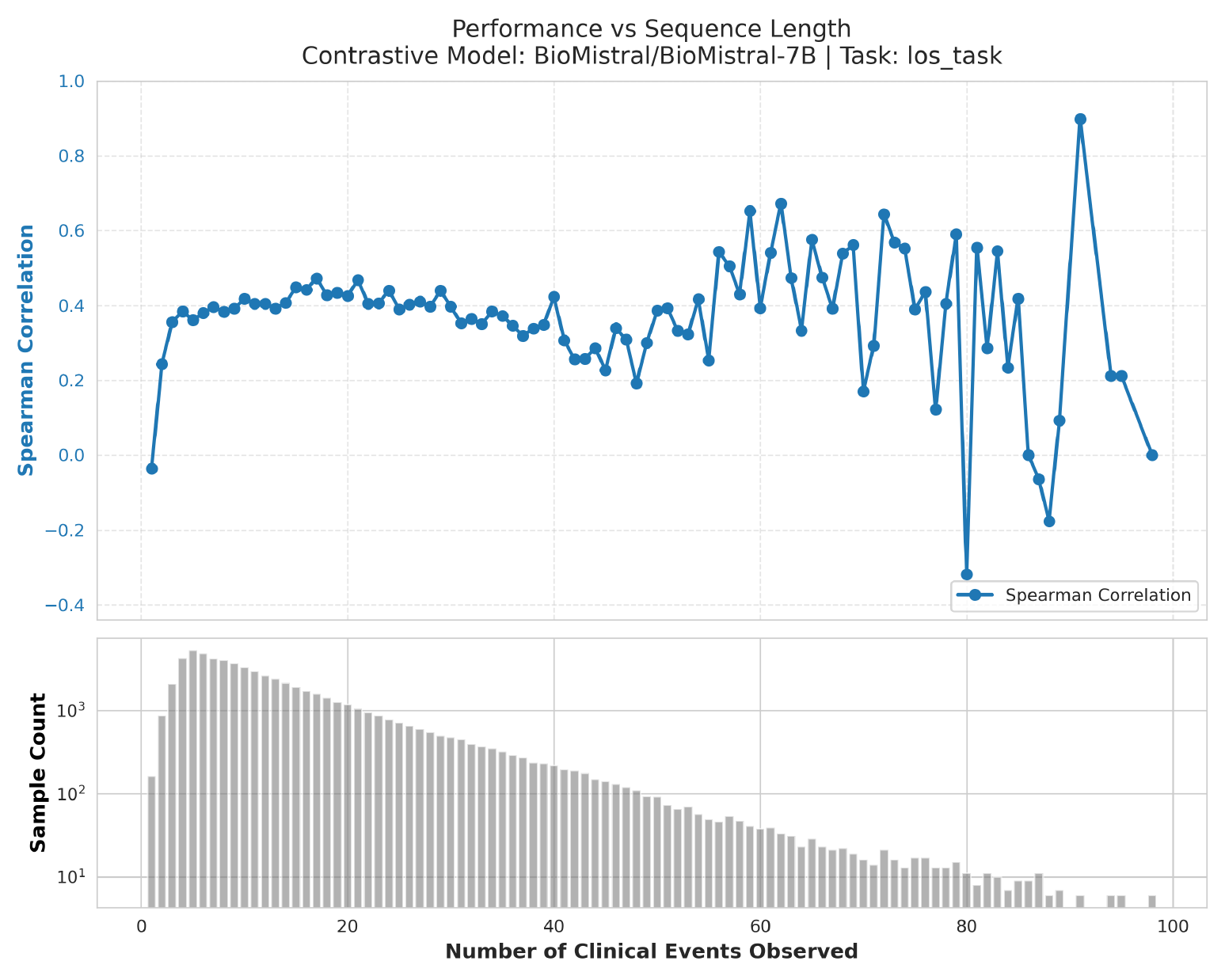}
        \caption{Contrastive (BioMistral)}
    \end{subfigure}\hfill
    \begin{subfigure}{0.32\textwidth}
        \centering
        \includegraphics[width=\linewidth]{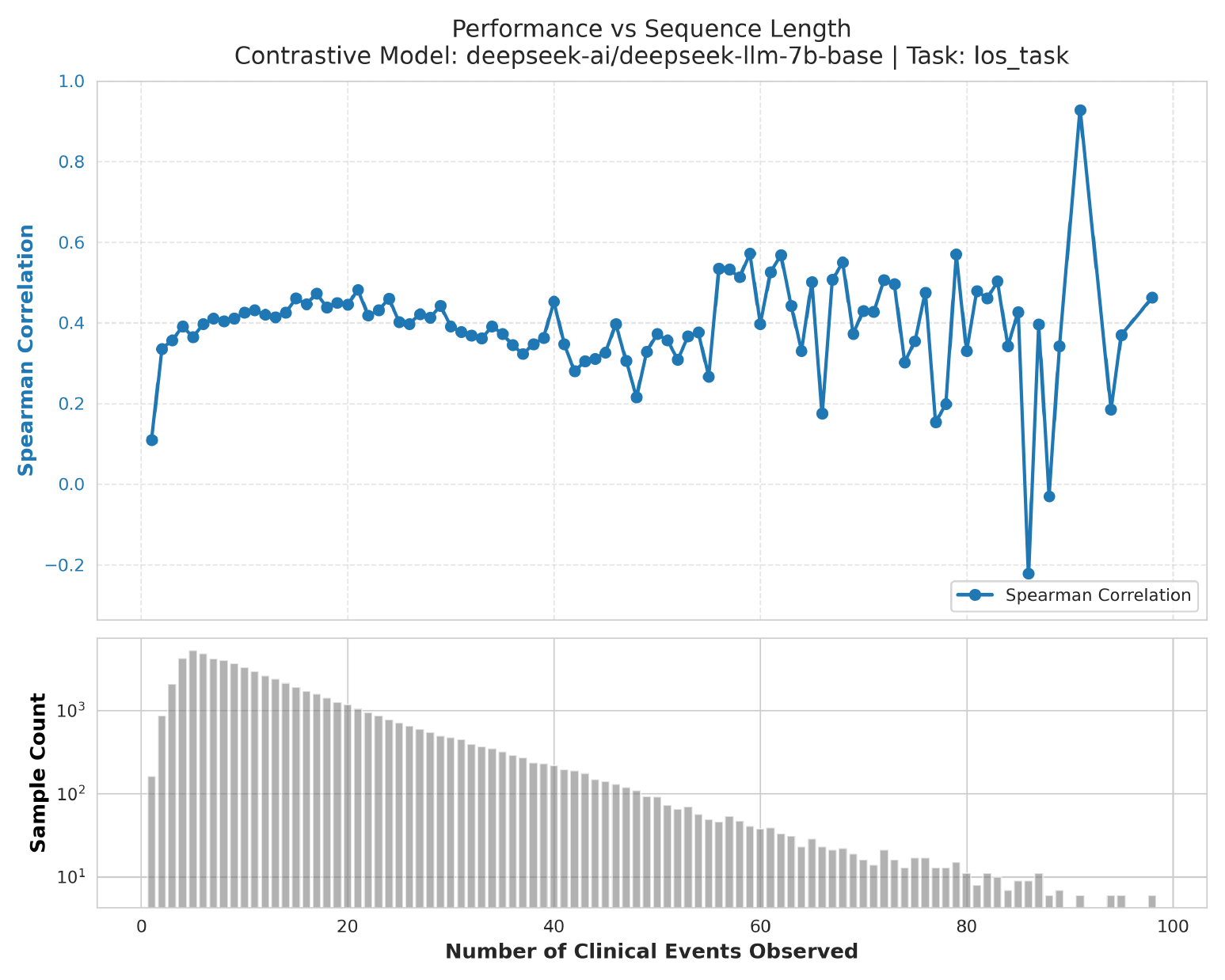}
        \caption{Contrastive (Deepseek)}
    \end{subfigure}\hfill
    \begin{subfigure}{0.32\textwidth}
        \centering
        \includegraphics[width=\linewidth]{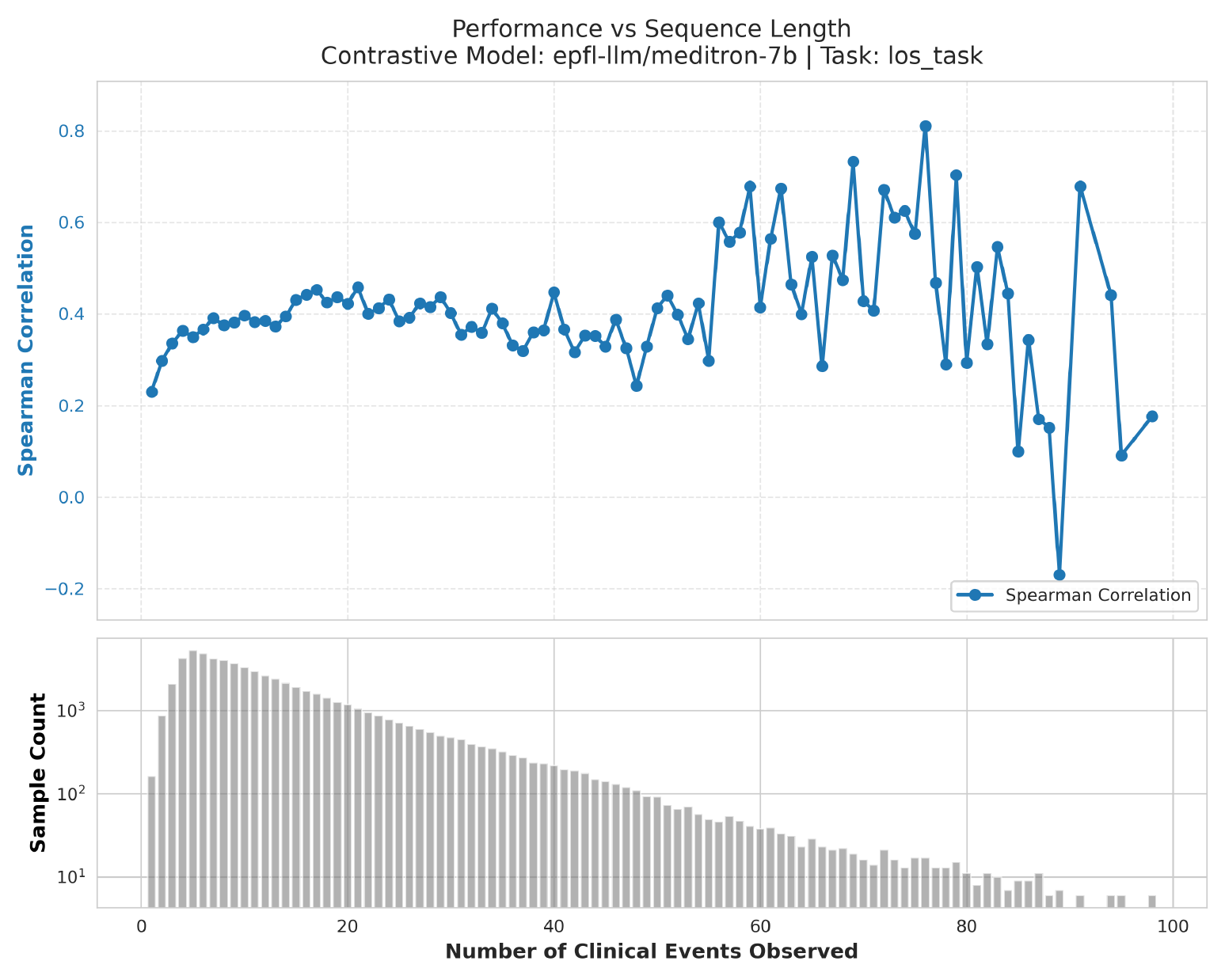}
        \caption{Contrastive (Meditron)}
    \end{subfigure}
    
    \caption{Length of Stay performance plots for contrastive sequential models evaluated on eICU. The Spearman Correlation Coefficient is plotted against the patient stays with the associated number of events in their sequences, along with a visualization of the patient stay lengths.}
    \label{fig:eicu_los_contrastive_trajectory}
\end{figure}

\begin{figure}[htbp]
    \centering
    % Row 1: E2E Baseline
    \begin{subfigure}{0.32\textwidth}
        \centering
        \includegraphics[width=\linewidth]{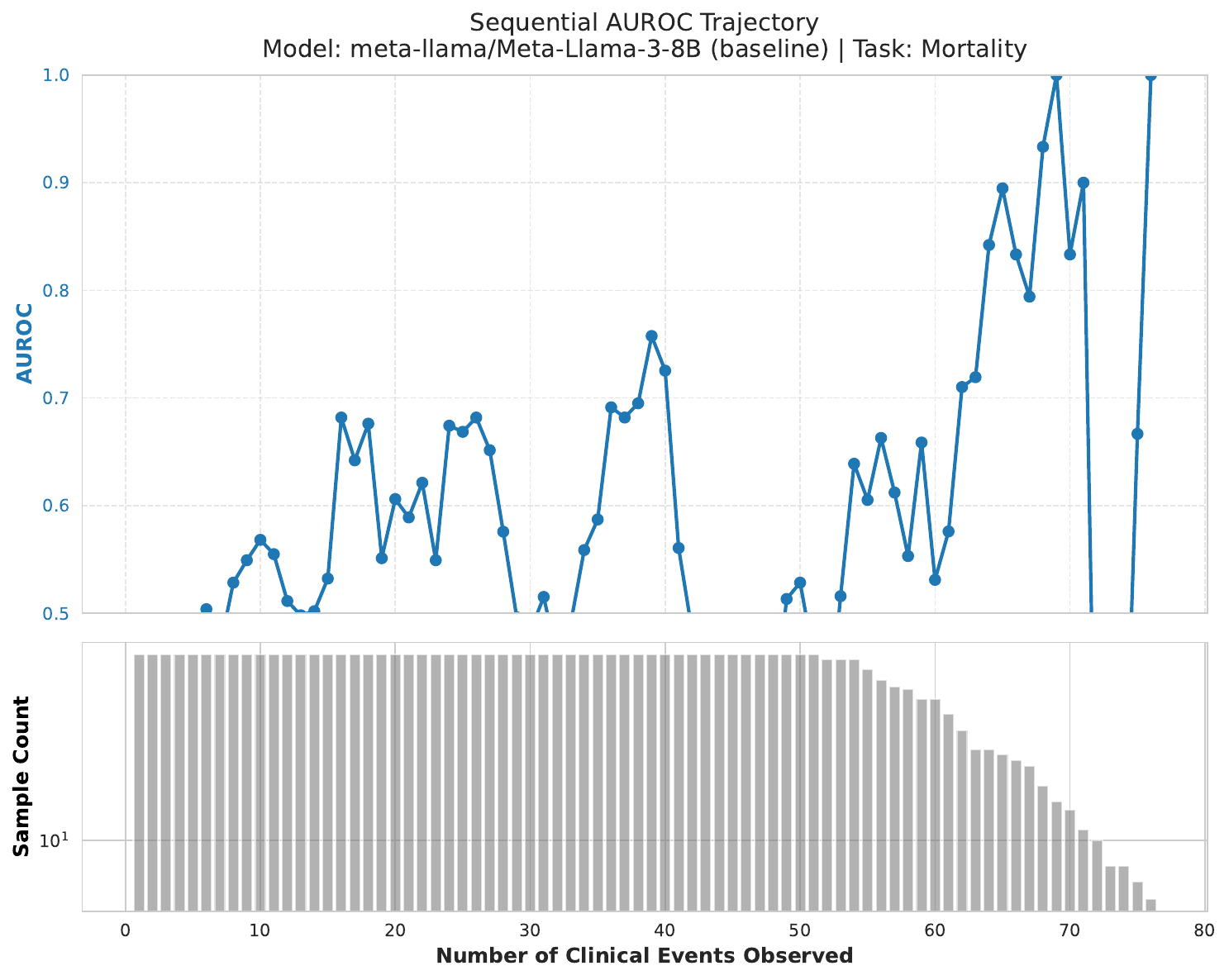}
        \caption{Non-Contrastive (Llama-3)}
    \end{subfigure}\hfill
    \begin{subfigure}{0.32\textwidth}
        \centering
        \includegraphics[width=\linewidth]{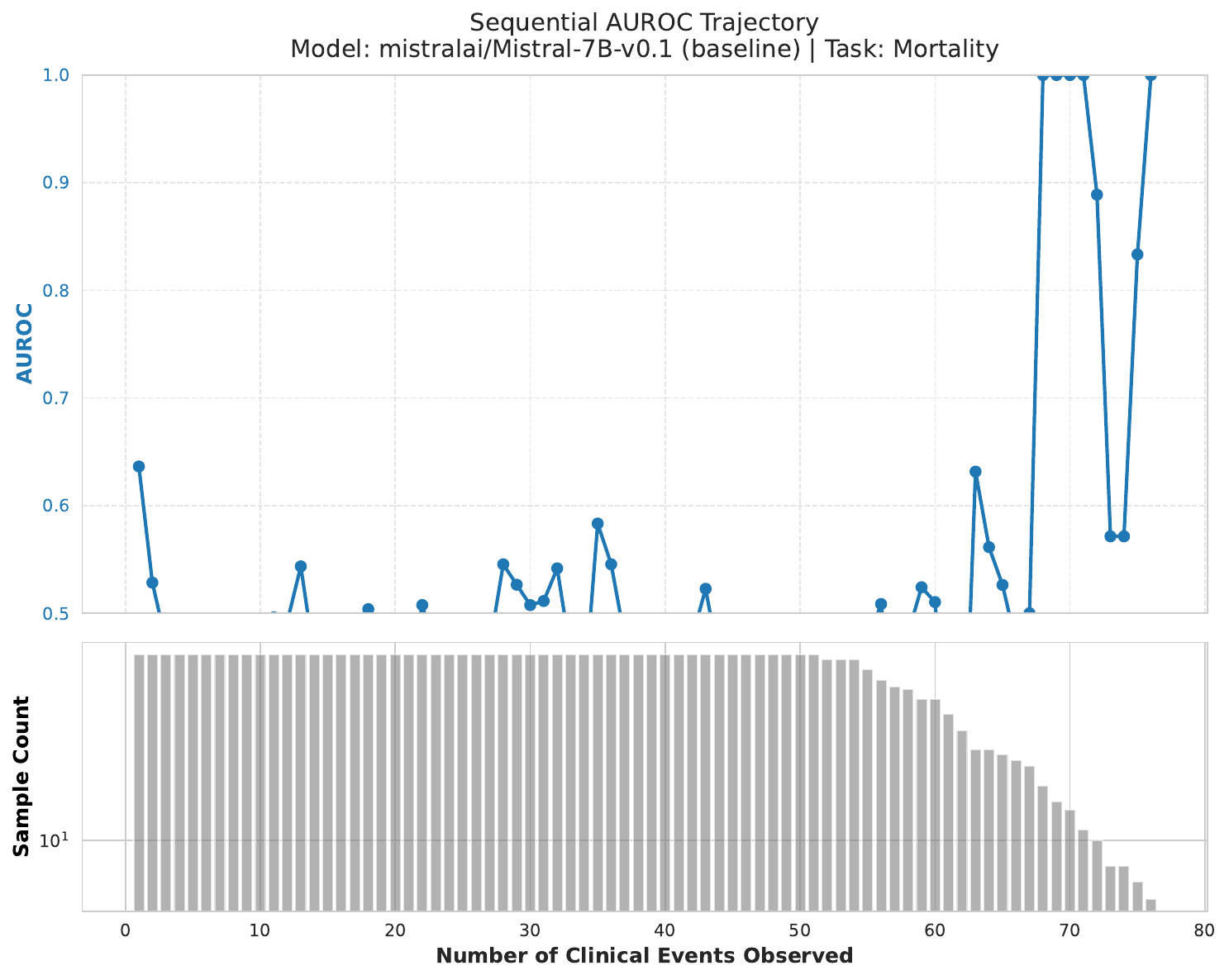}
        \caption{Non-Contrastive (Mistral)}
    \end{subfigure}\hfill
    \begin{subfigure}{0.32\textwidth}
        \centering
        \includegraphics[width=\linewidth]{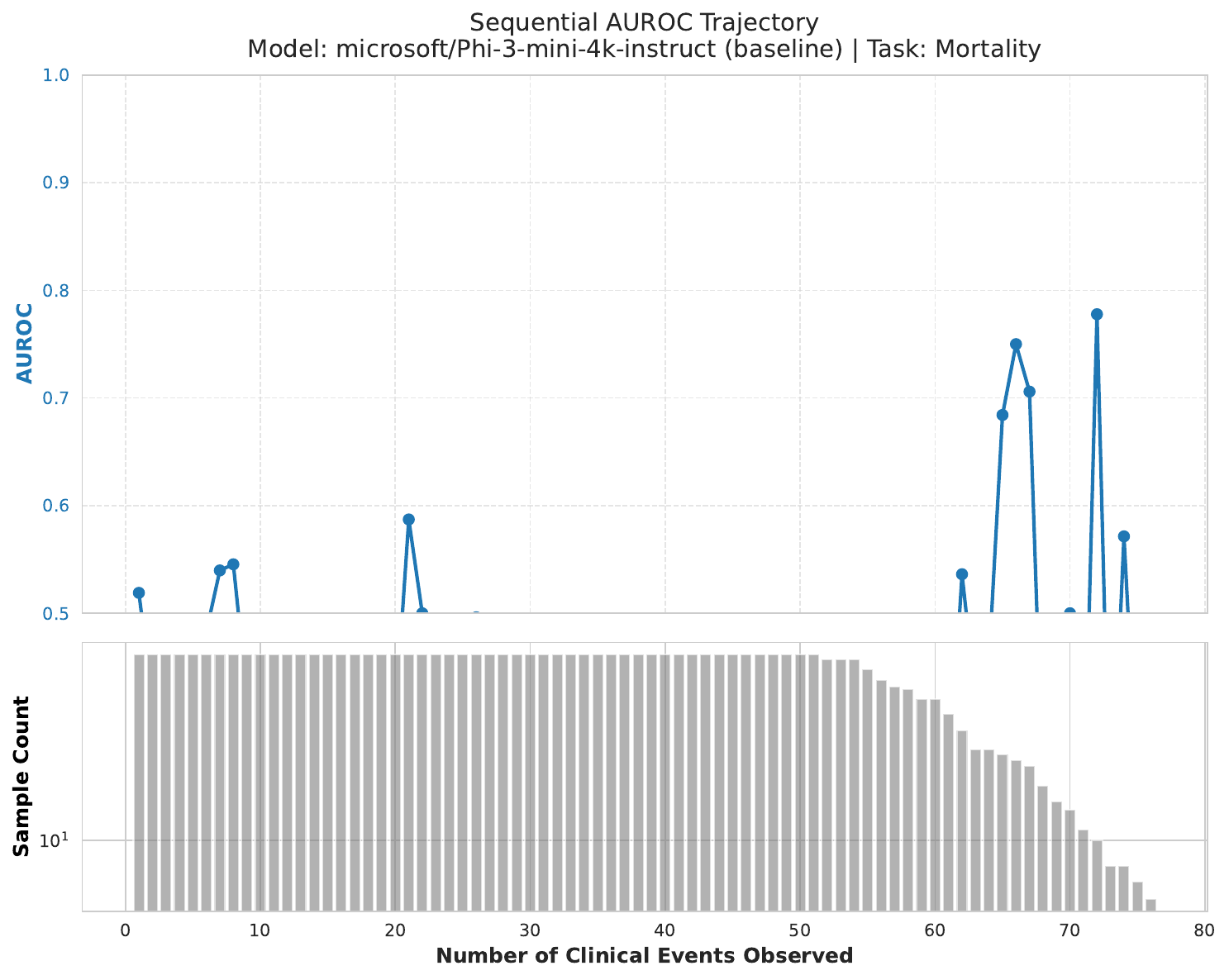}
        \caption{Non-Contrastive (Phi-3)}
    \end{subfigure}
    
    \vspace{1em}
    
    % Row 2: Contrastive
    \begin{subfigure}{0.32\textwidth}
        \centering
        \includegraphics[width=\linewidth]{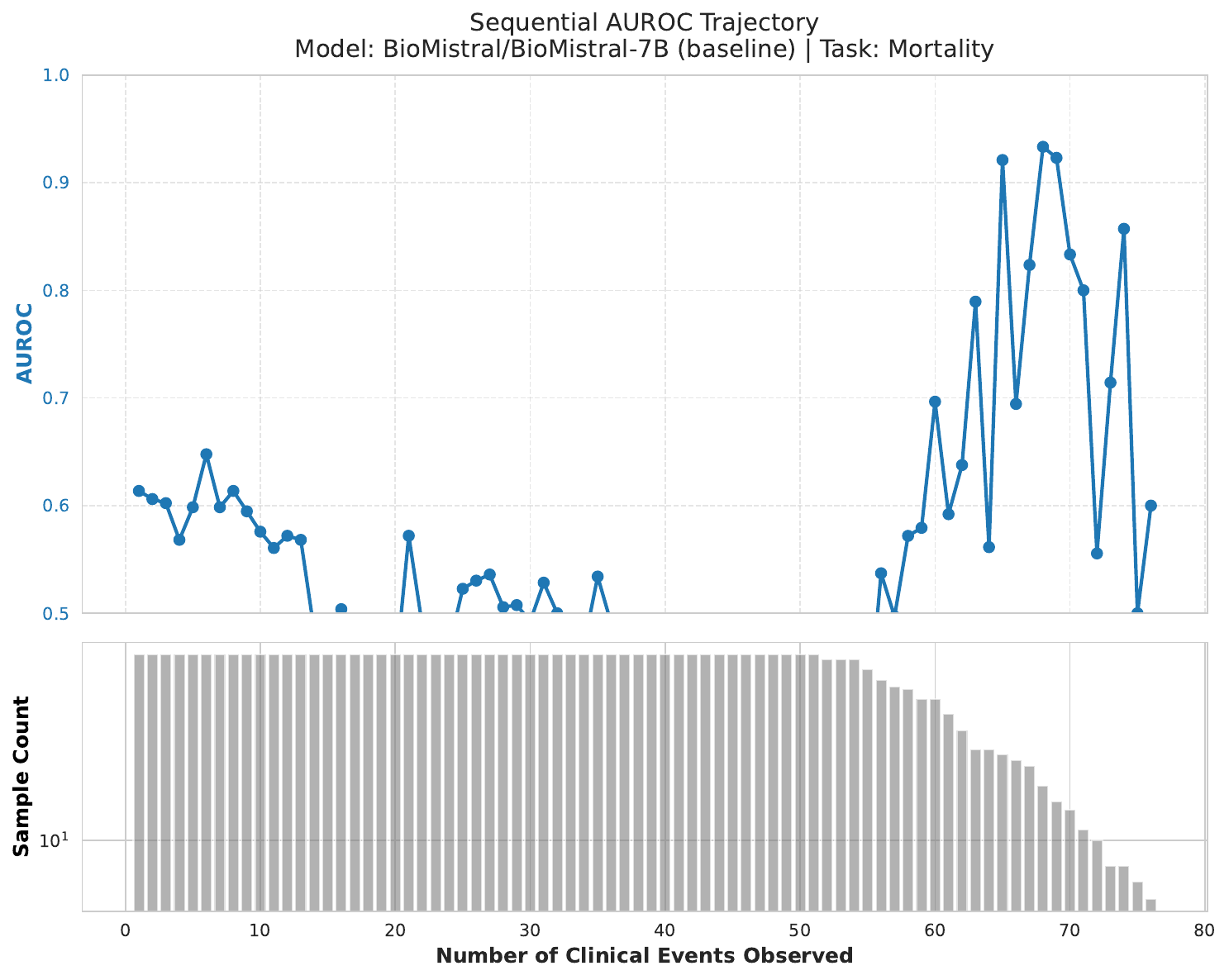}
        \caption{Non-Contrastive (BioMistral)}
    \end{subfigure}\hfill
    \begin{subfigure}{0.32\textwidth}
        \centering
        \includegraphics[width=\linewidth]{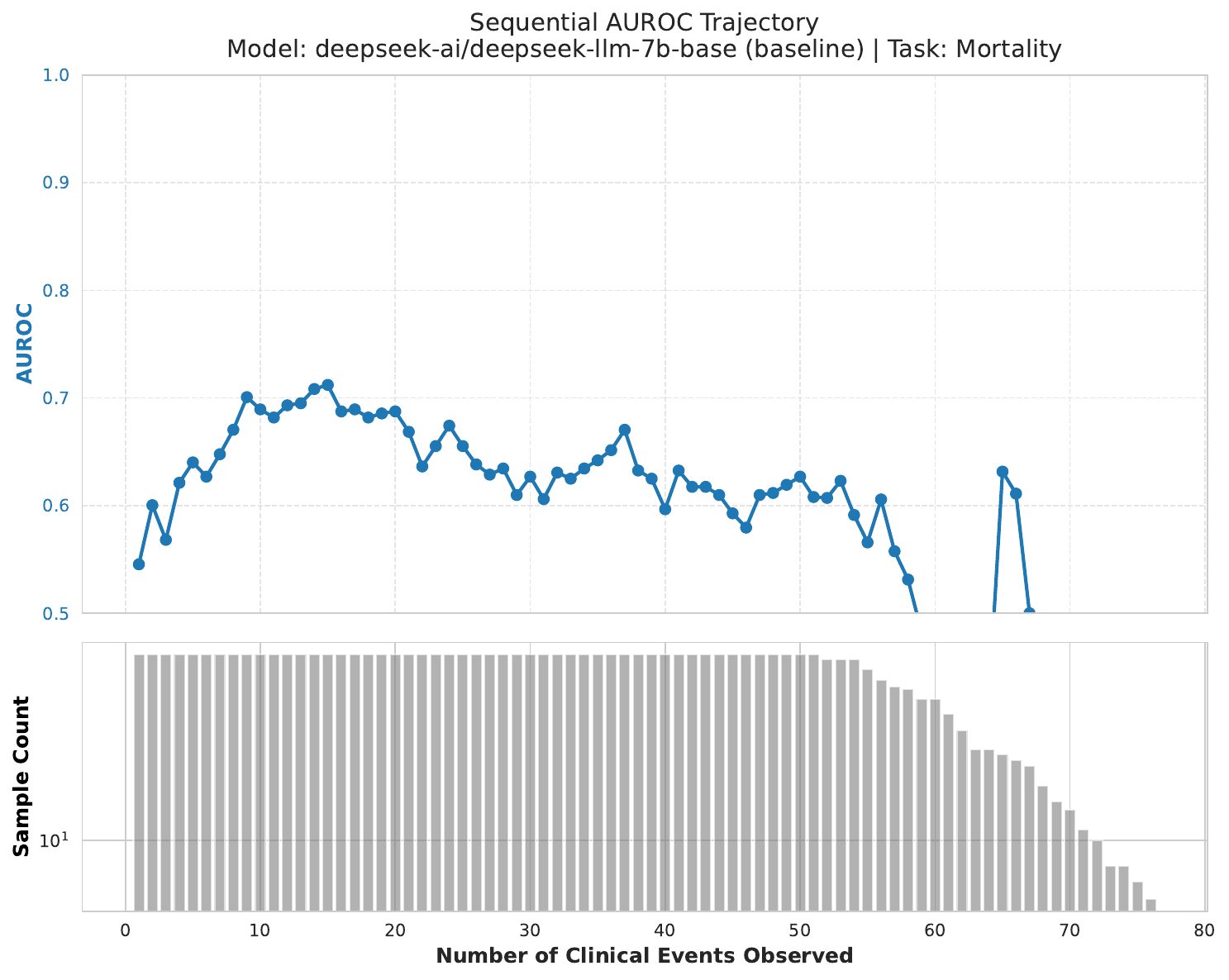}
        \caption{Non-Contrastive (Deepseek)}
    \end{subfigure}\hfill
    \begin{subfigure}{0.32\textwidth}
        \centering
        \includegraphics[width=\linewidth]{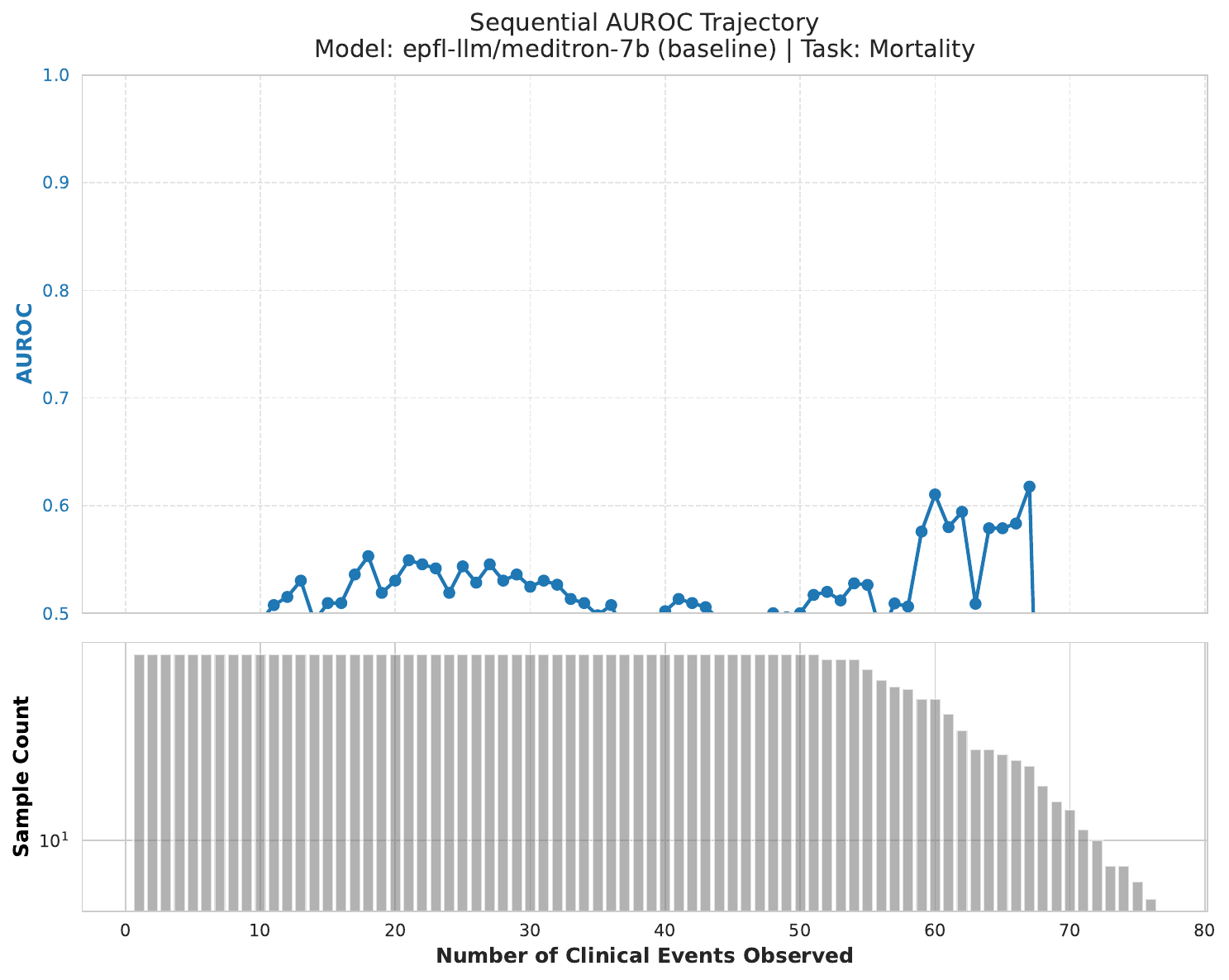}
        \caption{Non-Contrastive (Meditron)}
    \end{subfigure}
    
    \caption{Mortality performance plots for non-contrastive sequential models evaluated on MIMIC-IV. The AUROC performance is plotted against the patient stays with the associated number of events in their sequences, along with a visualization of the patient stay lengths.}
    \label{fig:MIMIC_mortality_base_trajectory}
\end{figure}

\begin{figure}[htbp]
    \centering
    % Row 1: E2E Baseline
    \begin{subfigure}{0.32\textwidth}
        \centering
        \includegraphics[width=\linewidth]{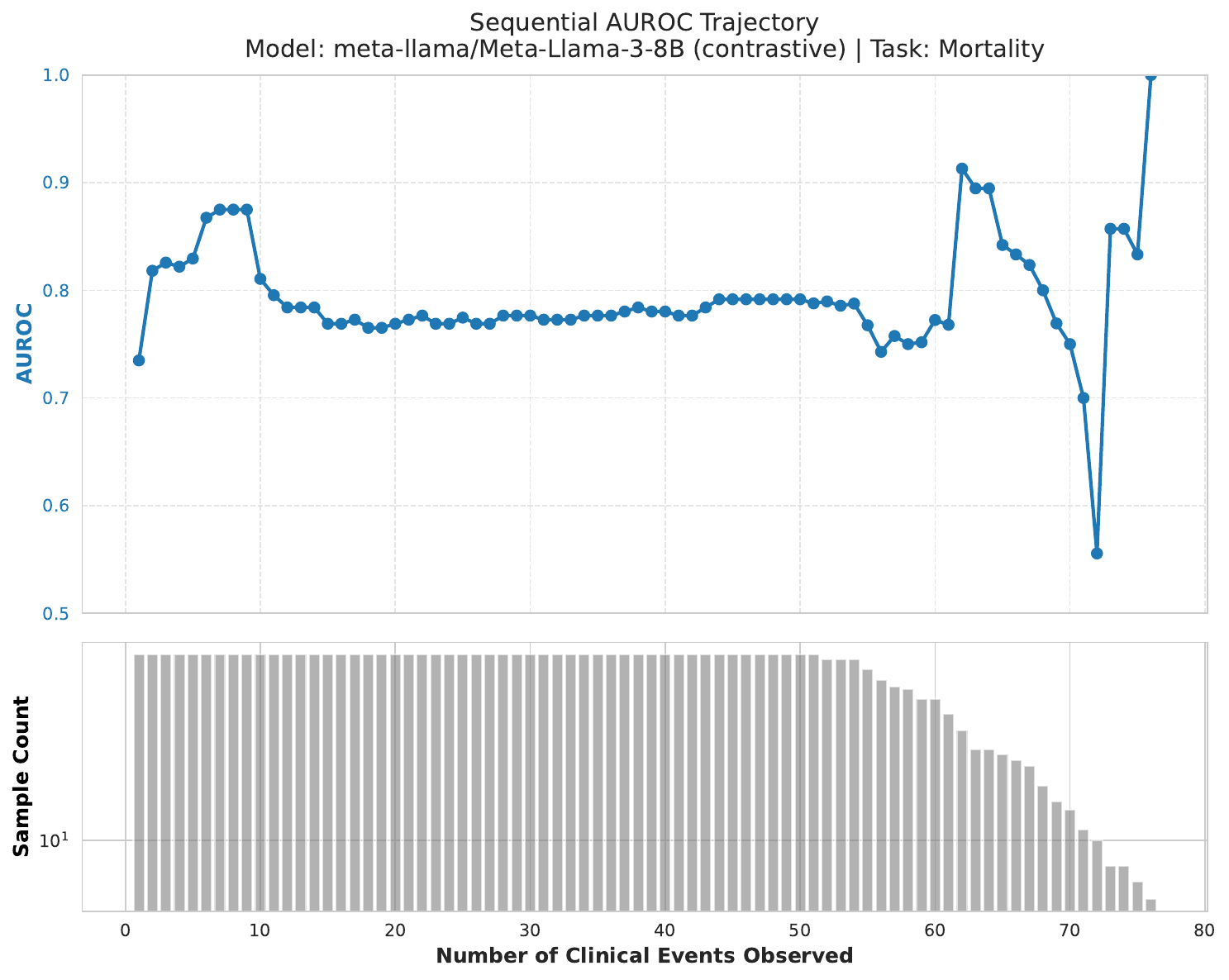}
        \caption{Contrastive (Llama-3)}
    \end{subfigure}\hfill
    \begin{subfigure}{0.32\textwidth}
        \centering
        \includegraphics[width=\linewidth]{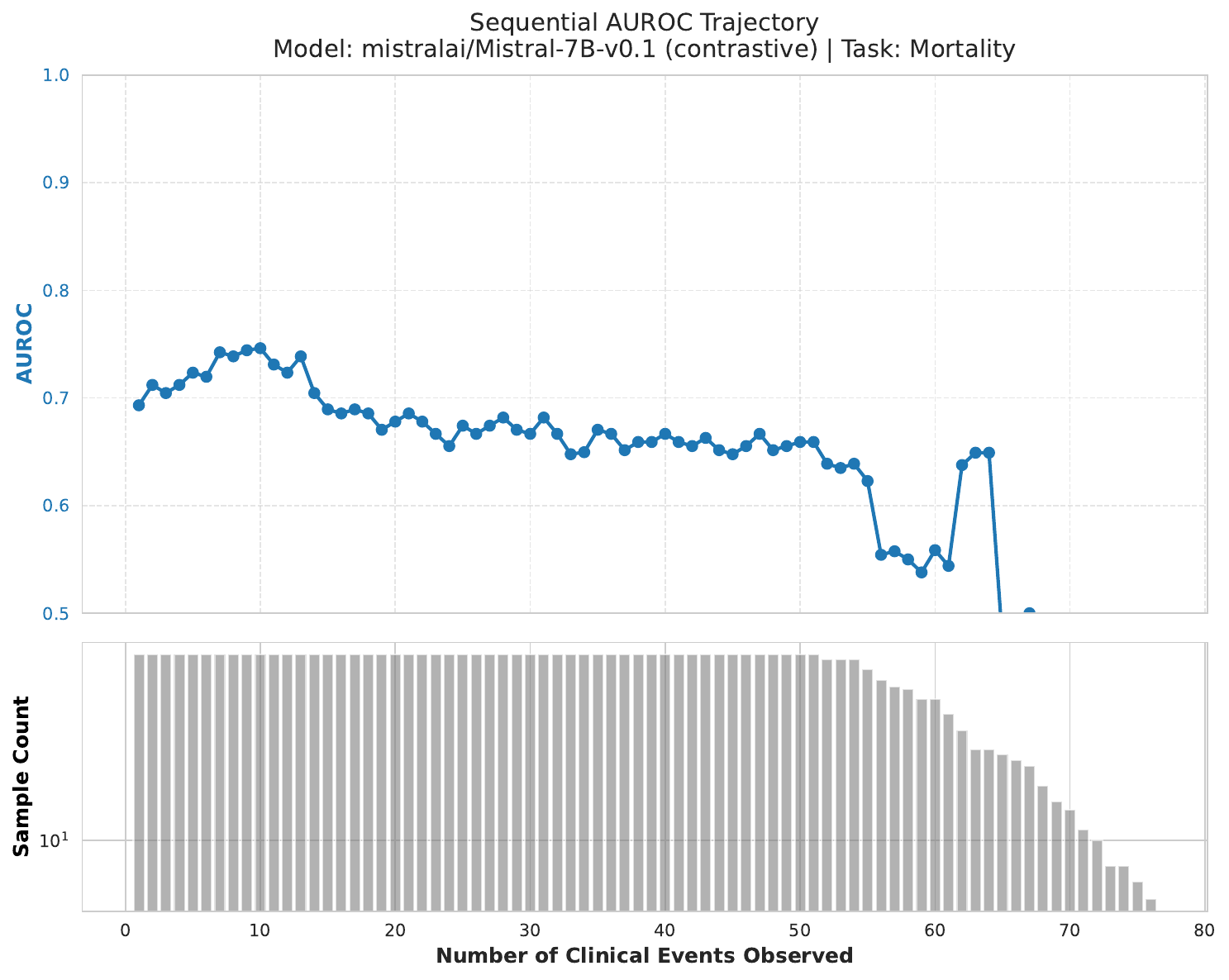}
        \caption{Contrastive (Mistral)}
    \end{subfigure}\hfill
    \begin{subfigure}{0.32\textwidth}
        \centering
        \includegraphics[width=\linewidth]{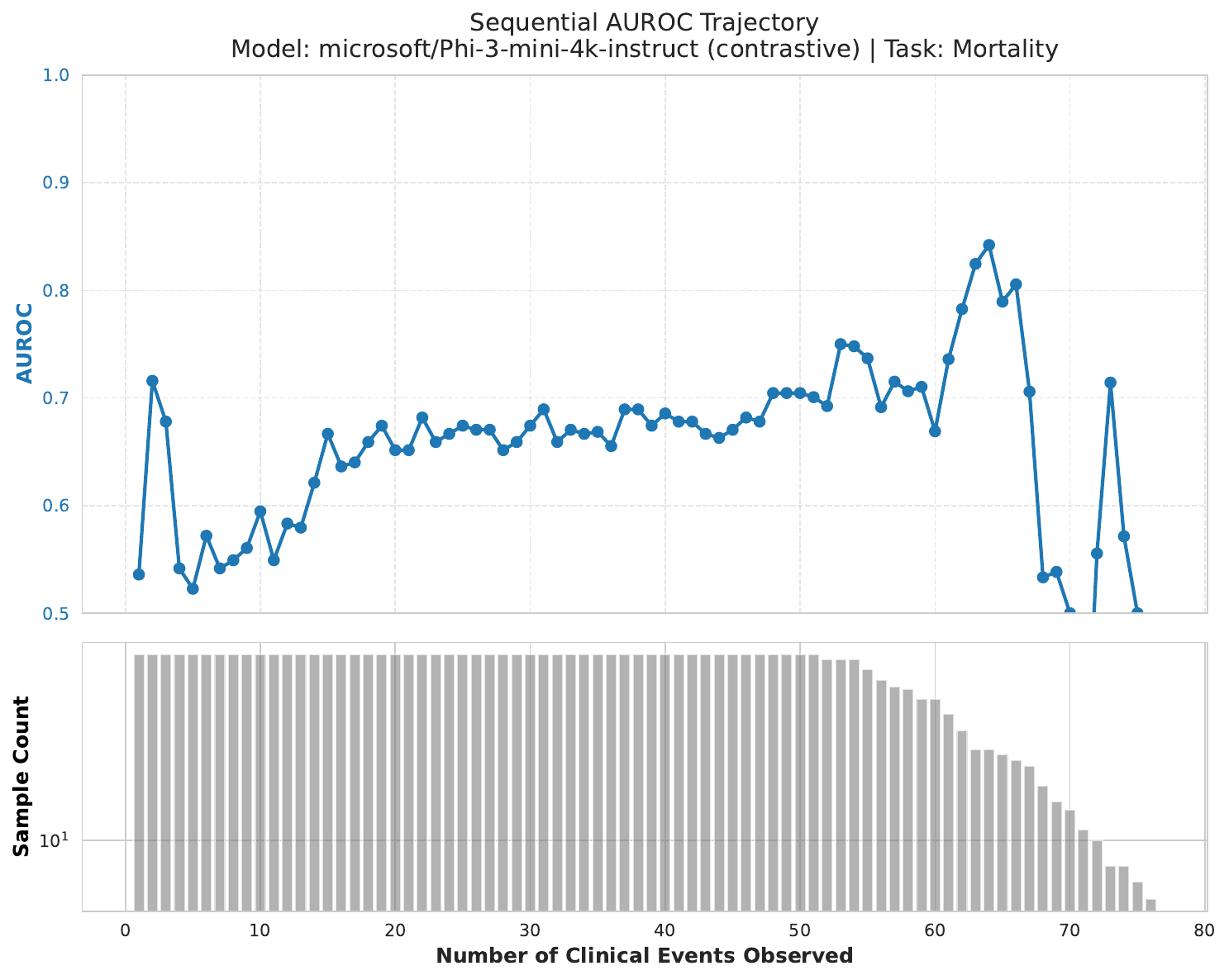}
        \caption{Contrastive (Phi-3)}
    \end{subfigure}
    
    \vspace{1em}
    
    % Row 2: Contrastive
    \begin{subfigure}{0.32\textwidth}
        \centering
        \includegraphics[width=\linewidth]{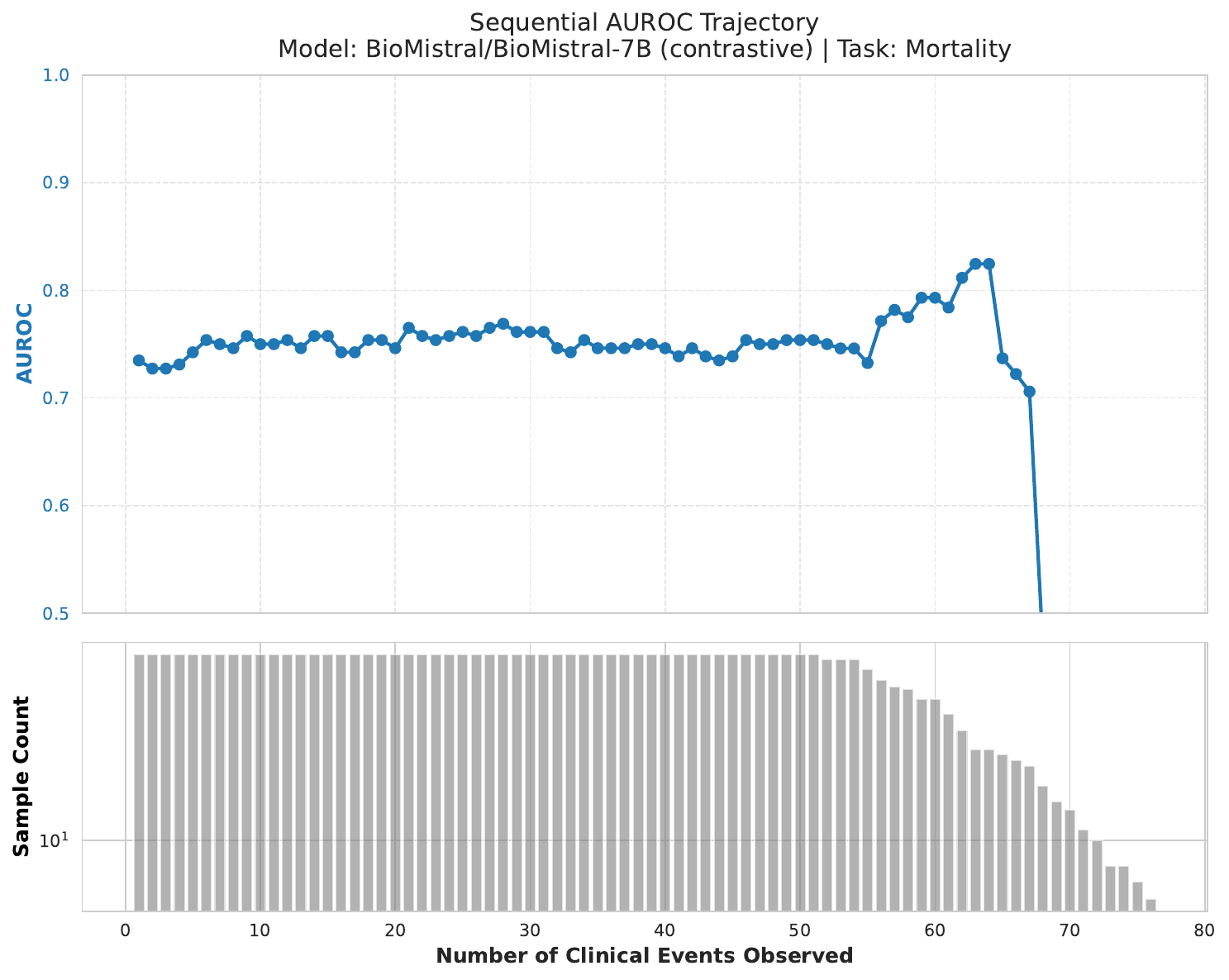}
        \caption{Contrastive (BioMistral)}
    \end{subfigure}\hfill
    \begin{subfigure}{0.32\textwidth}
        \centering
        \includegraphics[width=\linewidth]{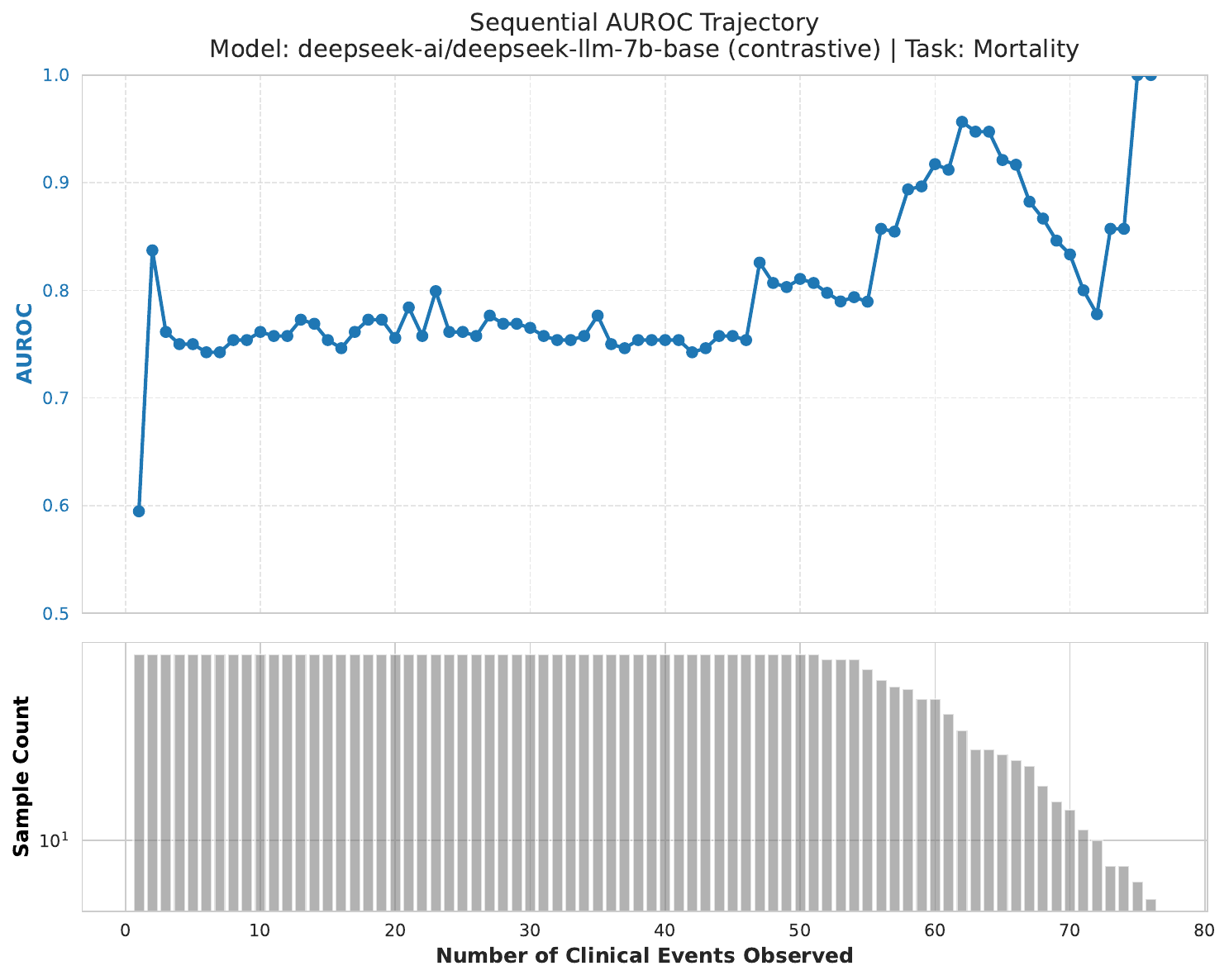}
        \caption{Contrastive (Deepseek)}
    \end{subfigure}\hfill
    \begin{subfigure}{0.32\textwidth}
        \centering
        \includegraphics[width=\linewidth]{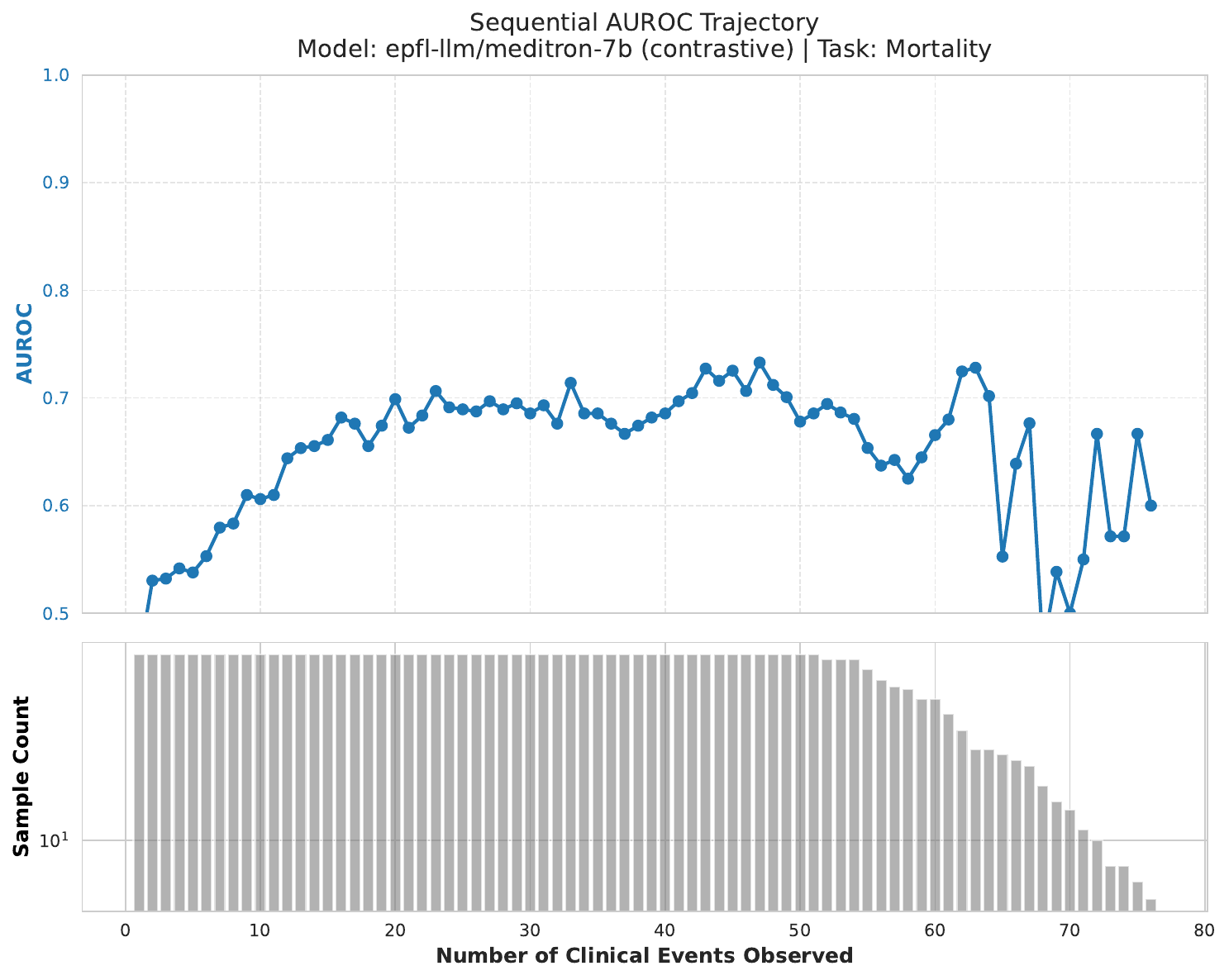}
        \caption{Contrastive (Meditron)}
    \end{subfigure}
    
    \caption{Mortality performance plots for contrastive sequential models evaluated on MIMIC-IV. The AUROC performance is plotted against the patient stays with the associated number of events in their sequences, along with a visualization of the patient stay lengths.}
    \label{fig:MIMIC_mortality_contrastive_trajectory}
\end{figure}

\begin{figure}[htbp]
    \centering
    % Row 1: E2E Baseline
    \begin{subfigure}{0.32\textwidth}
        \centering
        \includegraphics[width=\linewidth]{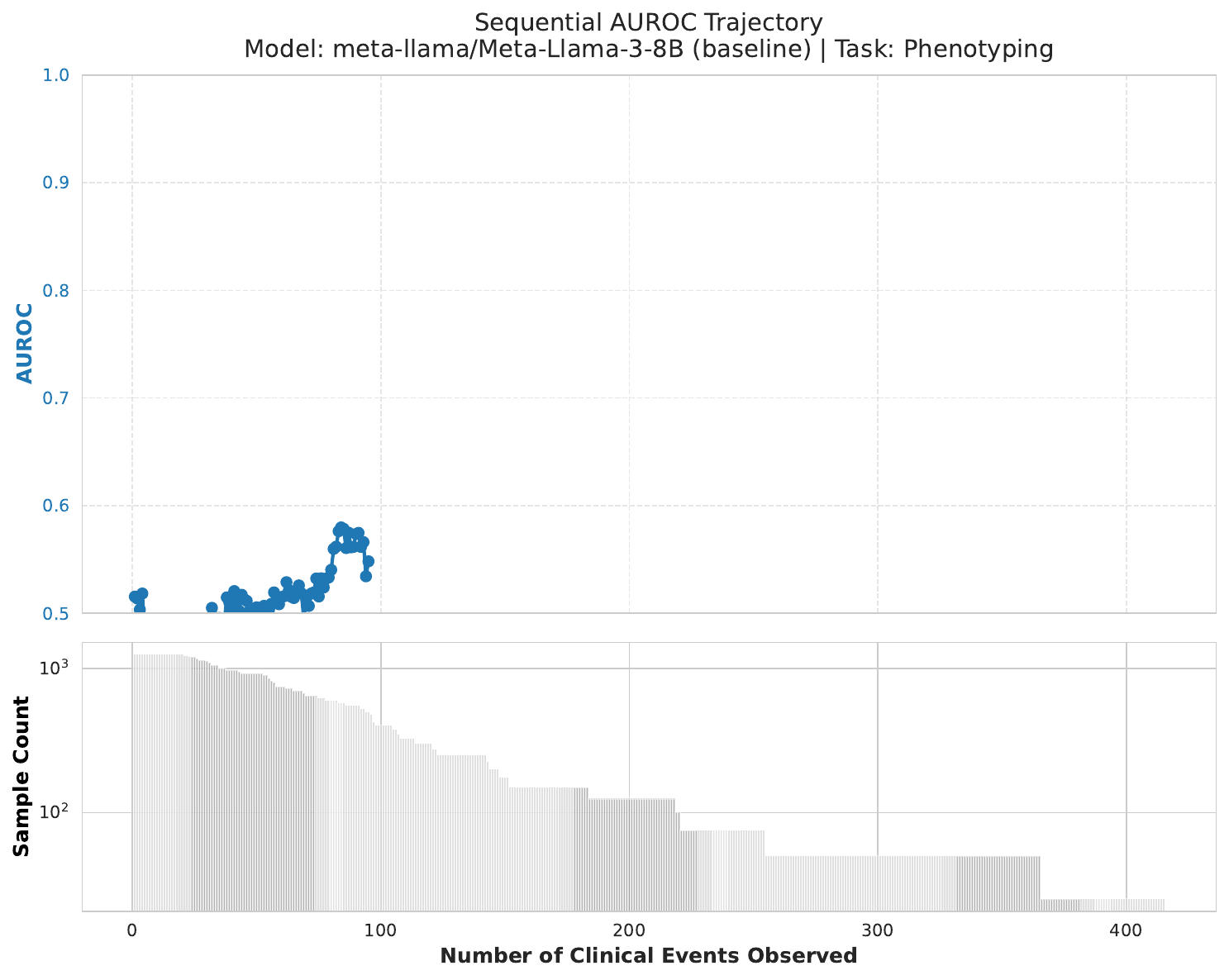}
        \caption{Non-Contrastive (Llama-3)}
    \end{subfigure}\hfill
    \begin{subfigure}{0.32\textwidth}
        \centering
        \includegraphics[width=\linewidth]{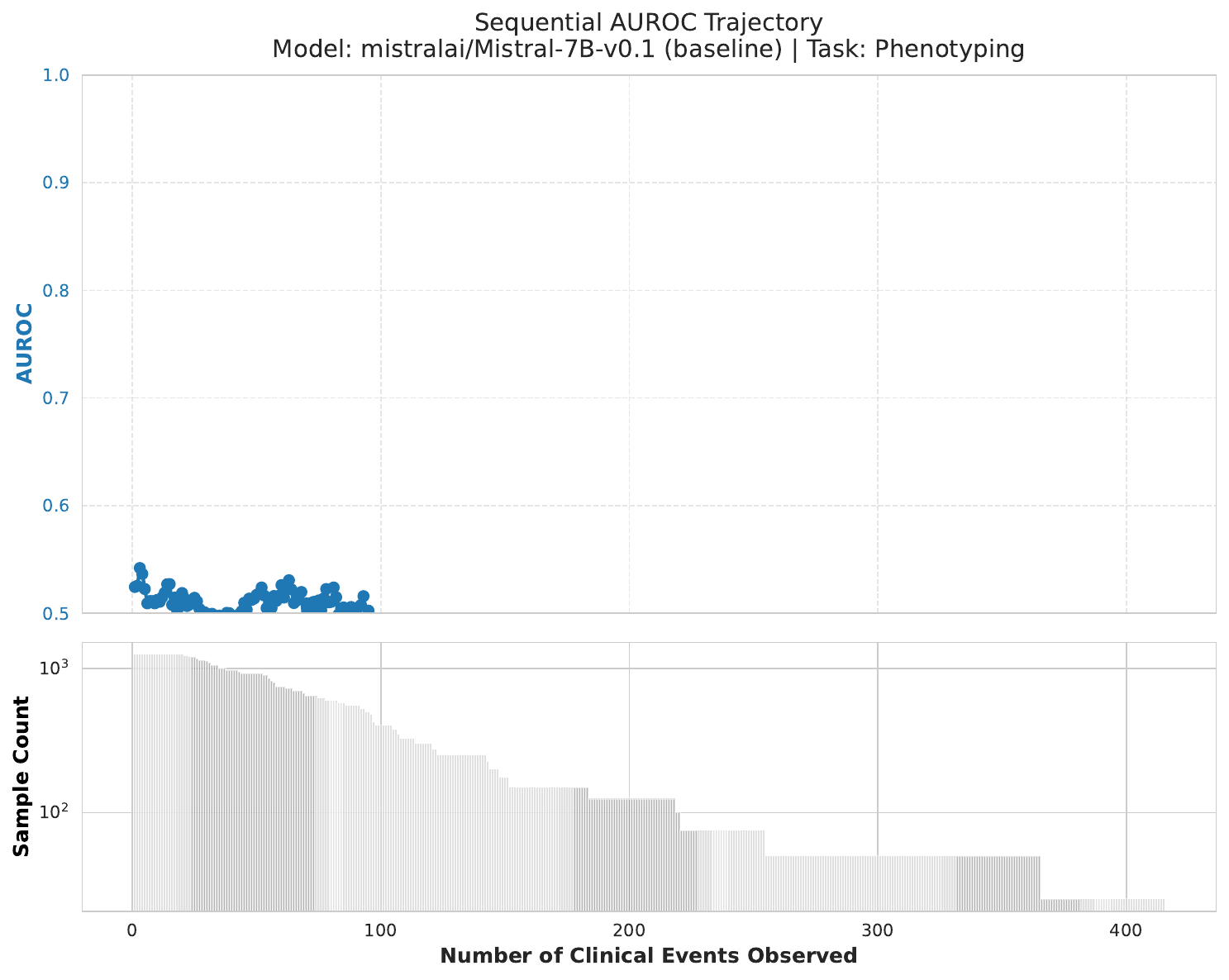}
        \caption{Non-Contrastive (Mistral)}
    \end{subfigure}\hfill
    \begin{subfigure}{0.32\textwidth}
        \centering
        \includegraphics[width=\linewidth]{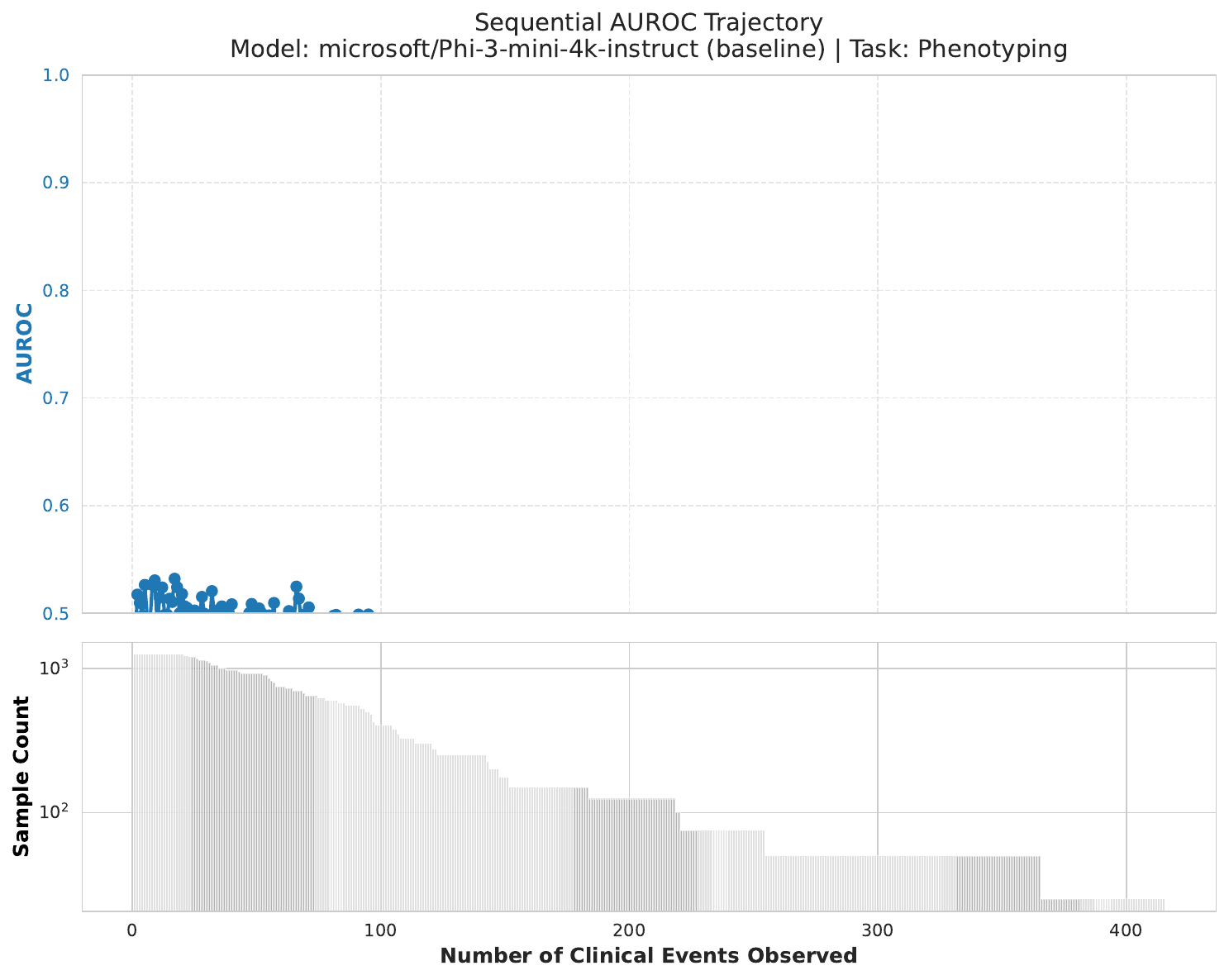}
        \caption{Non-Contrastive (Phi-3)}
    \end{subfigure}
    
    \vspace{1em}
    
    % Row 2: Contrastive
    \begin{subfigure}{0.32\textwidth}
        \centering
        \includegraphics[width=\linewidth]{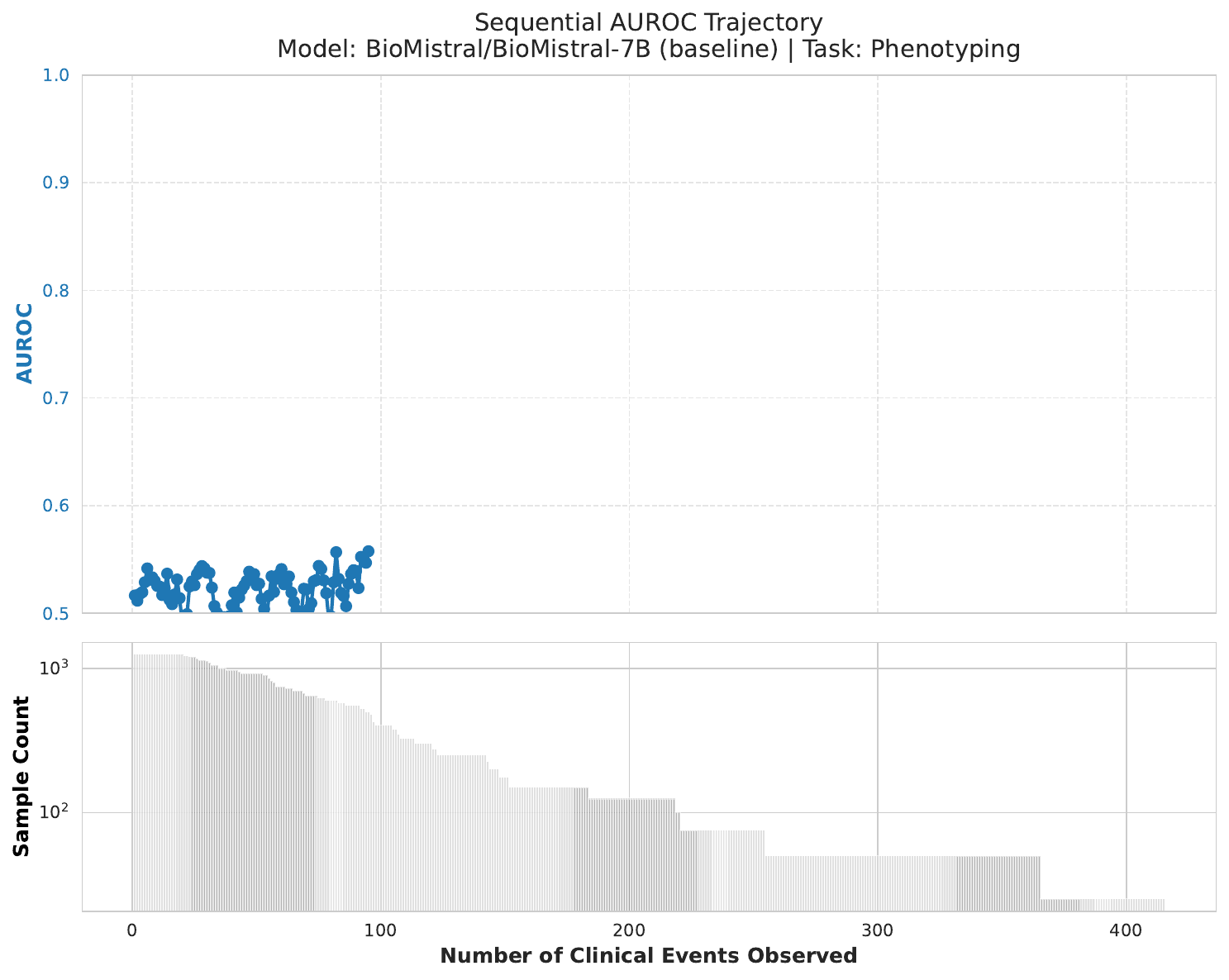}
        \caption{Non-Contrastive (BioMistral)}
    \end{subfigure}\hfill
    \begin{subfigure}{0.32\textwidth}
        \centering
        \includegraphics[width=\linewidth]{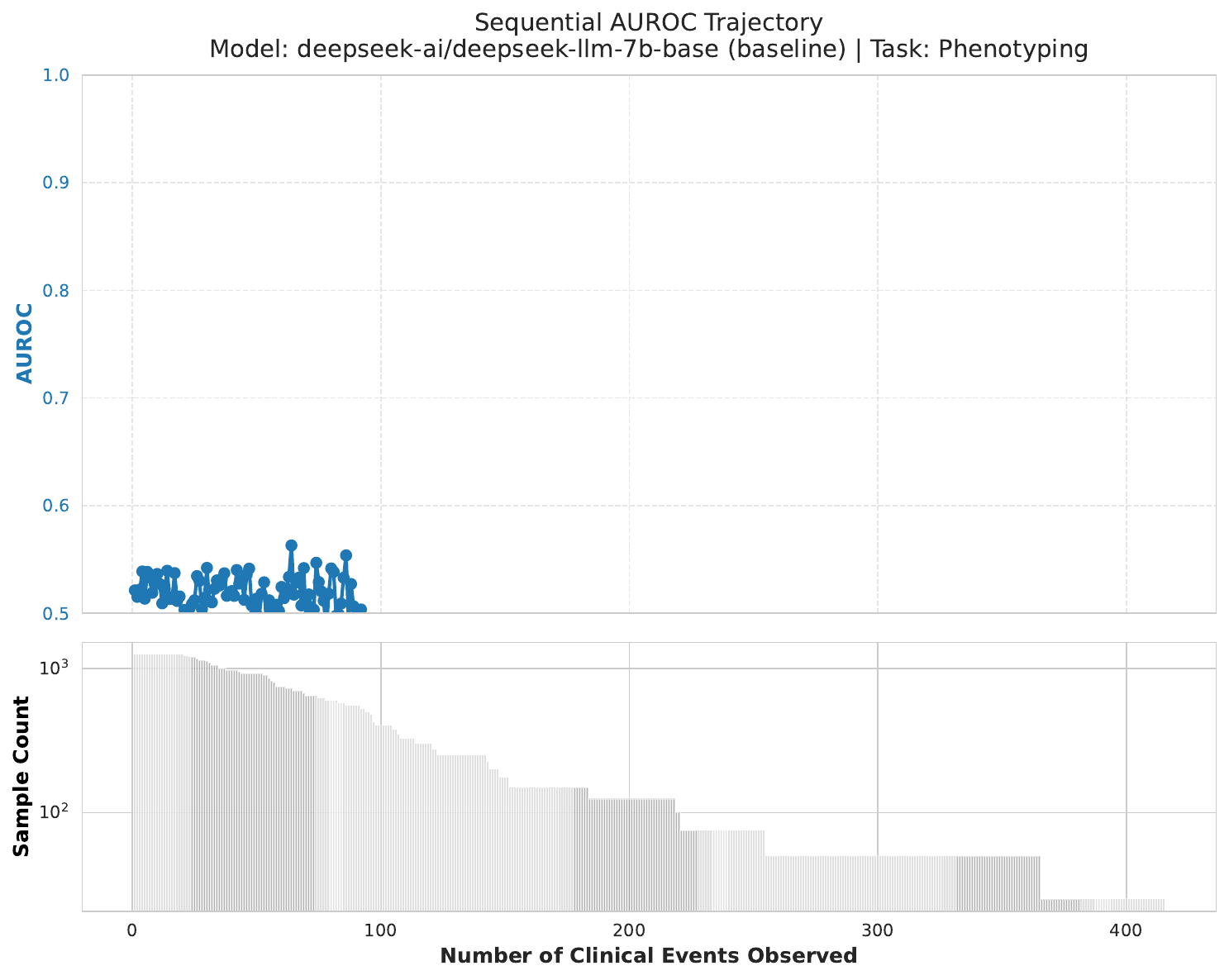}
        \caption{Non-Contrastive (Deepseek)}
    \end{subfigure}\hfill
    \begin{subfigure}{0.32\textwidth}
        \centering
        \includegraphics[width=\linewidth]{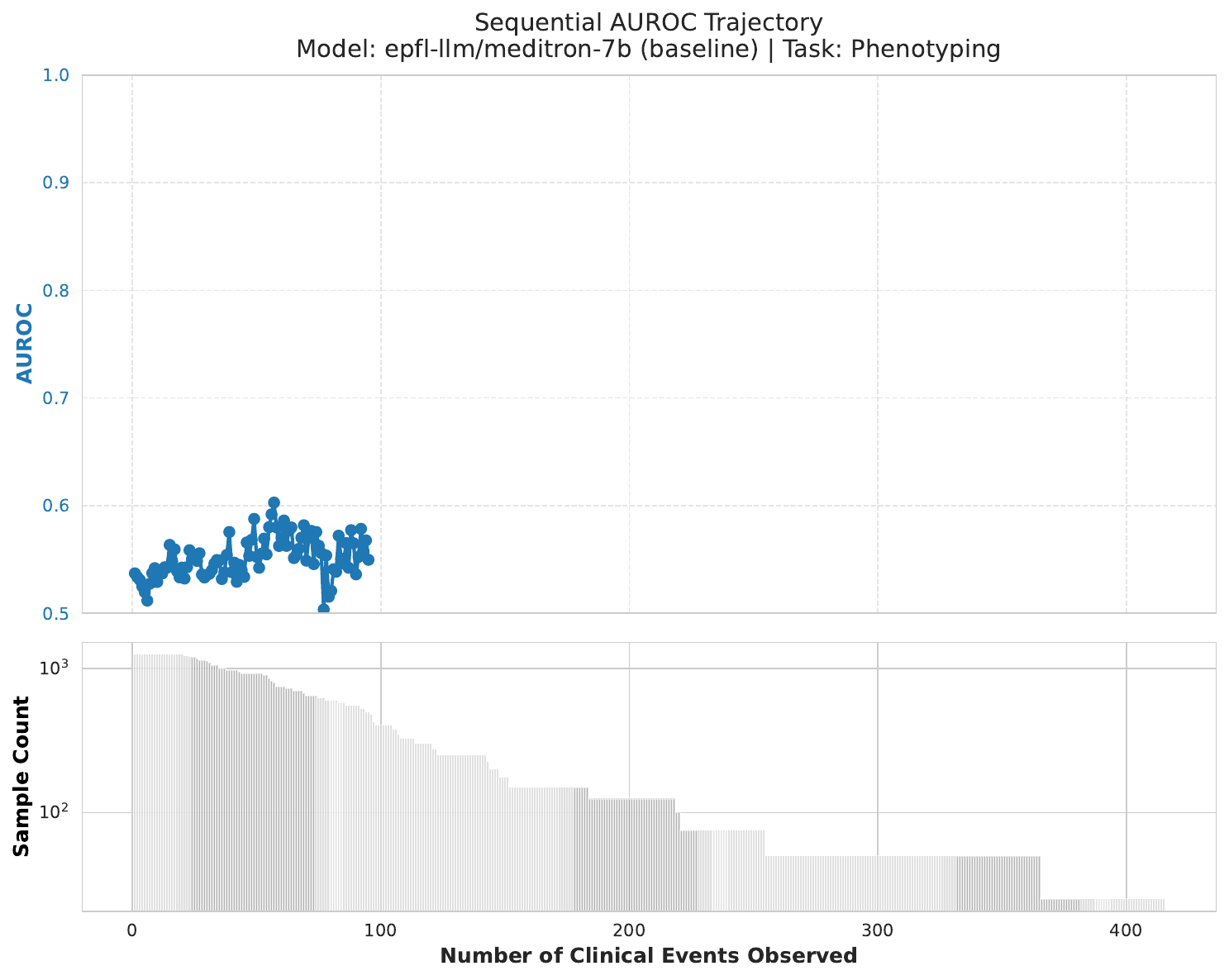}
        \caption{Non-Contrastive (Meditron)}
    \end{subfigure}
    
    \caption{Phenotyping performance plots for non-contrastive sequential models evaluated on MIMIC-IV. The AUROC performance is plotted against the patient stays with the associated number of events in their sequences, along with a visualization of the patient stay lengths.}
    \label{fig:MIMIC_phenotyping_base_trajectory}
\end{figure}

\begin{figure}[htbp]
    \centering
    % Row 1: E2E Baseline
    \begin{subfigure}{0.32\textwidth}
        \centering
        \includegraphics[width=\linewidth]{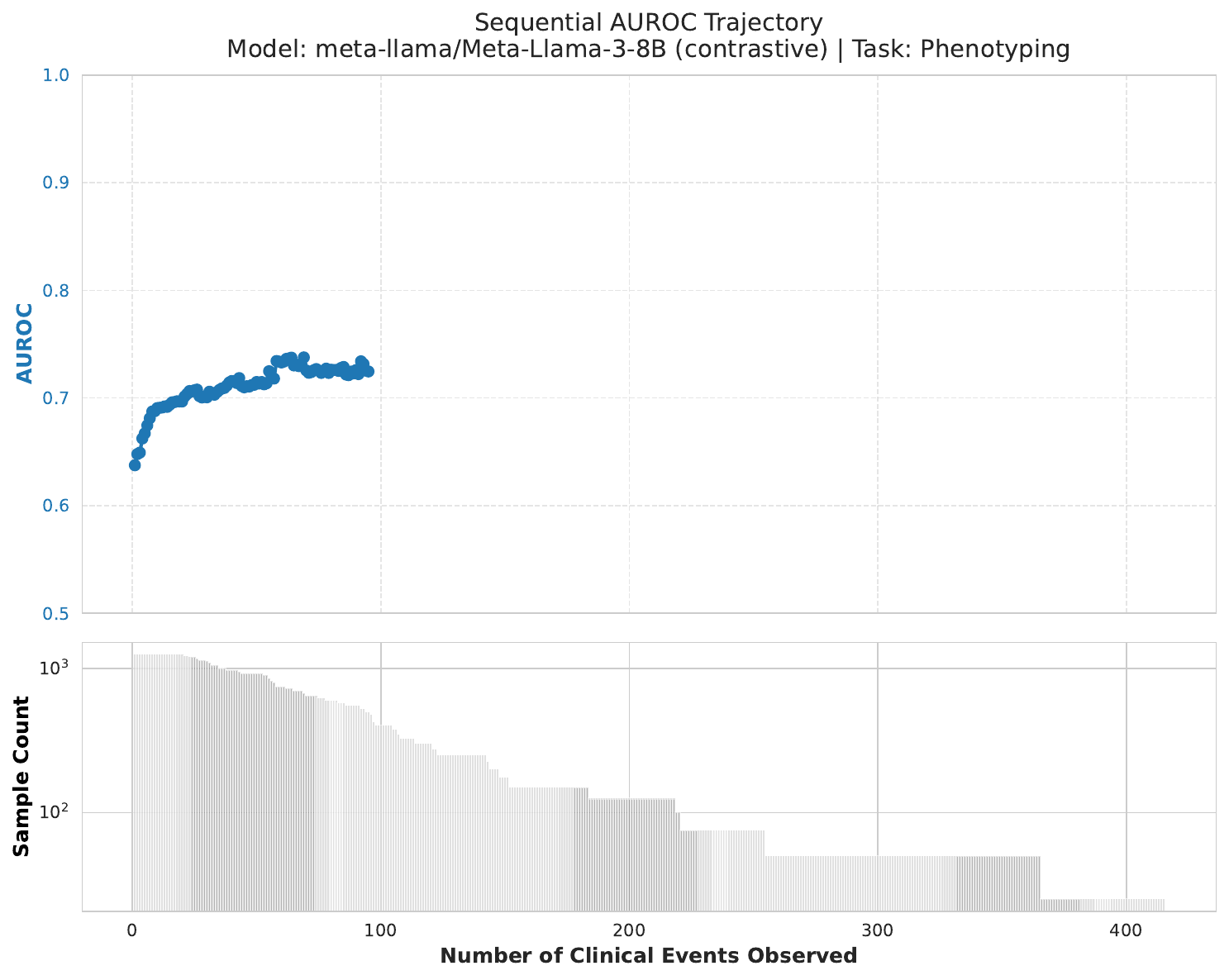}
        \caption{Contrastive (Llama-3)}
    \end{subfigure}\hfill
    \begin{subfigure}{0.32\textwidth}
        \centering
        \includegraphics[width=\linewidth]{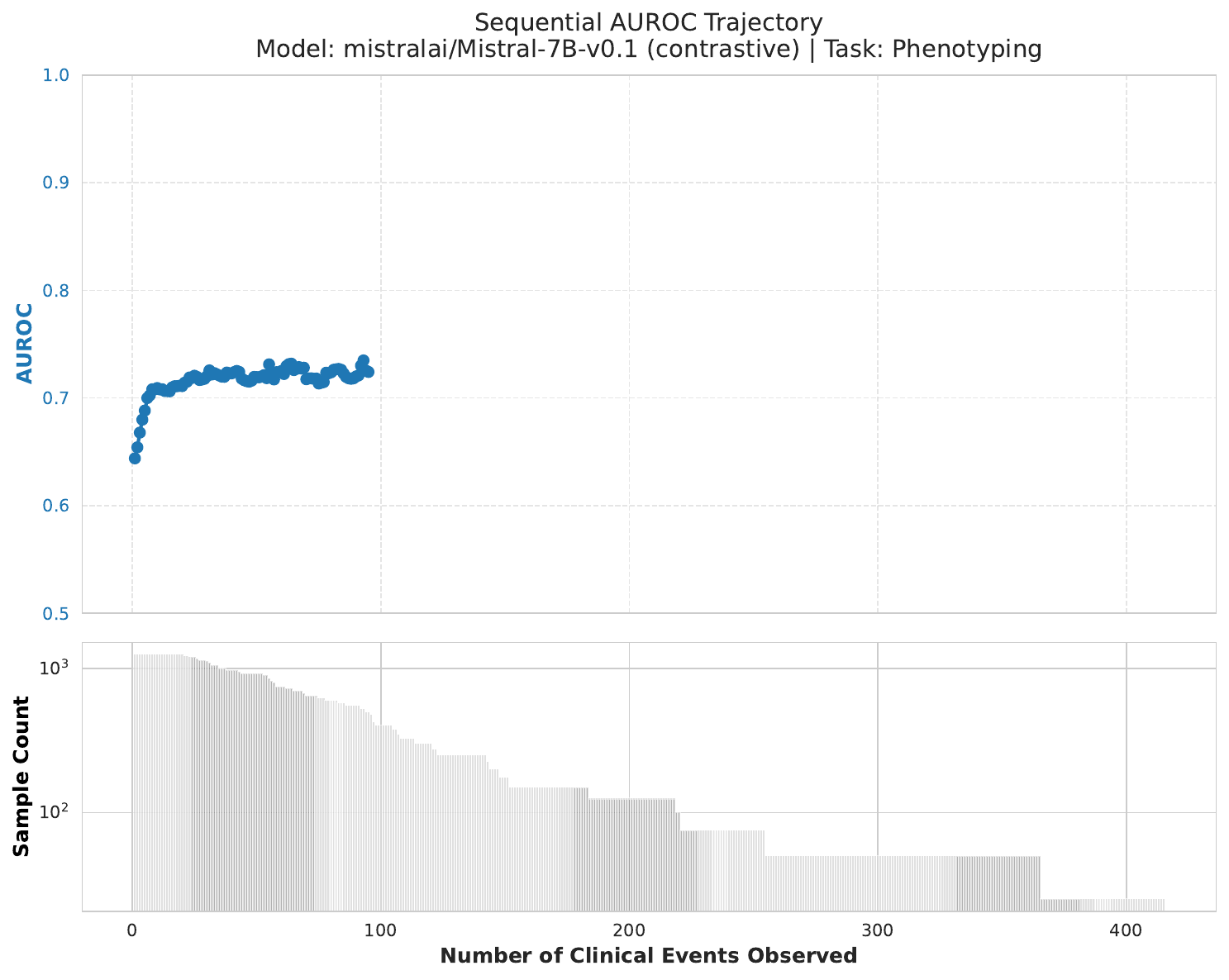}
        \caption{Contrastive (Mistral)}
    \end{subfigure}\hfill
    \begin{subfigure}{0.32\textwidth}
        \centering
        \includegraphics[width=\linewidth]{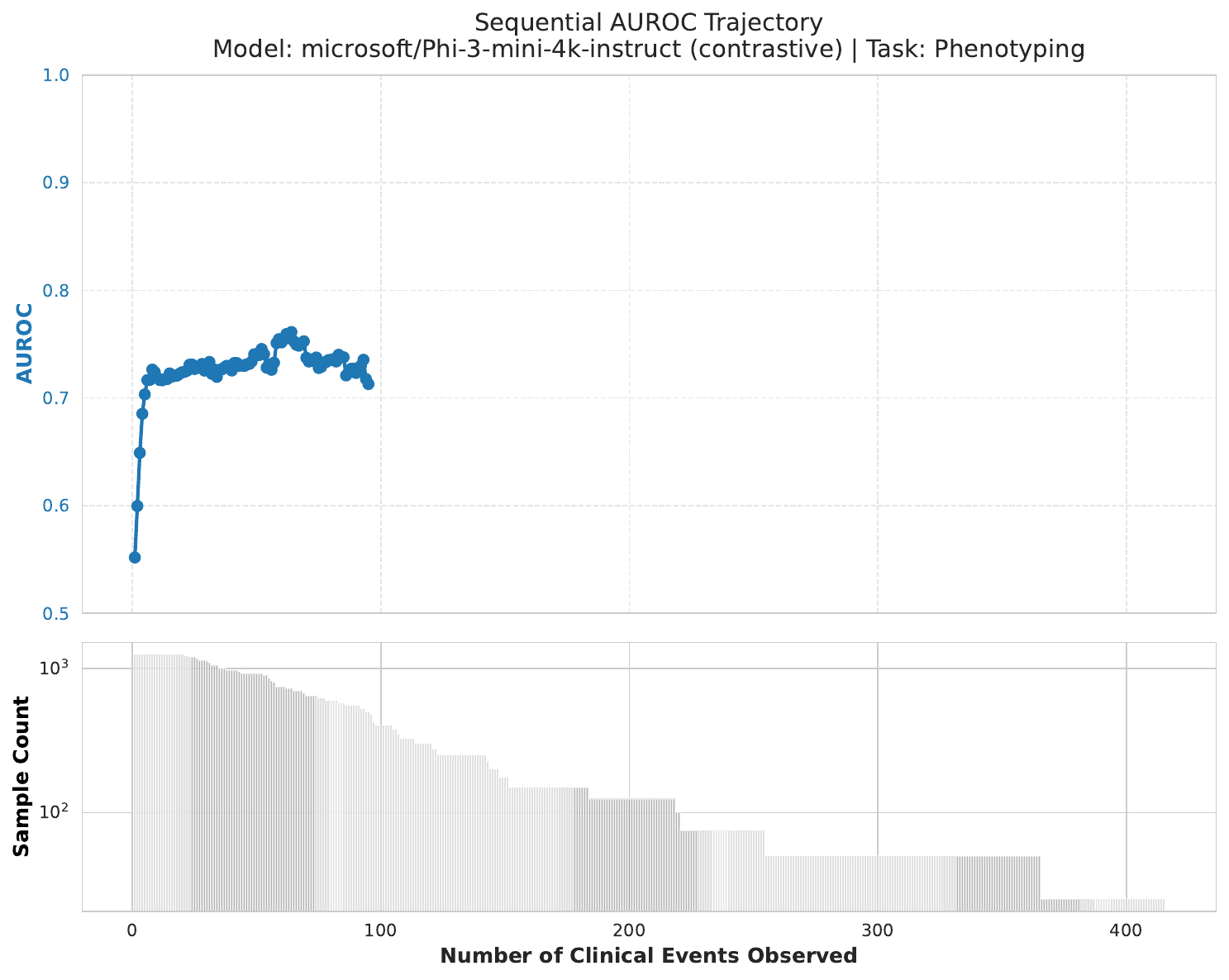}
        \caption{Contrastive (Phi-3)}
    \end{subfigure}
    
    \vspace{1em}
    
    % Row 2: Contrastive
    \begin{subfigure}{0.32\textwidth}
        \centering
        \includegraphics[width=\linewidth]{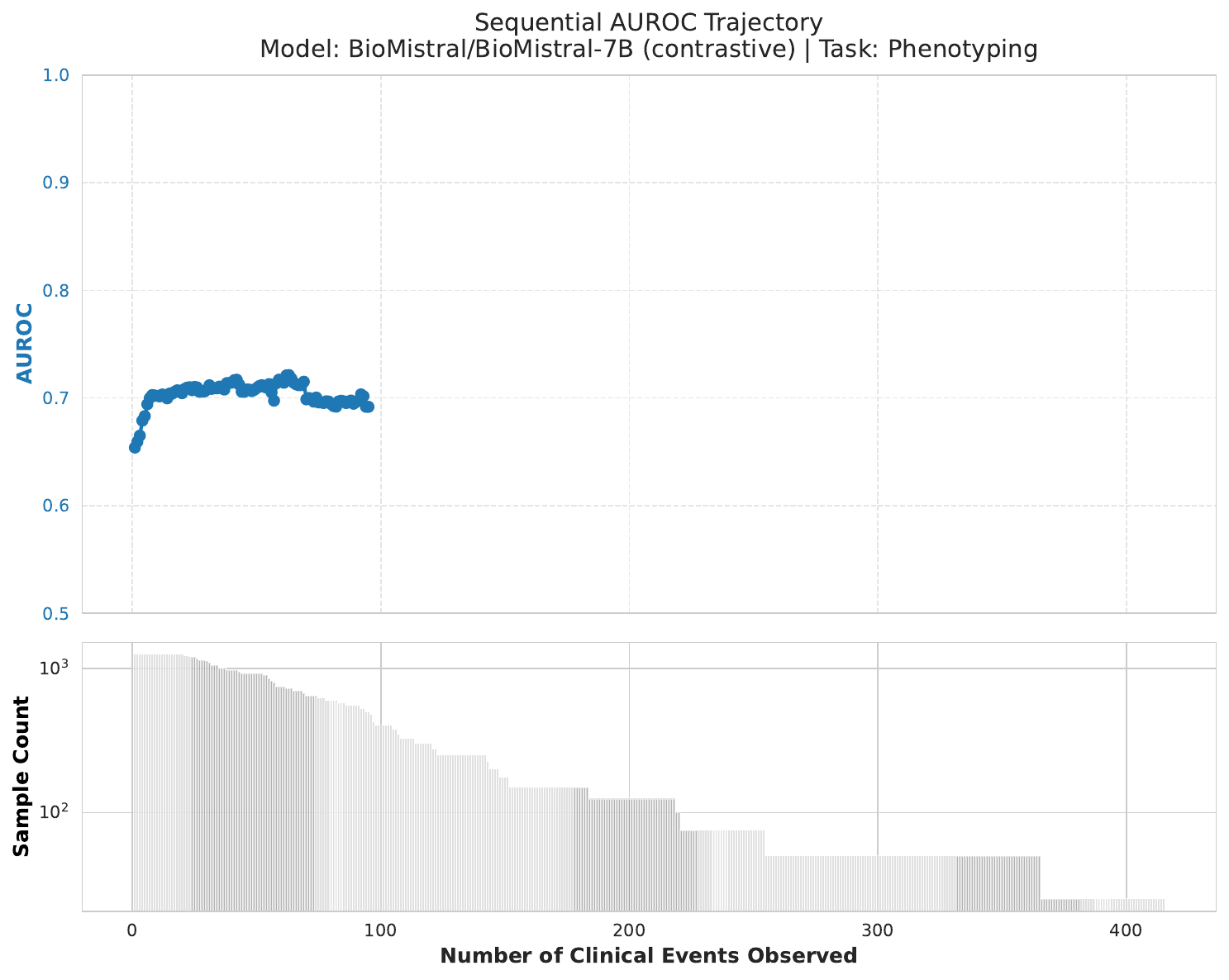}
        \caption{Contrastive (BioMistral)}
    \end{subfigure}\hfill
    \begin{subfigure}{0.32\textwidth}
        \centering
        \includegraphics[width=\linewidth]{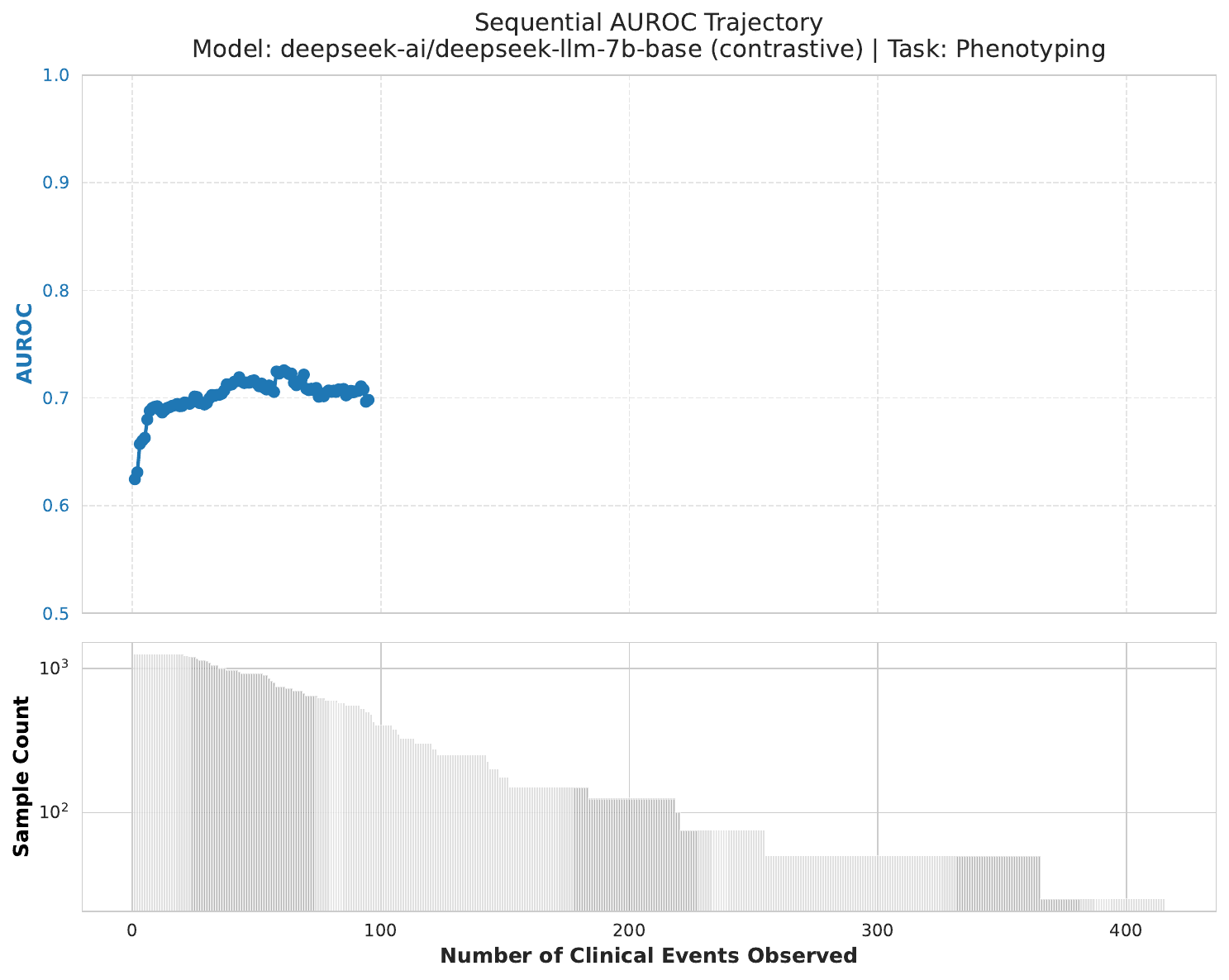}
        \caption{Contrastive (Deepseek)}
    \end{subfigure}\hfill
    \begin{subfigure}{0.32\textwidth}
        \centering
        \includegraphics[width=\linewidth]{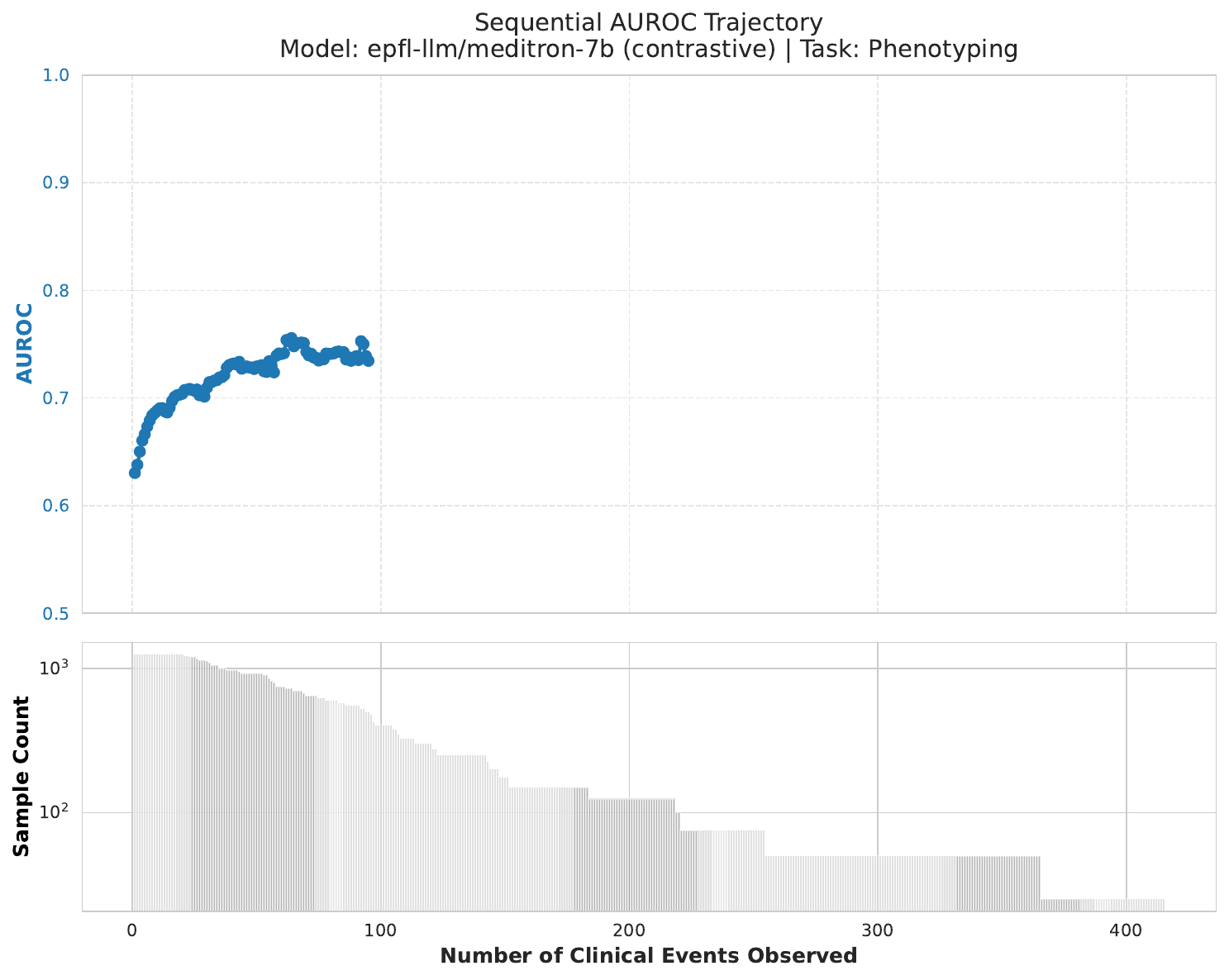}
        \caption{Contrastive (Meditron)}
    \end{subfigure}
    
    \caption{Phenotyping performance plots for contrastive sequential models evaluated on MIMIC-IV. The AUROC performance is plotted against the patient stays with the associated number of events in their sequences, along with a visualization of the patient stay lengths.}
    \label{fig:MIMIC_phenotyping_contrastive_trajectory}
\end{figure}

\clearpage

\section{Additional Case Studies and Interpretation}
\label{sec:more_interp}    
\subsection{Raw data associated with \ref{fig:reliance}:}

\begin{table}[htbp]
    \centering
    % Shrink the table to fit the exact width of the page
    \resizebox{\textwidth}{!}{%
        \begin{filecontents*}{ehr-first-hour-safe.csv}
Unnamed: 0,Hours,Capillary refill rate,Diastolic blood pressure,Fraction inspired oxygen,Glascow coma scale eye opening,Glascow coma scale motor response,Glascow coma scale total,Glascow coma scale verbal response,Glucose,Heart Rate,Height,Mean blood pressure,Oxygen saturation,Respiratory rate,Systolic blood pressure,Temperature,Weight,pH
0,0.0333333333333333,,115.0,,,,,,,,,118.0,100.0,,129.0,,,
1,0.05,,,,,,,,,92.0,,,,,,,,
2,0.0666666666666666,,,,,,,,,,,,,16.0,,,,
3,0.1833333333333333,,,,,,,,,84.0,,,,23.0,,37.5,,
4,0.2333333333333333,,,,,,,,96.0,,,,,,,,,
5,0.5666666666666667,,,,,,,,,,,,,,,,,7.3
6,0.6833333333333333,,52.0,,To Speech,Localizes Pain,,No Response-ETT,,66.0,,63.0,99.0,22.0,97.0,,,
\end{filecontents*}
\pgfplotstabletypeset[
            col sep=comma,
            string type,
            empty cells with={}, % Safely handles missing EHR data
            assign column name/.style={/pgfplots/table/column name={\detokenize{#1}}}, % Escapes underscores in headers
            every head row/.style={before row=\toprule, after row=\midrule},
            every last row/.style={after row=\bottomrule}
        ]{ehr-first-hour-safe.csv}
    }
    \caption{Raw EHR Time Series Data (Stay ID 37647460, \ref{fig:reliance}). Displaying only the first hour of admission. Missing measurements are natively represented as empty cells.} 
    \label{tab:ehr_ts_data_first_hour}
\end{table}
\begin{table}[tbhp]
    \centering
    \caption{Initial Radiology Note}
    \label{app:pre-test_ex}
    
    \begin{minipage}{\textwidth}
        \scriptsize
        \textbf{Stay ID:} 37647460 \\
        \textbf{Figure: \ref{fig:reliance}}
        \begin{lstlisting}[breaklines=true, basicstyle=\ttfamily\scriptsize, frame=single, escapeinside={(*@}{@*)}]
 INDICATION:  Trauma

TECHNIQUE:  Single supine view portable view of the chest

COMPARISON:  None.

FINDINGS: 

The endotracheal tube is seen, terminating just above the level the carinal
appear recommend withdrawal by approximately 2-3 cm for more optimal
positioning. Enteric tube is seen coursing below the diaphragm. Site for may
be at the GE junction and left could be advanced with a well within the
stomach.  Left basilar opacity may be due to atelectasis although underlying
aspiration or pulmonary contusion not excluded in the appropriate clinical
setting. No large pleural effusion or pneumothorax is seen. The cardiac
silhouette is top-normal. Mediastinal contours are grossly unremarkable. No
displaced fracture is identified.

IMPRESSION: 

Endotracheal tube low in position, terminating just above the level the
carina. Recommend withdrawal by approximately 2-3 cm for more optimal
positioning. This was discussed with Dr. at 20:15 on  via telephone.

Enteric tube courses below the level the diaphragm with side port at the level
of the GE junction and could be advanced so that it is well within the
stomach.

Left basilar opacity may be due to atelectasis and/or aspiration.
\end{lstlisting}
\end{minipage}
\end{table}

\subsection{Raw data associated with \ref{fig:mimic_safety}:}

\begin{table}[htbp]
    \centering
    \resizebox{\textwidth}{!}{%
        \begin{filecontents*}{demographics-summary-safe.csv}
admission_type,admission_location,insurance,language,marital_status,ethnicity,gender,anchor_age,anchor_year,anchor_year_group
SURGICAL SAME DAY ADMISSION,PHYSICIAN REFERRAL,Medicare,ENGLISH,MARRIED,WHITE,M,68.0,2184.0,2011 - 2013
\end{filecontents*}
\pgfplotstabletypeset[
            col sep=comma,
            string type,
            empty cells with={},
            assign column name/.style={/pgfplots/table/column name={\detokenize{#1}}},
            every head row/.style={before row=\toprule, after row=\midrule},
            every last row/.style={after row=\bottomrule}
        ]{demographics-summary-safe.csv}
    }
    \caption{Summarized Patient Demographics (Stay ID 36499784, figure \ref{fig:mimic_safety}). Extracted representing a static snapshot of the patient's admission profile.} 
    \label{tab:raw_demo_data}
\end{table}

\begin{table}[htbp]
    \centering
    \resizebox{\textwidth}{!}{%
        \begin{filecontents*}{hm.csv}
Hours,Capillary refill rate,Diastolic blood pressure,Fraction inspired oxygen,Glascow coma scale eye opening,Glascow coma scale motor response,Glascow coma scale total,Glascow coma scale verbal response,Glucose,Heart Rate,Height,Mean blood pressure,Oxygen saturation,Respiratory rate,Systolic blood pressure,Temperature,Weight,pH
0.9111111111111112,,,,,,,,228.0,,,,,,,,,7.35
\end{filecontents*}
\pgfplotstabletypeset[
            col sep=comma,
            string type,
            empty cells with={},
            assign column name/.style={/pgfplots/table/column name={\detokenize{#1}}},
            every head row/.style={before row=\toprule, after row=\midrule},
            every last row/.style={after row=\bottomrule}
        ]{hm.csv}
    }
    \caption{Raw EHR Time Series Data (Stay ID 36499784, figure \ref{fig:mimic_safety}). Displaying only the first hour of admission. Sparse or unmeasured clinical variables are natively represented as empty cells.} 
    \label{tab:ts_ehr_data}
\end{table}

\begin{figure}[htbp]
    \centering
    % Subfigure A using the subcaption package environment
    \begin{subfigure}[b]{0.48\textwidth}
        \centering
        \includegraphics[width=\textwidth]{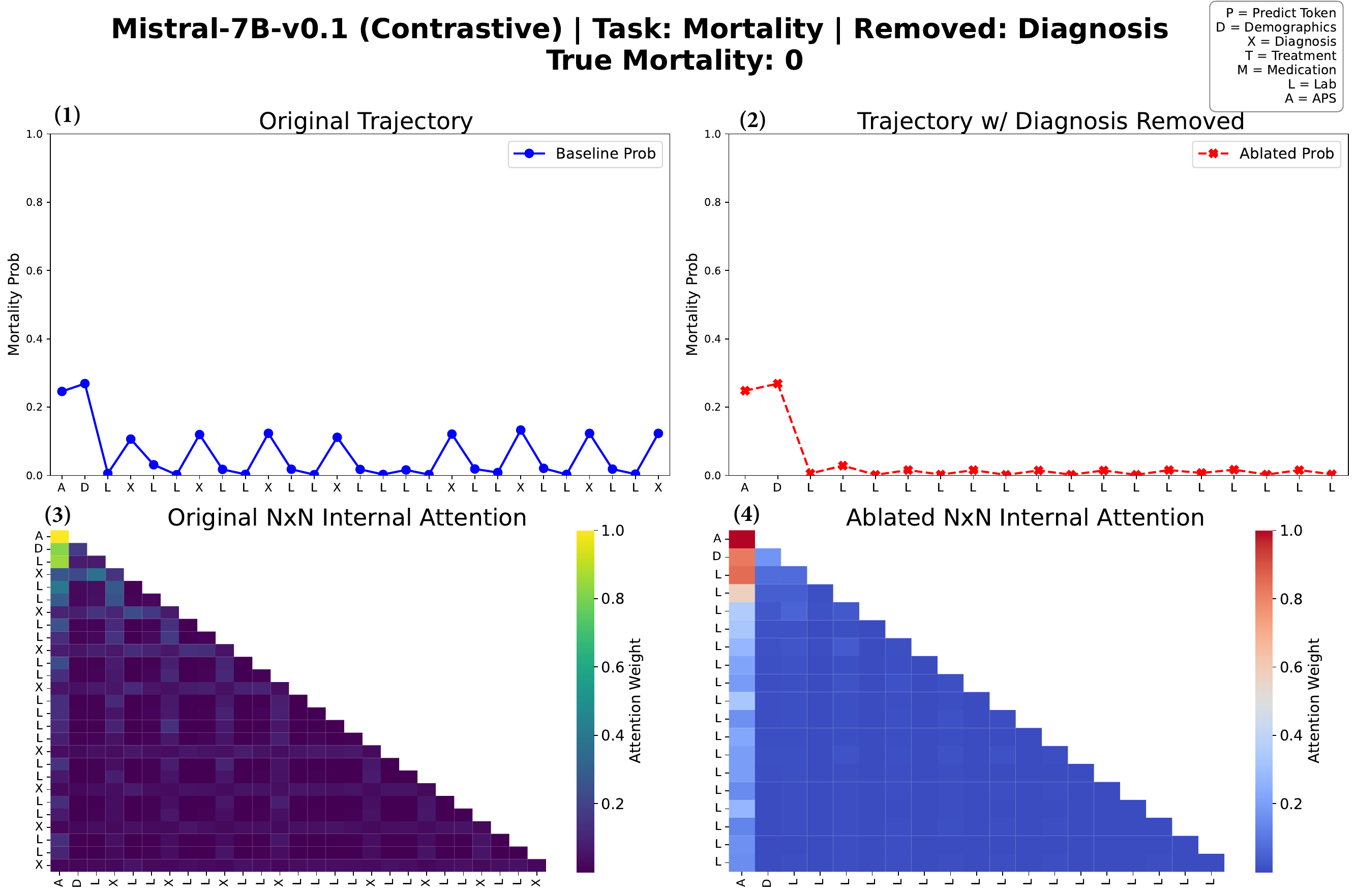}
        \caption{Sequential architecture fine-tuned on contrastively learned embeddings.}
        \label{fig:contrastive_bias_mistral}
    \end{subfigure}
    \hfill % Adds flexible space between the two subfigures
    % Subfigure B
    \begin{subfigure}[b]{0.48\textwidth}
        \centering
        \includegraphics[width=\textwidth]{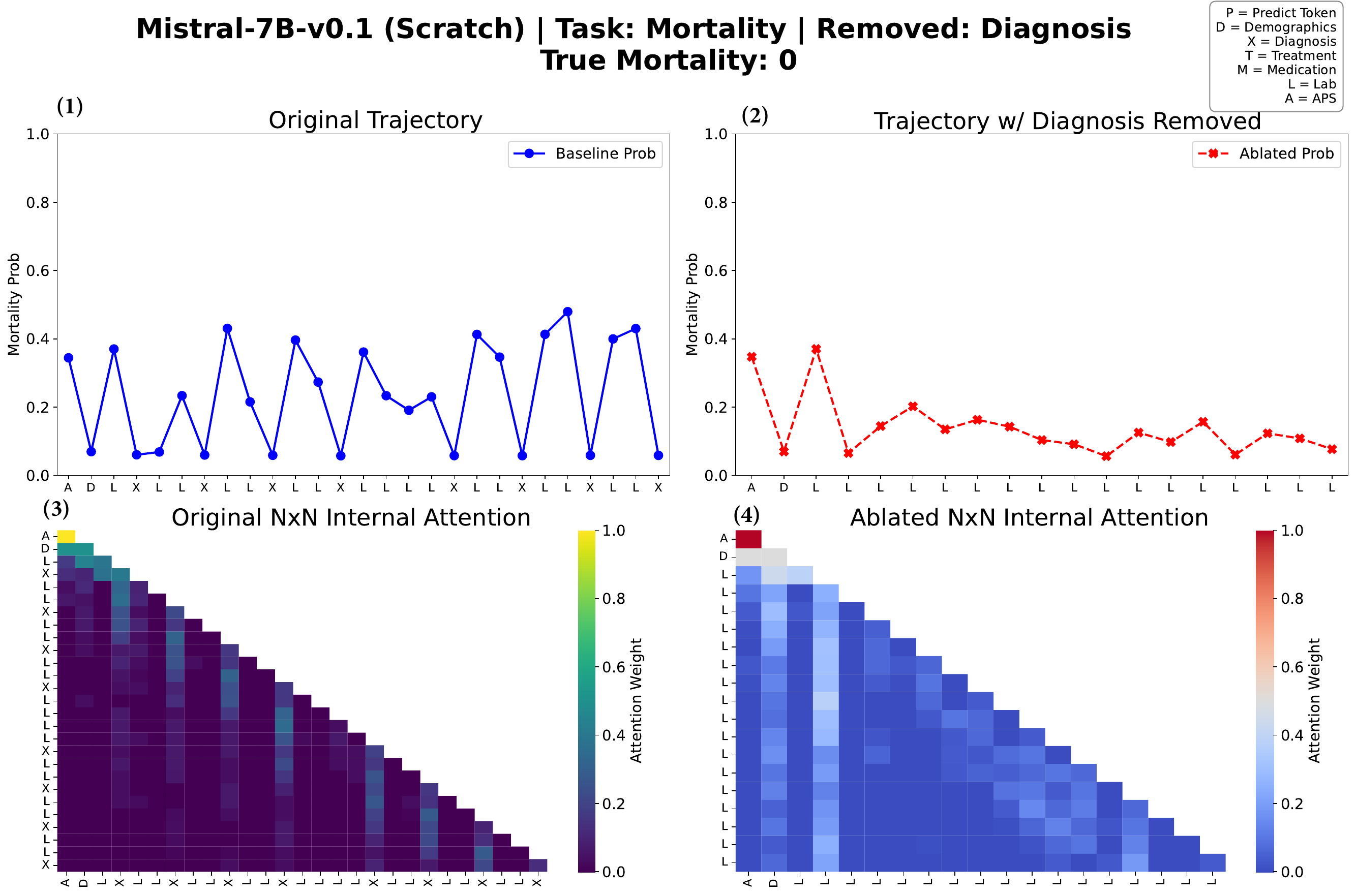}
        \caption{Sequential architecture fine-tuned on non-contrastive embeddings.}
        \label{fig:scratch_bias_mistral}
    \end{subfigure}
    
    % Main overarching caption
    \caption{We experimentally remove the diagnosis embeddings associated with a given eICU stay. For the contrastively fine-tuned model, this does not result in a change in the prediction trajectory [\textbf{(a.1)} to \textbf{(a.2)}], nor divergent behavior in the internal attention [\textbf{(a.3)} to \textbf{(a.4)}]. Removing this modality also does not alter the prediction trajectory for the non-contrastive model [\textbf{(b.1)} to \textbf{(b.2)}]. However, internal attention patterns reveal that this change introduces a new reliance on the second lab token as well as a lesser dependence on demographic information [\textbf{(b.3)} to \textbf{(b.4)}], indicating the model reached the correct conclusion based on hidden dependencies on other modalities.}
    \label{fig:mimic_safety_appendix}
\end{figure}

\end{document}